# Data-driven method for enhanced corrosion assessment of reinforced concrete structures

Woubishet Zewdu Taffese

# Abstract

Corrosion is a major problem affecting the durability of reinforced concrete structures. Corrosion related maintenance and repair of reinforced concrete structures cost multibillion USD per annum globally. It is often triggered by the ingression of carbon dioxide and/or chloride into the pores of concrete. Estimation of these corrosion causing factors using the conventional models results in suboptimal assessment since they are incapable of capturing the complex interaction of parameters. Hygrothermal interaction also plays a role in aggravating the corrosion of reinforcement bar and this is usually counteracted by applying surface-protection systems. These systems have different degree of protection and they may even cause deterioration to the structure unintentionally.

  The overall objective of this dissertation is to provide a framework that enhances the assessment reliability of the corrosion controlling factors. The framework is realized through the development of data-driven carbonation depth, chloride profile and hygrothermal performance prediction models.

  The carbonation depth prediction model integrates neural network, decision tree, boosted and bagged ensemble decision trees. The ensemble tree based chloride profile prediction models evaluate the significance of chloride ingress controlling variables from various perspectives. The hygrothermal interaction prediction models are developed using neural networks to evaluate the status of corrosion and other unexpected deteriorations in surface-treated concrete elements. Long-term data for all models were obtained from three different field experiments.

  The performance comparison of the developed carbonation depth prediction model with the conventional one confirmed the prediction superiority of the data-driven model. The variable importance measure revealed that plasticizers and air contents are among the top six carbonation governing parameters out of 25. The discovered topmost chloride penetration controlling parameters representing the composition of the concrete are aggregate size distribution, amount and type of plasticizers and supplementary cementitious materials. The performance analysis of the developed hygrothermal model revealed its prediction capability with low error. The integrated exploratory data analysis technique with the hygrothermal model had identified the surface-protection systems that are able to protect from corrosion, chemical and frost attacks.

All the developed corrosion assessment models are valid, reliable, robust and easily reproducible, which assist to define proactive maintenance plan. In addition, the determined influential parameters could help companies to produce optimized concrete mix that is able to resist carbonation and chloride penetration. Hence, the outcomes of this dissertation enable reduction of lifecycle costs.

# Tiivistelmä

Korroosio vaikuttaa merkittävästi teräsbetonirakenteiden kestävyyteen, ja sen aiheuttamat maailmanlaajuiset taloudelliset kustannukset rakenteiden huolto- ja korjaustöitten takia ovat vuosittain miljardiluokkaa. Korroosion alkusyynä on usein hiilidioksidin ja/tai kloridin tunkeutuminen huokoiseen betoniin. Perinteiset menetelmät korroosiota aiheuttavien tekijöiden arviointiin eivät ota riittävän hyvin huomioon eri tekijöiden ja parametrien vuorovaikutuksia. Hygroterminen vuorovaikutus vaikuttaa vahvisterakenteiden korroosion etenemiseen, ja sitä pyritään estämään erilaisilla pinnoiteratkaisuilla. Näiden ratkaisujen suojaavat ominaisuudet vaihtelevat, ja osin saattavat jopa aiheuttaa ei-haluttua rakenteiden heikkenemistä.

Tässä tutkielmassa kehitetään viitekehys, jonka avulla voidaan paremmin arvioida korroosiolta suojaavien ratkaisujen luotettavuutta. Kehyksen pohjalta rakennetaan datavetoisia malleja karbonatisoitumisen, kloridirasituksen sekä hygrotermisen suorituskyvyn ennustamiseen.

Karbonatisoitumista ennustetaan integroidulla mallilla, jossa käytetään neuroverkkoja sekä tehostettuja ja bootstrap-aggregoituja kokoonpanopäätöspuita. Kokoonpanopäätöspuuhun pohjautuva kloridirasituksen ennustemalli arvioi kloridien tunkeutumiseen vaikuttavien tekijöiden merkitystä monista eri näkökulmista. Hygrotermisen vuorovaikutuksen mallintamiseen kehitetään neuroverkko, joka pyrkii ennustamaan korroosiota ja muita odottamattomia pinnoitettujen betonirakenteiden vaurioita. Kaikkia malleja varten kerättiin dataa pitkältä aikaväliltä kolmesta eri kenttäkokeesta.

Karbonatisoitumista ennustavan mallin suorituskyky paljastui vertailussa paremmaksi kuin perinteisen menetelmän. Tutkimuksessa paljastui, että pehmittimen ja ilman koostumuksen vaikutukset karbonatisoitumiseen ovat kuuden tärkeimmän tekijän joukossa kahdestakymmenestäviidestä tutkitusta. Keskeisimmät betonin rakenteeseen vaikuttavista tekijöistä kloridirasituksen suhteen ovat soran raekoko, käytettyjen pehmittimien määrä ja laatu, sekä käytetyt sementiittilisämateriaalit. Hygrotermisen suorituskyvyn mallin havaittiin ennustavan suorituskykyä hyvin. Integroitu hygrotermisen suorituskyvyn malli ja data-analyysimenetelmä tunnisti pinnan suojausmenetelmät, jotka suojaavat korroosiolta, kemiallisilta sekä roudan aiheuttamilta vaurioilta.

Kaikki tutkielmassa korroosion arviointiin kehitetyt menetelmät ovat luotettavia, robusteja sekä helposti toistettavia, mikä auttaa ennakoivien kunnossapitosuunnitelmien tekemisessä. Lisäksi tutkielmassa havaittujen merkittävien korroosiota aiheuttavien tekijöiden tunnistaminen auttaa yrityksiä optimoimaan sekoitussuhteita betoninvalmistuksessa niin, että tuloksena saatavat teräsbetonirakenteet kestävät entistä paremmin karbonatisoitumista sekä kloridirasitusta, samalla vähentäen rakennusten elinkaarikustannuksia.

# Acknowledgements

The doctoral studies that led to this dissertation have been a long and challenging journey. I started my study at Department of Civil Engineering, Aalto University and finalized it at Department of Future Technologies, University of Turku. This work was made possible by financial support from various sources. I am grateful to the University of Turku Graduate School (UTUGS), Finnish Foundation for Technology Promotion (TES), Foundation for Aalto University Science and Technology, Auramo-Säätiö Foundation, Kerttu and Jukka Vuorinen Fund for providing me financial support that help me to fully concentrate on my research.

There are few people including colleagues and family members who have contributed to accomplish this journey. Firstly, I would like to express my deepest gratitude to my supervisors at University of Turku: Associate Professor Tapio Pahikkala, Professor Jukka Heikkonen and Professor Jouni Isoaho for their guidance and giving me the opportunity to finalize my doctoral study at the Department of Future Technologies. The doctoral degree would not have been possible to be realized without their support. Furthermore, I am grateful to Professor Jouni Isoaho for his time on following up closely all the official processes despite his busy schedules. Most of the research work was carried out within the Department of Civil Engineering at Aalto University. I am greatly thankful to D.Sc. (Tech.) Esko Sistonen for giving me the opportunity to start my research work, support in getting experimental data and more importantly for being my external supervisor at University of Turku. My appreciation also extended to Professor Jouni Punkki and D.Sc. (Tech.) Fahim Al-Neshawy for their kind assistance to all my requests at Aalto University.

I am also greatly thankful to the pre-examiners, Professor Mahmut Bilgehan and Associate Professor Marek Słoński, for reviewing the thesis. I would also like to thank Prof. Artūras Kaklauskas for his time and interest to this work as an opponent.

D.Sc. (Tech.) Antti Hakkala is warmly acknowledged for helping me in translating my research proposal and thesis abstract to Finnish language. It was not possible to secure some of the grants without his kind help. I am thankful to Dr. Mikhail Barash, Scientific Coordinator at TUCS, for his kind assistance during the publication of this thesis. I am grateful to D.Sc. (Tech.) Khalid Latif for trying kindly to find solutions for the challenges I had encountered. I would also like to thank my family and friends who have encouraged me throughout the years.

   Finally, my sincere gratitude with love goes to my wife Ethiopia. I am very grateful for her unconditional love, encouragement, and endurance during this long journey. She has stood by me throughout this journey and has made countless sacrifices for helping me get through the difficult times in the most positive way. She has been my inspiration and motivation for continuing to enhance my knowledge and move my research forward. Words cannot express my heartfelt appreciation and love for her infinite support, and for being most intellectually curious partner that I could hope for.

Turku, June 2020

Woubishet Zewdu Taffese

# Contents

x

# List of abbreviations and symbols

| | |
|---|---|
| AEA | Air-entraining agents |
| BFS | Blast-furnace slag |
| CV | Cross validation |
| EDA | Exploratory data analysis |
| EN | European Standard |
| FA | Fly ash |
| HW | Highway |
| ITZ | Interfacial transition zone |
| LSBoost | Least-squares boosting |
| MAE | Mean-absolute error |
| MSE | Mean-square error |
| NARX | Nonlinear autoregressive with exogenous inputs |
| R | Correlation coefficient |
| RC | Reinforced concrete |
| RCM | Rapid chloride migration |
| RH | Relative humidity |
| RMSE | Root-mean-square error |
| SCC | Self-compacting concrete |
| T | Temperature |
| VI | Variable importance |
| w/b | Water- to-binder ratio |
| w/c | Water-to-cement ratio |

# List of publications

The work discussed in this doctoral dissertation is based on the synthesis of the following original publications.

## Publications included in the thesis

1. Publication I

    **Taffese, Woubishet Zewdu**; Sistonen, Esko. 2017. Machine learning for durability and service-life assessment of reinforced concrete structures: Recent advances and future directions. Elsevier. Journal of Automation in Construction, volume 77, pages 1–14. ISSN 0926-5805. https://doi.org/10.1016/j.autcon.2017.01.016.

2. Publication II

    **Taffese, Woubishet Zewdu**; Sistonen, Esko; Puttonen, Jari. 2015. CaPrM: Carbonation prediction model for reinforced concrete using machine learning methods. Elsevier. Journal of Construction and Building Materials, volume 100, pages 70–82. ISSN 0950-0618. https://doi.org/10.1016/j.conbuildmat.2015.09.058.

3. Publication III

    **Taffese, Woubishet Zewdu**; Sistonen, Esko. 2017. Significance of chloride penetration controlling parameters in concrete: Ensemble methods. Elsevier. Journal of Construction and Building Materials, volume 139, pages 9–23. ISSN 0950-0618. https://doi.org/10.1016/j.conbuildmat.2017.02.014.

4. Publication IV

   **Taffese, Woubishet Zewdu;** Sistonen, Esko. 2016. Neural network based hygrothermal prediction for deterioration risk analysis of surface-protected concrete façade element. Elsevier. Journal of Construction and Building Materials, volume 113, pages 34–48. ISSN 0950-0618. https://doi.org/10.1016/j.conbuildmat.2016.03.029.

**Publications not included in the thesis**

The following publications are related to the research but are not appended in the thesis.

5. Publication V

   **Taffese, Woubishet Zewdu**; Nigussie, Ethiopia; Isoaho, Jouni. 2019. Internet of things based durability monitoring and assessment of reinforced concrete structures. Elsevier. Procedia Computer Science, volume 155, pages 672–679. ISSN 1877-0509. https://doi.org/10.1016/j.procs.2019.08.096.

6. Publication VI

   **Taffese, Woubishet Zewdu**; Al-Neshawy, Fahim; Sistonen, Esko; Ferreira, Miguel. 2015. Optimized neural network based carbonation prediction model, International Symposium on Non-Destructive Testing in Civil Engineering (NDT-CE), Berlin, Germany. pages 1074–1083.

7. Publication VII

   **Taffese, Woubishet Zewdu**; Sistonen, Esko; Puttonen, Jari. 2015. Prediction of concrete carbonation depth using decision tree. Proceedings of the 23rd European Symposium on Artificial Neural Networks, Computational Intelligence and Machine Learning (ESANN 2015), Bruges, Belgium. pages 415–420.

8. Publication VIII

   **Taffese, Woubishet Zewdu**; Sistonen, Esko. 2013. Service-life prediction of repaired concrete structure using concrete recasting method: State-of-the-art. Elsevier. Procedia Engineering, volume 57, pages 1138–1144, ISSN 1877-7058 https://doi.org/10.1016/j.proeng.2013.04.143.

9. Publication IX

   **Taffese, Woubishet Taffese**; Al-Neshawy, Fahim; Piironen, Jukka; Sistonen, Esko; Puttonen, Jari. 2013. Monitoring, evaluation and long-term forecasting of hygrothermal performance of thick-walled concrete structure. Proceedings of OECD/NEA WGIAGE Workshop on Nondestructive Evaluation of Thick-Walled Concrete Structures, Prague, Czech Republic. pages 121–143.

# Chapter 1

# Introduction

## 1.1 Background

Corrosion is one of the foremost critical problems affecting the durability of reinforced concrete (RC) structures throughout the world [1–3]. Several studies revealed that corrosion related maintenance and repair of RC structures cost multibillion USD per annum globally. Repairing of corrosion-induced damage in Western Europe alone causes the loss of 5 billion EUR annually [4]. Some developed countries even expend about 3.5% of their gross national product for corrosion associated damage and its control [5]. Even in repaired RC structures, continued corrosion of reinforcement bar accounts for 37% of the failure modes [6], causing costly repairs of repairs [6–8].

The process of corrosion in RC structures is divided into two general phases: initiation and propagation. In the initiation phase the aggressive substances, carbon dioxide ($CO_2$) and chloride ions ($Cl^-$), are transported through the concrete pores towards the surface of the reinforcement bar. Since concrete is alkaline with a pore solution pH of 12–13 that protects the embedded reinforcement bar from corrosion by forming a thin oxide layer on its surface. This layer deteriorates in the presence of $Cl^-$ or due to the carbonation of concrete [9,10]. Carbonation is a physicochemical phenomenon induced naturally by the ingression of $CO_2$ into the concrete pores from the surrounding environment and reacts with hydrated cement. The propagation phase covers the time from the onset of reinforcement bar corrosion to structural failure. In this phase, the deterioration of reinforcement bar depends on the corrosion rate, which is mainly governed by two environmental agents: moisture and temperature. These agents control the corrosion rate through their effect on the electrochemical reactions [11,12].

Carbonation- and chloride-induced corrosion can diminish the cross-sectional area and the elongation capacity of the reinforcement bar. This causes severe cracking as well as reduction in the load-bearing capacity of the structure. Cracked concrete could allow an additional entry of moisture, aggressive gasses and ions, aggravating reinforcement bar corrosion and concrete degradation. Subsequently, the strength, safety and serviceability of the RC structures will be declined.

In order to perform in-time and cost-effective maintenance and repair decisions, the initiation time of corrosion has to be reliably estimated. In practice, simplified Fick's second law based models are extensively applied for predicting carbonation depth and chloride concentration inside the concrete. Most of these models and the associated value of input parameters have been oversimplified, incomplete, and/or unsuitable for the prevailing conditions [13–15]. The use of these oversimplified models lacks the ability to capture the complex interactions among the involving parameters, causing suboptimal or even improper choice of design and maintenance strategies. A better understanding of the complex interacting parameters that controls the corrosion of reinforcement bar is a crucial step towards the development of reliable models. Indeed, examination of $CO_2$ and $Cl^-$ transport in concrete is performed for several years to acquire a better understanding of various controlling parameters. Nevertheless, it is usually challenging to isolate the influences of particular parameters because other controlling parameters are also varying naturally at the same time [16,17]. Identifying the influential predictors using traditional statistical methods, such as linear regression method is unachievable since the penetration of these aggressive substances in field concrete is a highly complicated process involving several nonlinear interactions among the parameters. Determining powerful predictors based on linear regression method is only applicable for linear or nearly linear models. Hence, alternative approaches that are capable of managing multidimensional nonlinear parameters are necessary in order to determine the influential predictors reliably.

The conventional carbonation depth and chloride concentration prediction models were established based on short-term tests. It has been proven from many experimental data that these simplified models, especially those developed to predict chloride profile, can only characterize the chloride penetration under the exposure conditions for a short period of time close to the conditions for which the input parameters of the model were deduced [18]. There is no a straightforward method that can be applied for the translation of short-term field data into values that can express the long-term performance of concrete structures exposed to field environment. Fortunately, several long-term field data are nowadays available from real structures and specimens made with several mix compositions exposed to carbonation and chloride environment. With the advancement of nondestructive technologies for monitoring of corrosion controlling parameters in concrete structure, the availability of field data subjected to different exposure conditions will increase significantly. Huge efforts are being made in the past in other industries to establish powerful data-driven methods that can be utilized to perform accurate predictions and extract useful information. Thus, in construction industry, examining the possibilities of predicting corrosion controlling parameters using readily available long-term field data as well as translating them into useful knowledge is essential.

In Finland, corrosion-induced damage on prefabricated RC façade is projected to be around 15 million m$^2$ per annum and will grow 2% each year [19]. It accounts for about 11–40% of the overall repair costs depending on the surface-finishing material types [20]. Substantial attempts have been made into devising cost-effective repair methods to control the rate of corrosion of reinforcement bar. Surface-protection system is one of the economical methods that are widely used to curb the corrosion rate by controlling the moisture of concrete [11,21]. The surface-protection system may have different degrees of protection against moisture even with identical generic chemical composition. They may even cause unintended damage to the structure since the compositions of the materials differ extensively [21]. Therefore, clear understanding of the hygrothermal performance of surface-protected concrete is essential since uncontrolled hygrothermal interaction may accelerate the corrosion of the embedded reinforcement bar. Certainly, hygrothermal transport phenomena in concrete and several other building materials are thoroughly understood and numerical models have already been established [22]. Though models to predict the hygrothermal behaviour of concrete have been suggested in the past, none has explicitly integrated diverse surface-protection materials and application procedures in their model. It is also challenging to provide satisfactory analytical methods to assess the performance of the protective measures since understanding the interaction of various types of surface-protection systems with the substrate concrete in very dynamic environmental conditions is highly complex. In-service hygrothermal monitoring utilizing sensors is a better alternative for assessing the performance of surface-protected RC concrete elements. Using the sensor data, the hygrothermal interaction can be forecasted, which leads to understanding of the condition of corrosion. In addition, using the forecasted data and exploratory data analysis appropriate surface-protection system can be determined.

## 1.2 Research problem

The research problem of the dissertation is: *how to make valid, reliable, and robust corrosion assessment of RC structures through realization of the complex processes of carbonation, chloride ingress and hygrothermal interaction?* The research problem is approached through the following more detailed research questions:

**Research question one:** How to eliminate or mitigate the uncertainties observed in the traditional corrosion assessment methods?

**Research question two:** How to develop accurate carbonation depth prediction model that considers the complex parameter interactions? What are the predominant carbonation depth predictors?

**Research question three:** What are the significant parameters that describe the chloride concentration into concrete?

**Research question four:** How to predict the hygrothermal interaction inside surface-protected concrete while identifying the appropriate surface-protection system?

Each of the above research questions is answered in one of the annexed publications and each article contributes towards addressing the overall research problem. The logic on how the publications are interconnected in answering the research problem is illustrated in Figure 1.1. The first publication answers the research question one. This publication examines the recent advances and current practices of corrosions assessment of RC structures and recommends methods that mitigate the uncertainties observed in the conventional approaches. Research question two is addressed by Publication II, which presents the development of an optimized and integrated data-driven carbonation prediction model. It also determines influential carbonation depth predicting parameters. Research question three is answered by Publication III. This publication evaluates the importance of variables that characterize the chloride concentration in concrete. The fourth publication provides answer for the research question four. It demonstrates how hygrothermal interaction inside surface-protected concrete can be predicted using data-driven models while identifying the appropriate surface-protection system.

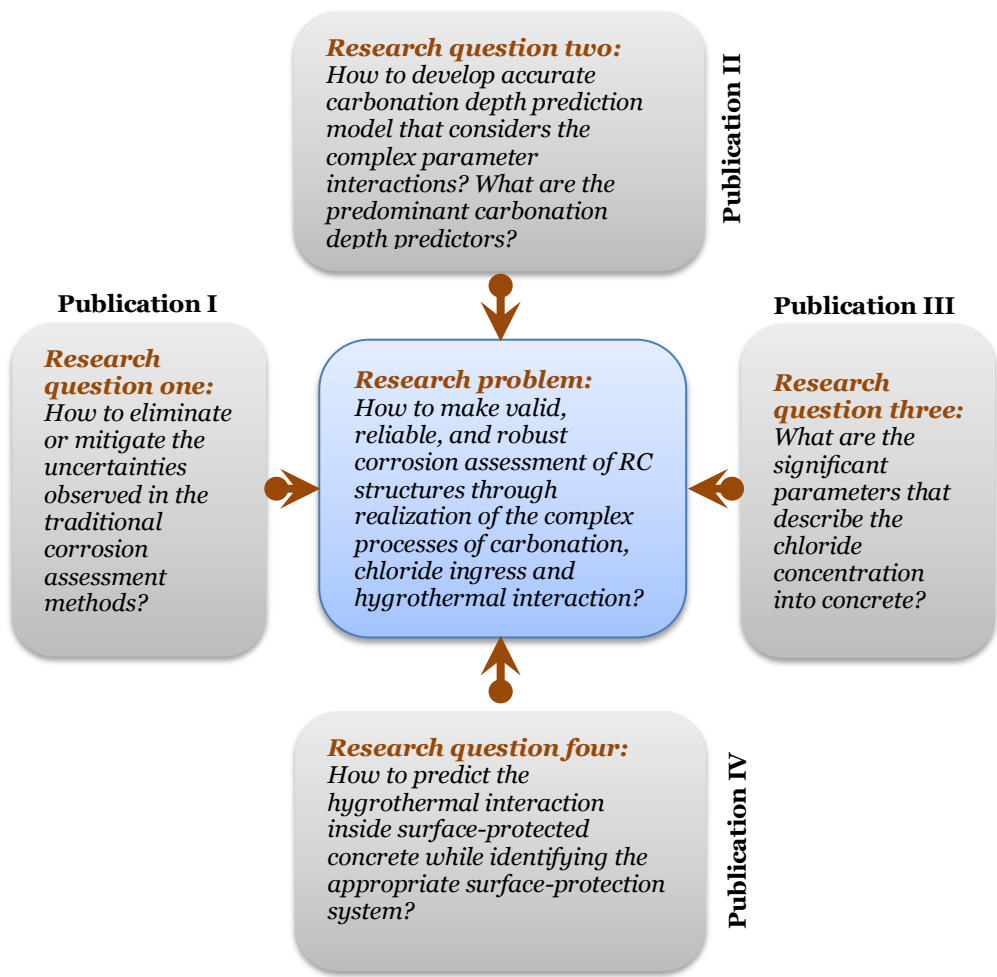

**Figure 1.1.** The logic on how the publications are interconnected in answering the research problem.

## 1.3 Aim and objectives of the research

The aim of this dissertation is *to provide a framework that enhances the assessment reliability of the corrosion controlling factors*. The framework is realized through the development of data-driven models that evaluate corrosion causing and accelerating factors. The objectives are: i) to reliably foresee the carbonation depth and to determine the influential predictors of carbonation depth; ii) to identify the significance of chloride penetration characterizing parameters; and iii) to evaluate the hygrothermal behaviour of surface-protected concrete elements.

A reliable prediction of carbonation depth, chloride profile and hygrothermal behaviour is instrumental for realizing optimal design and maintenance plans for RC structures. The determination of influential predictors will also yield scientific significance that could assist concrete researchers to design concrete mix that resist carbonation and chloride penetration. All these enable a considerable minimization in lifecycle costs, which in turn prevents economic loss.

## 1.4 Research methodology and dissertation structure

All the methods that are devised to assess corrosion causing and aggravating factors in this dissertation rely on machine learning algorithms and implemented using MATLAB programming language. The methods are utilized for developing data-driven models. As any data-driven models, the development process of these models primarily consists of data, data preprocessing and training. The experimental data employed for model development and testing are gathered from field experiments.

The thesis is divided into six chapters. Chapter Two presents the theoretical foundation of the thesis by focusing on corrosion causing and controlling factors as well as the limitations of the conventional carbonation and chloride profile prediction models. It also discusses the need for hygrothermal-behaviour prediction approaches for surface-protected concrete structures. The advantages of data-driven methods that mitigate the limitations of the conventional models and the fundamental concepts of the machine learning algorithms that are used in developing the data-driven models are discussed in the same chapter.

The materials and methods that are employed to address the research problem of the dissertation are discussed in Chapter Three. It presents the development process of data-driven models that predict carbonation depth, chloride profile and hygrothermal behaviour. As data-driven methodologies, the materials part of this chapter discusses the utilized experimental data for the development of carbonation, chloride and hygrothermal models in details.

The core results of the dissertation are presented in Chapter Four. It has four sections where each section answers the corresponding research questions. It presents the advantages of data-driven approaches in mitigating uncertainties that are observed in the conventional corrosion prediction methods. The performances of the developed models for carbonation depth prediction, and for determining significant predictors of carbonation depth and chloride concentration in concrete are discussed. The performance of the hygrothermal behaviour prediction model for surface-protected concrete element is also discussed in the same chapter.

Chapter Five presents the discussion of the research results in the context of theoretical and practical implications of the research as well as its reliability and validity. In addition, recommendations for further research are proposed in the same chapter. Finally, the conclusion of the dissertation is presented in Chapter Six.

## 1.5   Scope of the research

The scope of the dissertation is to examine and model corrosion causing and controlling factors by focusing on carbonation, chloride penetration and hygrothermal interaction. All the case structures or experiments in the dissertation represent natural Finnish climate. The hygrothermal, carbonation and chloride case structures are located in Vantaa, Espoo and Kotka, respectively. The environmental exposure classes for examination of carbonation, chloride ingress and hygrothermal are XC3 (moderate humidity), XD3 (cyclic wet and dry), and XF1 (moderate water saturation, without deicing agent), respectively. The maximum exposure time of the case structures is two, six, and seven years for hygrothermal, chloride, and carbonation investigation, respectively.

# Chapter 2

# Theoretical foundation

In this chapter, the corrosion process and factors that cause and control corrosion are presented. The limitations of the conventional corrosion assessment methods are also discussed. The fundamental principles of novel data-driven approach that are able to address the recognized limitations of the conventional methods are presented.

## 2.1 Corrosion process

Corrosion of reinforcement bar is an electrochemical process. The electrochemical potentials to generate the corrosion cells are induced in two mechanisms [9,10,12,23]: i) when two different types of metal are embedded in concrete or when there is a considerable dissimilarities in the surface characteristics of the reinforcement bar, and ii) when there is concentration differences in the dissolved ions at the vicinity of the reinforcement bar surface. As a result, in cases of mechanism (i), one of the two metals becomes anodic and the other cathodic. While, in case of mechanism (ii), some parts of the reinforcement bar begin to be anodic and the other part of the reinforcement bar becomes cathodic. The primary chemical changes occurring at the anodic and cathodic areas as well as the resulting rust formation are described by Equations (2.1) to (2.4). Furthermore, the corrosion of reinforcement bar in concrete as an electrochemical process is schematically illustrated in Figure 2.1.

At the anode, oxidation of iron occurs:

$$\text{Fe} \rightarrow \text{Fe}^{++} + 2\text{e}^-. \tag{2.1}$$

At the cathode, reduction of atmospheric oxygen with water occurs:

$$\tfrac{1}{2}\text{O}_2 + \text{H}_2\text{O} + 2\text{e}^- \rightarrow 2(\text{OH})^-. \tag{2.2}$$

When anodic and cathodic reaction products combine:

$$Fe^{++} + 2(OH)^- \rightarrow Fe(OH)_2. \qquad (2.3)$$

Subsequent oxidation reaction results in the formation of rust:

$$FeO.(H_2O)_x. \qquad (2.4)$$

The anodic and cathodic reaction is responsible for the formation of primary corrosion product of metal, $Fe(OH)_2$, but the action of $O_2$ and $H_2O$ can yield other corrosion products with different colours [9,24,25]. Corrosion is often accompanied by a loss of reinforcement bar cross-sectional area and accumulation of corrosion products which invade a larger volume (usually 2 to 6 times) than the original reinforcement bar [9,11]. The corrosion product exerts substantial tensile stresses, causing cracking and spalling of the concrete cover. So, structural distress may gradually occur due to the bond loss between reinforcement bar and concrete or due to loss of reinforcement bar cross-sectional area [26,27].

The concrete cover provides both chemical and physical protections against corrosion of reinforcement bar. Chemical changes caused by the ingression of aggressive substances into concrete deteriorate the oxide layer at the surface of the reinforcement bar, causing initiation of corrosion. The concrete can also lose its protection capability due to cracking. Cracking of concrete let penetration of corrosion causing and accelerating agents, such as moisture, aggressive gasses, and ions to the vicinity of reinforcement bar surface.

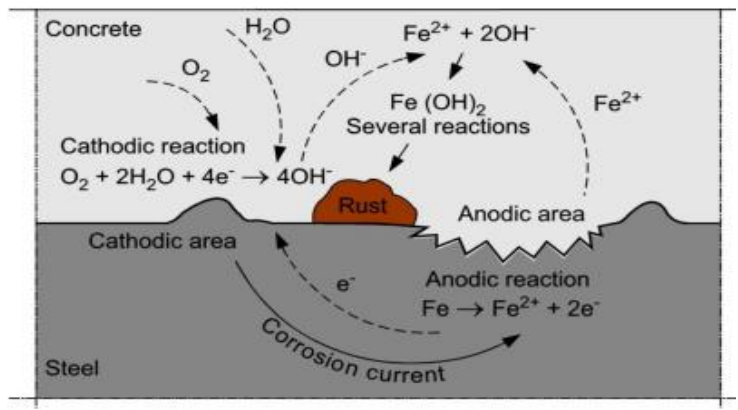

**Figure 2.1.** Schematic illustration of rebar corrosion in concrete as an electrochemical process [19].

## 2.2 Corrosion causing and controlling factors

Corrosion of reinforcement bar is typically triggered by the ingression of $CO_2$ and $Cl^-$ into the concrete pores. Once the corrosion of reinforcement bar is initiated, the corrosion rate is mainly controlled by the environmental agents, which are moisture and temperature. Since the focus of the dissertation is on the above corrosion initiating and controlling factors, the process of carbonation, chloride ingression and hygrothermal interaction are discussed below.

### 2.2.1 Carbonation and chloride ingress

Normally, concrete is alkaline with a pore solution pH of 12–13 that passivizes the embedded reinforcement bar. The passivation of reinforcement bar is breakdown due to the existence of $Cl^-$ or by the carbonation of the concrete [9–11]. Carbonation is a natural physicochemical process caused by the ingression of $CO_2$ from the neighbouring environment into the concrete through pores in the matrix where the $CO_2$ reacts with hydrated cement [28,29]. The chemical reaction of carbonation process is expressed in Equation (2.5). Calcium hydroxide ($Ca(OH)_2$) in contact with carbon dioxide ($CO_2$), in the presence of moisture, forms calcium carbonate ($CaCO_3$). This chemical reaction slowly lowers the alkalinity of the pore fluid from a pH value of about 13 to 9 [10,29–32]. Though the depletion of alkalinity caused by carbonation alters the chemical composition of concrete, its major consequence is that it destroys the passive oxide layer of reinforcement bar which ultimately initiates corrosion [10,28–35].

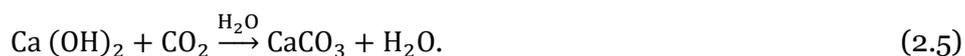

$$\text{Ca (OH)}_2 + \text{CO}_2 \xrightarrow{H_2O} \text{CaCO}_3 + H_2O. \qquad (2.5)$$

The ingression of chloride ions is also the principal cause for corrosion of reinforcement bar in concrete. Similar to carbonation, the ingression of chloride ions does not normally cause damage to the concrete directly. Nevertheless, when the amount of chloride concentration at the reinforcement bar reaches a certain threshold, depassivation occurs that initiates corrosion [36–40]. Chloride ions exist in the cement paste surrounding the reinforcement bar react at anodic sites to form hydrochloric acid which obliterates the passive protection layer on the surface of reinforcement bar. The surface of the reinforcement bar then becomes activated locally to form the anode, while the passive surface forming the cathode, resulting corrosion initiation in the form of localized pitting [10,11,41]. Chloride attack is one of the primary threats for the durability of RC structures that are exposed to marine environment and deicing salts containing chloride [36–40]. The ingression of $Cl^-$ is notable in countries at those latitudes where substantial amounts of deicing salts are spread on the roads to melt the ice during the winter. The melted ice slurry with intensely concentrated $Cl^-$ from deicing salt splashes to RC structures by the moving vehicles. There is even a study that claim $Cl^-$ from deicing salt is detected as high as 60[th] floor of RC building structure located 1.9 km from a busy highway [42].

As presented above, the alkaline environment that shields the reinforcement bar is vulnerable to deterioration, either by carbonation or chloride attack. The penetration rate of these aggressive substances to breakdown the passive film is a function of the quality and the thickness of the concrete cover as well as the surrounding environment. Generally, the corrosion process of reinforcement bar in concrete is divided into two stages: initiation and propagation. In case of carbonation-induced corrosion, the corrosion initiation stage corresponds to the time required for the carbonation front to arrive at the surface of reinforcement bar. In chloride-induced corrosion, the corrosion initiation stage corresponds to the period for the $Cl^-$ concentration to reach at a specific threshold level that damages the protective layer. Once the protective layer has broken, corrosion can onset and accelerate very fast in the presence of moisture and oxygen. The time taken from onset of corrosion to concrete failure is known as the propagation period. The initiation stage compared with the propagation period is long. It means that, if appropriate measures are not timely taken, the period relative to the whole service life from the corrosion onset to the structural failure is short. Due to this fact, the corrosion initiation period has been often utilized to measure the service life of RC structures [43,44]. A schematic representation of the conceptual model of corrosion stages of reinforcement bar is presented in Figure 2.2.

The premature failures due to corrosion of reinforcement bar in RC structures are among the major challenges of civil infrastructures which causes huge economic losses. For instance, the annual total direct cost of chloride-induced corrosion in US highway bridges alone exceeds eight billion USD. The indirect costs caused by traffic delays and lost productivity are projected to be ten times more than the cost of corrosion associated maintenance, repair and rehabilitation [45,46]. Even if chloride-induced corrosion is normally more pernicious and more expensive to repair, carbonation-induced corrosion of reinforcement bar affects a wider range of RC structures at a larger scale. Thus, it is a critical problem in several parts of the world and presently two-thirds of RC structures are subjected to environmental situations that favour carbonation-induced corrosion [35,47] .

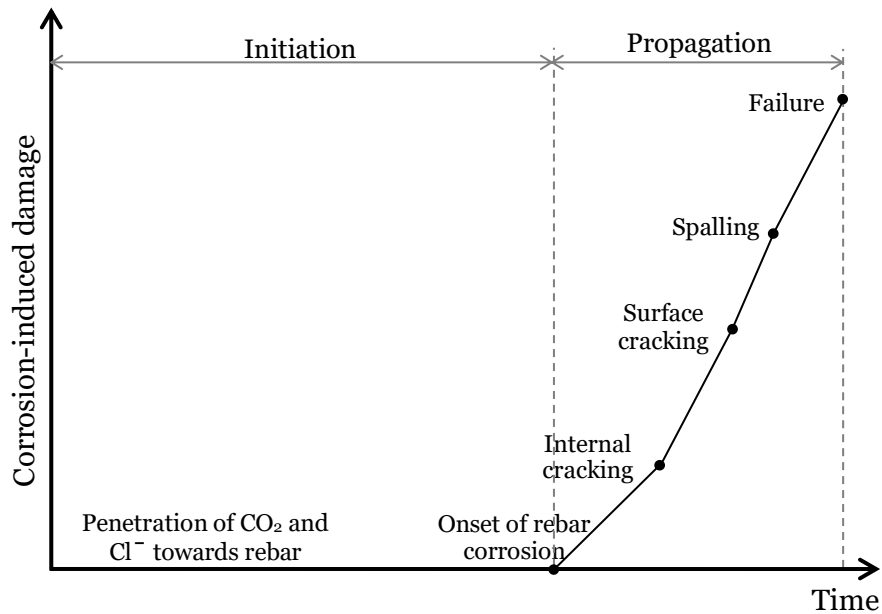

**Figure 2.2.** Corrosion of rebar in concrete structure: initiation and propagation periods (Publication I).

### 2.2.2 Hygrothermal behaviour

The amount of moisture within the concrete is the major factor that controls the corrosion rate through their influence on the electrochemical reactions at the reinforcement bar-concrete interface and ions transport between anodes and cathodes [11]. The surrounding temperature also governs the corrosion rate of reinforcement bar since it influences the electrochemical reactions and the amount of the moisture that the concrete retains [12]. For instance, the corrosion rate varies by more than a factor of ten in a regular seasonal temperature range of 5 to 30 °C [2,12,48]. Uncontrolled hygrothermal (moisture and temperature) can also cause other types of deterioration on concrete. In the presence of high moisture, some aggressive substances from internal or external sources can react with concrete ingredients leading to concrete damage [9,10,24,28]. For example, alkali reaction may take place when alkalis react with aggregates to form products that are deleterious to concrete. In low temperature, concrete may be damaged by freezing and thawing if the concrete pore system is filled with moisture and has reached a critical degree of saturation [28,49]. Hence, controlling the hygrothermal condition of RC structures is essential to prolong their service life.

The existence of a large number of RC structures that are subjected to corrosion of reinforcement bar and other deterioration mechanisms caused by uncontrolled hygrothermal interaction call for cost-effective maintenance measures. In the past few decades, substantial efforts have been put into devising economical methods to control the moisture penetration into RC structures. European Standard - EN 1504 proposes surface-protection systems to limit the amount of moisture content, and thus control the corrosion rate of reinforcement bar by increasing the concrete resistivity under rehabilitation principles P2 – *Moisture control of concrete* and P8 – *Increase of the electric resistivity of concrete* [21]. According to EN 1504, the surface-protection systems that can be applied for concrete are categorized into three groups: i) hydrophobic impregnation: produces a water-repellent surface with no pores filling effect; ii) impregnation: lessens the surface porosity with partial or full pores filling effect; and iii) coatings: forms a continuous protective film on the concrete surface.

The chemical compositions of commonly applied surface-protection materials to control the moisture penetration into concrete vary broadly. Due to this, the surface-coating systems may behave differently and even lead to unintentional damage to the concrete. It may also deliver dissimilar degrees of protection against moisture even for those surface-protection materials with similar generic chemical composition. All these facts make selection of suitable surface-protection system for a particular structure challenging.

In order to capture the hygrothermal behaviour inside surface-protected concrete members, details of the temporal change of properties of the applied coating materials under environmental and service conditions are necessary. Furthermore, a comprehensive understanding of the interaction of different surface-protection materials with the substrate concrete is indispensable. Some surface-treating materials penetrate into the pores of concrete and react with the hydration products of the concrete, but some other materials form a continuous layer at the surface of concrete. All these conditions create complexity on the hygrothermal behaviour assessment of surface-protected concrete.

## 2.3 Conventional corrosion assessment models

As discussed above, carbonation and chloride ion penetration into concrete cause initiation of corrosion of reinforcement bar. Reasonably accurate prediction of the depth of carbonation and concentration of chloride ions is crucial to optimize the design and maintenance programs for RC structures. In the past three decades, considerable attempts have been performed to develop durability models for RC structures subjected to environmental situations that favour carbonation- and chloride-induced corrosion. As a result, diverse models and input parameters have been established. The complexity degree of the proposed models differ from straightforward analytical models presuming uniaxial diffusion into concrete, to more complex numerical models which take the physical, chemical, and electrochemical processes of gases and ions transport into consideration [50–52]. Some of the adopted analytical models and the associated value of the input parameters have been suboptimal, incomplete, and/or unsuitable for the prevailing circumstances. Due to these facts, the prediction outcomes of different models vary considerably even for concrete elements with the same mix proportions that are exposed to identical environmental conditions [13]. Though the complex scientific models yield rationally precise predictions, they lack user friendliness and require well-skilled professionals, making them appropriate only for research but not for practical applications. In practice, durability models in the form of simple analytical equations on the basis of Fick's second law of diffusion are widely applied to predict carbonation depth and chloride concentration in concrete.

Conventionally, the depth of carbonation is evaluated using a simplified version of Fick's second law of diffusion, Equation (2.6) [28,31,34,53,54]. This model obeys the square root law and is utilized to predict the depassivation time by extrapolating the carbonation depth measured at a certain time to the future.

$$x_c(t) = k\sqrt{t}, \tag{2.6}$$

where $x_c(t)$ is carbonation depth at the time $t$ [mm], $k$ is coefficient of carbonation [mm/year$^{0.5}$] and expressed as $\sqrt{\frac{2 \cdot D_{CO_2}(C_1 - C_2)}{a}}$, where $D_{CO_2}$ is diffusion coefficient for $CO_2$ through carbonated concrete [mm²/year], $C_1$ is concentration of $CO_2$ for the surrounding environment [kg/m³], $C_2$ is concentration of $CO_2$ at the carbonation front [kg/m³], $a$ is mass of $CO_2$ per unit volume of concrete required to carbonate all the available calcium hydroxide [kg/m³], and $t$ is the time of exposure to the atmosphere containing $CO_2$ [year].

The assumptions in Equation (2.6) are: i) diffusion coefficient for $CO_2$ through carbonated concrete is constant; ii) the amount of $CO_2$ required to neutralize alkalinity within a unit volume of concrete is invariant; and iii) $CO_2$ concentration varies linearly between fixed boundary values of $C_1$ at the external surface and $C_2$ at the carbonation front. To evaluate $k$, the carbonation depth of concrete should be determined in advance. It can be examined using concrete cores taken from existing structures or by performing an accelerated test in laboratory. Indeed, carbonation depth is often evaluated by carrying out an accelerated test using higher $CO_2$ concentration in a controlled environment since carbonation is a slow process [55]. Then using the value of the laboratory measured depth of carbonation, the amount of the equivalent $k$ and thus the depassivation time of the reinforcement bar is computed. This is a common approach even if the accelerated test may not precisely explain the natural carbonation process consistently [53]. Equation (2.6) is plausible as far as the three assumptions are fulfilled. However, the assumptions are invalid in reality. For instance, diffusion coefficient of $CO_2$ varies both temporally and spatially. The reason for these variability is that the diffusion of $CO_2$ in concrete depends on multiple factors, including concrete mix composition, curing conditions and the macro- and micro-environment to which the concrete is exposed [31,55,56]. Due to this, Equation (2.6) often fails to represent the actual condition of the concrete structures, leading to inaccurate carbonation depth prediction [28,56,57]. To minimize some of the uncertainties, analytical models which consider direct account of some of the carbonation controlling parameters have been proposed. For example, the model proposed in fib-MC2010 [58] and DuraCrete framework [59], that is given below in Equation (2.7).

$$x_c(t) = \sqrt{2 \cdot k_e \cdot k_c \cdot R_{NAC,0}^{-1} \cdot C_a \cdot W(t)} \cdot \sqrt{t}, \qquad (2.7)$$

where $x_c(t)$ is carbonation depth at the time $t$ [mm], $t$ is exposure time [year], $k_e$ is environmental function [-], $k_c$ is execution transfer parameter [-], $C_a$ is $CO_2$ concentration in the air [kg/m³], $W(t)$ is weather function [-], $R_{NAC,0}^{-1}$ is inverse effective carbonation resistance of concrete [(mm²/year)/(kg/m³)] which is determined at a certain time $t_0$ using the natural carbonation test.

As observed in Equation (2.7), the fib and DuraCrete model adopt Equation (2.6) by linking the carbonation coefficient with parameters of the concrete property and the environment condition. There are also other models which follow the same principle as Equation (2.7). Summarized list of those models can be found in [60]. The majority of the models incorporate limited carbonation controlling parameters. The linked parameters of these models, such as exposure condition, water-to-cement ratio (w/c), and compressive strength, have ordinarily been considered as random variables. Though air permeability of concrete relies largely on the w/c, it is also governed by other parameters, e.g. mineral admixtures [61]. The combination of several assumptions and simplifications in the prevailing carbonation prediction models lead to a considerable uncertainty in their performance.

The chloride ion concentration in concrete is often estimated by adopting a simplified Fick's second law of diffusion based analytical formula described by Equation (2.8) [62].

$$C_x = C_i + (C_s - C_i)\left(1 - erf_{(x)}\left[\frac{x}{2\sqrt{D_{nss}t}}\right]\right), \tag{2.8}$$

where $C_x$ is the Cl¯ content measured at average depth $x$ [m] after exposure time $t$ [s][% by mass of concrete], $C_s$ is the Cl¯ content at the exposed surface [% by mass of concrete], $C_i$ is the initial Cl¯ content [% by mass of concrete], $D_{nss}$ is the diffusion coefficient of Cl¯ at nonsteady state [m²/s], and $erf_{(x)}$ is the error function [-].

The foremost limitations of Equation (2.8) are [15,36,63,64]: i) the surface chloride content is invariant; ii) the nonsteady diffusion coefficient remains constant; and iii) the $D_{nss}$ is assumed to be uninfluenced by different $C_s$. In real situation, $C_s$ and $D_{nss}$ cannot be recognized as constants. This is due to the fact that the transport properties of Cl¯ depend on the amount of Cl¯ concentration in the pore solution and the intrinsic permeability of the concrete. The Cl¯ concentration amount varies due to the continuous chemical reaction of Cl¯ with the dilute cement solution and the amount of diffused Cl¯. The concrete permeability property also varies during the cement hydration process with time. In another perspective, the change of the pore structure of concrete is governed by cement type, w/b, exposure time, type of admixtures, and exposure conditions. Due to these, both $C_s$ and $D_{nss}$ are temporally and spatially varying parameters [65,66]. It is also comprehended that the Cl¯ is accumulated in the pore solution of concrete during chloride diffusion process. As the amount of Cl¯ concentration increases, the mobility of free Cl¯ slowly becomes weak, and thus lessens the magnitude of $D_{nss}$. This demonstrates that $D_{nss}$ is a function of $C_s$ and this makes the assumption (iii) in Equation (2.8) invalid. Moreover, in Equation (2.8), the error function equation takes only diffusion mechanism into consideration.

Nevertheless, the ingression of Cl⁻ into concrete involves a complex chemical and physical process that integrates diverse transport mechanisms besides diffusion, such as capillary sorption, and permeation. All these facts describe the reason behind the failure of Equation (2.8) to offer accurate predictions in several cases [64]. In fact, in order to undertake the time dependency of $D_{nss}$ and the impact of other influential parameters some approaches have been suggested, e.g. fib-MC2010 [58] and DuraCrete framework [59]. Equation (2.9) is the most applied formula for estimating $D_{nss}$, but it also fails to eliminate the uncertainty fully because the input parameters exhibit substantial scatter.

$$D_{nss}(t) = k_e . k_t . k_c . D_0 . \left(\frac{t_0}{t}\right)^n, \tag{2.9}$$

where $k_e$ is environmental function [-], $k_t$ is test method function [-], $k_c$ is curing function [-], $D_0$ is experimentally determined chloride diffusion coefficient at time $t_0$ [m²/s], $t_0$ is age of concrete at $D_0$ is measured [year], $t$ is the exposure duration [year], and $n$ is the age factor describing the time dependency of the diffusion coefficient [-].

The age factor in Equation (2.9) explains the time dependency of the diffusion coefficient based on the concrete mix composition. The magnitude of the age factor is often determined based on different concrete specimens subjected to various environments for relatively short period of time and reveals significant scatter. Several studies demonstrated that the age factor is the most sensitive parameter in Equation (2.9) [14,36,67,68]. A minimal change in its magnitude leads to a considerable uncertainty in the prediction of chloride concentration. The combination of all the above discussed assumptions causes considerable uncertainty in the prediction of Cl⁻ concentration in concrete which ultimately affects estimation of the time to onset corrosion of reinforcement bar or evaluation of the service life of the structure [15,69].

As elaborated above, the ingression rate of the aggressive substances ($CO_2$ and Cl⁻) into the concrete pores is predominantly a function of concrete properties and environmental circumstances. In a given structure, the penetration rate of these substances cannot be constant and even they may alter in various parts of the structure. The presented corrosion assessment methods in the form of analytical equations are no better than their underlying conceptual base. So, the estimation of corrosion onset utilizing Fick's law based analytical models is uncertain. In addition, the rapidly increasing use of combinations of supplementary cementitious materials and new technologies are another factors which make the traditional models incapable to precisely evaluate the corrosion initiation time [69–72]. Hence, developing novel methods that estimate the carbonation depth and chloride profile accurately are crucial to mitigate premature failure of RC structures caused by corrosion of reinforcement bar and the related costs.

The complexity of several interacting parameters which control the initiation of corrosion of reinforcement bar in concrete calls for researchers to question *how to eliminate or mitigate the uncertainties observed in the traditional corrosion assessment methods?* This is one of the research questions of this dissertation, research question one. Indeed, the oversimplified corrosion assessment methods can be integrated with a semi-probabilistic uncertainty model to enhance the accuracy as in the DuraCrete framework. Nonetheless, this approach cannot eliminate the associated uncertainty completely. Uncertainties can be mitigated by utilizing readily available long-term data or gathering more and more relevant data, and then modelling it using machine learning methods. Such data-driven models estimate without assumptions by mapping the variables of the input to the output that closely approximate the target instances. They do also have the ability of extracting useful knowledge from the data, thus contributing to a better understanding of the complex and nonlinear interaction of multiple parameters. Fortunately, there are readily available long-term field data obtained from real structures and specimens made with several mix compositions exposed to carbonation and chloride environment. The increment of the availability of more and more field data exposed to different exposure conditions is evident since nowadays real-time monitoring of several corrosion controlling parameters in concrete structure is achievable.

In another perspective, the broadly utilized Fick's law based carbonation depth and chloride profile prediction models rely on limited number of parameters. Certainly, exploration of $CO_2$ and $Cl^-$ transport in concrete is conducted for several years to acquire a better understanding of the influential predictors. In order to recognize the effect of different parameters, a large number of experiments must be carried out since the concrete microstructure is immensely complex and its transport properties are controlled by several interacting factors. However, it is generally challenging to isolate the effects of particular parameters since other governing parameters also vary naturally at the same time [16,17]. Hence, determining parameters that characterize the carbonation process and the chloride penetration in concrete as well as their interdependency is crucial in order to develop parsimonious and precise model. But identifying the significance of the parameters using the traditional statistical methods is impossible since the carbonation process and the chloride penetration in concrete is a complicated process governed by several nonlinear factors. All the limitations presented above can be mitigated by adopting various types of machine learning methods since they are capable of handling highly nonlinear variables with multiple interactions.

Unlike carbonation and chloride penetration assessment, hygrothermal transport phenomena through concrete are well understood and numerical models that can be used in practice have already been developed. A comprehensive review of hygrothermal simulation models is presented in [22]. Although hygrothermal prediction models for concrete are available, it is challenging to include the ever increasing of surface-protection material types and their application techniques in the models' library. It is a known fact that numerical models can yield accurate predictions of any process if and only if the actual material properties are well studied and utilized. In addition, hygrothermal interaction in surface-protected concrete elements involves multiple temporally varying complex interactions. Such a complex problem needs an approach where the most important features with the involved multiple interactions are modelled so that the behaviour of the system could be reasonably predicted. These features can be gathered through long-term in-service monitoring using appropriate sensors and predicting the hygrothermal behaviour from the gathered data using machine learning techniques is an attractive alternative. Furthermore, the gathered and the predicted data, with the help of exploratory data analysis technique, could assist to evaluate the performance of the applied protection materials while obtaining valid information regarding the real hygrothermal behaviour of the concrete.

## 2.4 Machine learning

Machine learning is a major subfield of artificial intelligence that deals with the design and implementation of algorithms to recognize complex patterns from data and make intelligent decisions [73–81]. Machine learning based models can be predictive to carryout prediction or descriptive to discover knowledge from data, or both without assuming a predetermined equation as a model [77,82,83]. Even if machine learning grew out of the quest for artificial intelligence, its scope and potential are more generic. It draws upon ideas from a diverse set of disciplines, including Probability and Statistics, Information Theory, Psychology and Neurobiology, Computational Complexity, Control Theory and Philosophy [80].

The design process of a machine learning model involves a number of choices, including the learning types, the target performance function to be learned, a representation of the target function and an algorithm (a sequence of instructions used for learning the target function). Based on the training conditions, machine learning is classified as supervised, unsupervised, semi-supervised and reinforcement learning [77,84]. The supervised and unsupervised learning are the most commonly implemented learning types in several area of applications [83].

*Supervised learning*: in this learning type, the training dataset comprises pairs of $N$ input instance $x$ and a desired output (target) value $y$: $\{x_i, y_i\}_{i=1}^{N}$. A supervised learning algorithm analyses the training data and generates an inferred function, which can be applied for mapping new instances. A best scenario will allow the algorithm to accurately estimate the target for unseen input instances. The process of learning from the training dataset can be thought of as a teacher supervising the learning process, hence the name *supervised learning*. The algorithm iteratively performs predictions on the training data and is rectified by the teacher. The learning process halts when the algorithm attains an acceptable performance level. Based on the nature of the target variable, supervised learning problem is categorized into two: *classification* and *regression*. Supervised learning problems where the target variable is defined as a finite set of discrete values are called *classification* whereas those in which the value of the target variable is continuous are referred to as *regression*.

*Unsupervised learning*: in this learning type, the training dataset consists of only $N$ input instances $\{x_i\}_{i=1}^{N}$ and no corresponding target values. The goal of unsupervised learning is inferring a function to describe hidden structure from the unlabelled data. The most common unsupervised learning problem is clustering where the goal is to partition the training instances into subsets (clusters) so that the data in each cluster display a high level of proximity. Unlike supervised learning, there is no a teacher providing supervision as to how the instances should be handled. Also, there is no evaluation of the correctness of the structure that is output provided by the adopted relevant algorithm.

In order to perform predictions and/or discover new useful knowledge from data using machine learning methods, an algorithm that is capable of learning the target function from training data is required. The algorithms in machine learning implement different types of methods from various fields, such as, pattern recognition, data mining, statistics, and signal processing. This allows machine learning to undertake the synergy benefits from all these disciplines, and thus results in robust solutions that integrates various knowledge domains [81]. Some of the commonly adopted powerful algorithms that have been applied in supervised and unsupervised learning types are demonstrated in Figure 2.3. It can be observed that some algorithms operate under both supervised and unsupervised learning types to solve several diverse problems.

Supervised learning type is the focus of this dissertation since the research questions two, three and four deal with regression problems. Typically, regression types of problems are handled by developing a functional model which is the best predictor of $y$ given input $x$ employing a particular training data $D = \{y_i, x_i\}_1^{N}$ as expressed in Equation (2.10).

$$y = \hat{F}(x_1, x_2, \ldots, x_n) = \hat{F}(X), \tag{2.10}$$

where $y_i$ is the output variable, $x_i$ is the input vector made of all the variable values for the $i^{th}$ observation, $n$ is the number of variables, and $N$ is the number of instances.

A typical roadmap for building machine learning models is illustrated in Figure 2.4. It consists of the following three major steps:

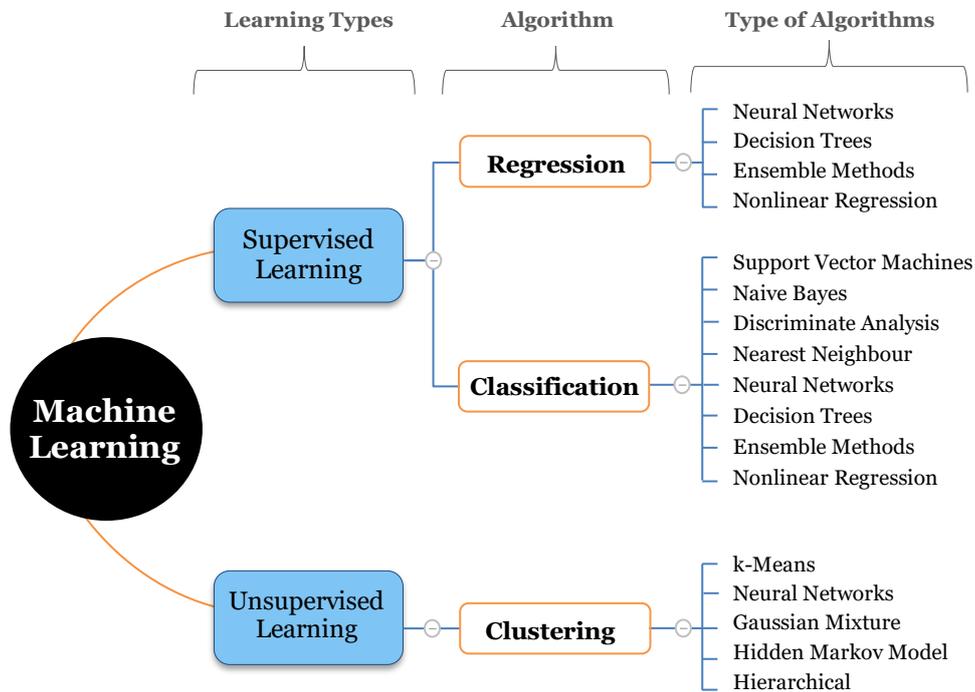

**Figure 2.3.** Commonly adopted machine learning algorithms (Publication I).

1. *Data preprocessing*: this step is the most critical one in machine learning applications since raw data usually comes in the form that is unsuited for the optimal performance of a learning algorithm. Due to this, the first step in data preprocessing activity is data cleaning. This activity includes replacing missed data, removing outliers, and smoothing noisy data. Data cleaning is often followed by integration of multiple data sources and data transformation to a specific range (normalization) and dimension reduction for optimal performance. Removing redundant features by compressing the features onto a lower dimensional subspace while holding most of the relevant information is also essential to make the learning process faster. In addition, in data preprocessing phase, the data is randomly divided into training and test set. The training set is applied to train and optimize the machine learning model, while the test set is used to assess the performance of the final model.

2. *Training and selecting a predictive model:* as presented earlier, there are a wide range of machine learning algorithms that have been developed to solve different problems. Each algorithm has different feature and the choice mainly depends on the types of the problem to be resolved and the available data. So, in this phase, at least a handful of different algorithms shall be trained in order to select the best performing one (a model that fitted well on the training dataset). The most commonly performance evaluating metric is the mean-square error (the mean of the squared difference between the target and its predicted value).
3. *Evaluating models and predicting unseen data instances:* the last step is dedicated to model assessment. The main issue that machine learning models face is how well they model the underlying data. The model can be too specific if it memorized the training dataset and unable to generalize to unseen examples. This means that the model *overfits* the data employed for training. The model can also be too generic if it is incapable to capture the relationship between the input instances and the target values. In another word, it *underfits* the training data. Both *overfitted and underfitted* models are useless for making valid predictions. Hence, the purpose of this step is to evaluate how well the model has generalized to new (unseen) data. To do so, the test dataset is applied. If the model performs well in the test data, it can be applied to predict new future data since it generalized the training dataset.

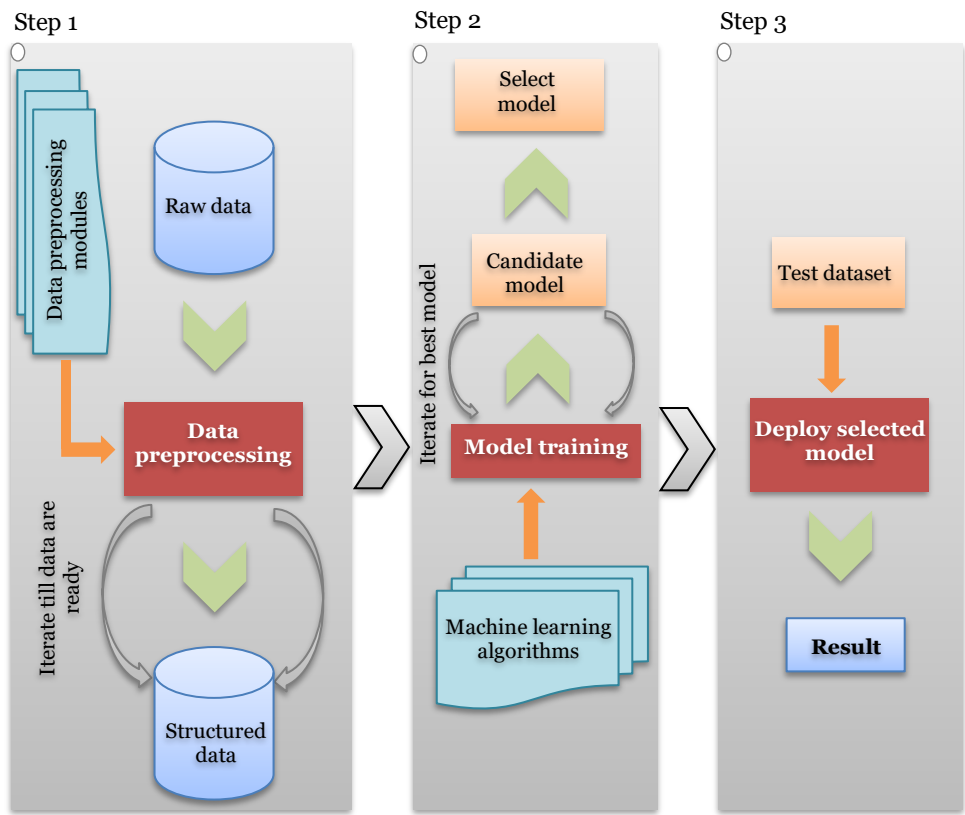

**Figure 2.4.** The fundamental steps of a typical machine learning workflow.

Machine learning is one of the most significant technological developments in recent history. There are a wide spectrum of successful practical applications of machine learning in different domains, such as, computational finance [85–87], image and speech processing [88–90], property valuation [91–93], computational biology [94–96], and energy production [97–99]. Although employing machine learning is becoming a regular practice in diverse engineering fields, its application for assessing durability of concrete is yet limited.

In this dissertation, supervised learning type is employed since estimation of carbonation depth, chloride ingress and hygrothermal performance are regression problems and the input and the target data are known in the training dataset. The adopted learning algorithms are neural network, decision tree, and boosted and bagged ensemble methods. The fundamental principles of these algorithms are discussed as follows.

### 2.4.1 Neural network

Neural network is a computational network inspired by biological neural networks which comprises partially or fully interconnected simple processing units called artificial neurons [100–102]. Each neuron processes data locally using similar concepts as learning in the brain. Neural network is ideal for supervised learning since the connections within the network can be systematically adjusted based on inputs and outputs. They are usually categorized based on their architecture (pattern of connections between the neurons) and the architecture is intimately connected with the learning algorithm utilized to train the network. The common classification are: single-layer feedforward, multilayer feedforward, and recurrent networks [100]. Multilayer feedforward and recurrent networks were adopted in Publications II and IV to solve the research questions two and four, respectively. In this section, only the details of the utilized types of neural networks are discussed.

Multilayer feedforward architecture along with backpropagation training technique is broadly employed to solve complex nonlinear regression problems [103–106] and this approach is adopted in Publication II of the dissertation. This architecture often has three or more layers: input, hidden, and output layers. Figure 2.5 shows the architecture of a multilayer feedforward neural network (multilayer perceptron) with a single hidden layer. The first and the last layers are known as input and output layers, respectively. The intermediate layer is called hidden layer which assists to execute the necessary computations before conveying the input data to the output layer. This network can be seen as a nonlinear parametric function from a set of inputs, $x_i$, to a set of outputs, $y_m$. First, linear combinations of the weighted inputs are formed. This includes the additional external inputs provided to the network, which is known as bias, the neurons represented by blue colour in Figure 2.5. Biases have no effect on the performance of the network, but they increase the flexibility of the network to fit the data. After linear combinations, they are translated to new values using an activation function $\varphi(.)$, Equation (2.11) [105–107]. Then, the outputs of the first-hidden layer neurons are multiplied by the interconnection weights of the layer that connect them to neurons of the next layer as expressed by Equation (2.12). If the network has multiple hidden neurons, this activity continues until the output neurons compute the values of the output.

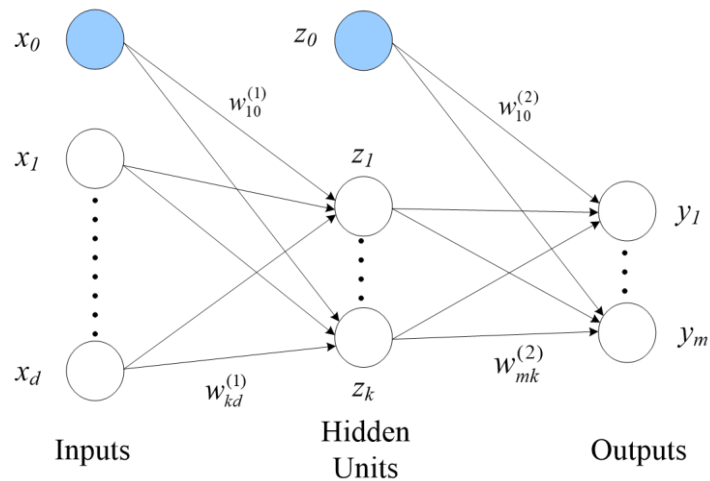

**Figure 2.5.** A multilayer feedforward neural network with a single hidden layer (Publication II).

$$z_j = \varphi\left(\sum_i w_{ji}^{(1)} x_i\right), \tag{2.11}$$

$$y_m = \sum_j w_{mj}^{(2)} z_j, \tag{2.12}$$

where $w_{ji}^{(1)}$ and $w_{mj}^{(2)}$ are the weights of the network which are initially set to random values, and then adjusted during training by backpropagation using the response data.

The purpose of the activation function is to control the amplitude of the output of a neuron in terms of the induced local field $v$ and break the linearity of a neural network, allowing it to learn more complex functions. Depending on the characteristics of the problems, various forms of activation functions can be defined [107]. The common ones are *linear*, *logistic* and *hyperbolic tangent* activation functions, which are expressed by Equations (2.13) to (2.15) and illustrated in Figure 2.6. The vertical and the horizontal axes represent the unit's output and its input, respectively. It can be noticed that the *hyperbolic tangent* function has the same form as the *logistic*. However, the function *hyperbolic tangent* covers the range from [-1 1] whereas the *logistic* function covers from [0 1]. The *linear* function simply outputs a value proportional to the summed inputs. To build nonlinear models, the activation function of the network must be nonlinear and a single type of activation function is applied for neurons in the same layer.

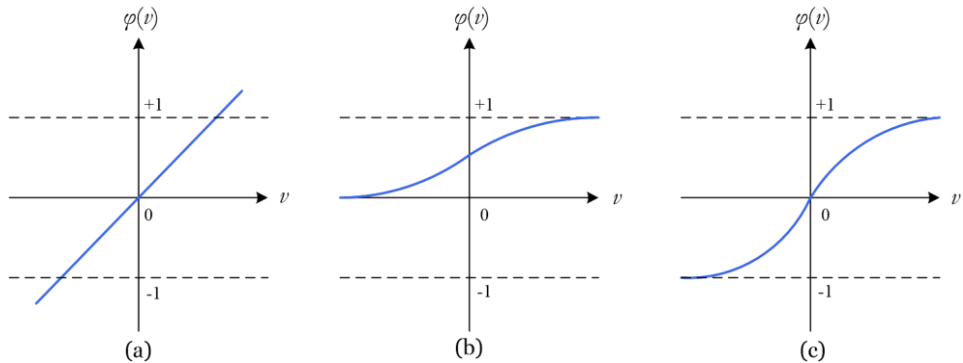

**Figure 2.6.** Activation functions (a) linear, (b) logistic, and (c) hyperbolic tangent.

Linear function: $$\varphi(v) = v. \tag{2.13}$$

Logistic function: $$\varphi(v) = \frac{1}{1+e^{-av}}. \tag{2.14}$$

Hyperbolic tangent function: $$(v) = \frac{e^{2v}-1}{e^{2v}+1}. \tag{2.15}$$

Recurrent neural network differentiates itself from a feedforward neural network by having arbitrary feedback connections, including neurons with self-feedback [100]. Self-feedback refers to a circumstance where the output of a neuron is fed back into its own input. The architectural layout of recurrent network takes many different forms depending on the kinds of time-series problems. The adopted subclass of recurrent network in Publication IV is called nonlinear autoregressive with exogenous inputs (NARX). It is one of the popular network and has high capability in capturing long-term dependencies since it uses the feedback derived from the output at explicit time lags as part of the input data [100,101]. The general architecture of the NARX network is illustrated in Figure 2.7. The network has a single input that is applied to a tapped-delay-line memory of $q$ units. It has a unique output that is fed back to the input through another tapped-delay-line memory also of $q$ units. The values of these two tapped-delay-line memories are utilized to feed the input layer of the multilayer perceptron. The present input value of the network is represented by $u(n)$, and the respective output value of the network is expressed by $\hat{y}(n+1)$. This mean that the output is ahead of the fed-back input by one-time unit. So, the data window supplied to the input layer of the multilayer perceptron can be denoted as follows:

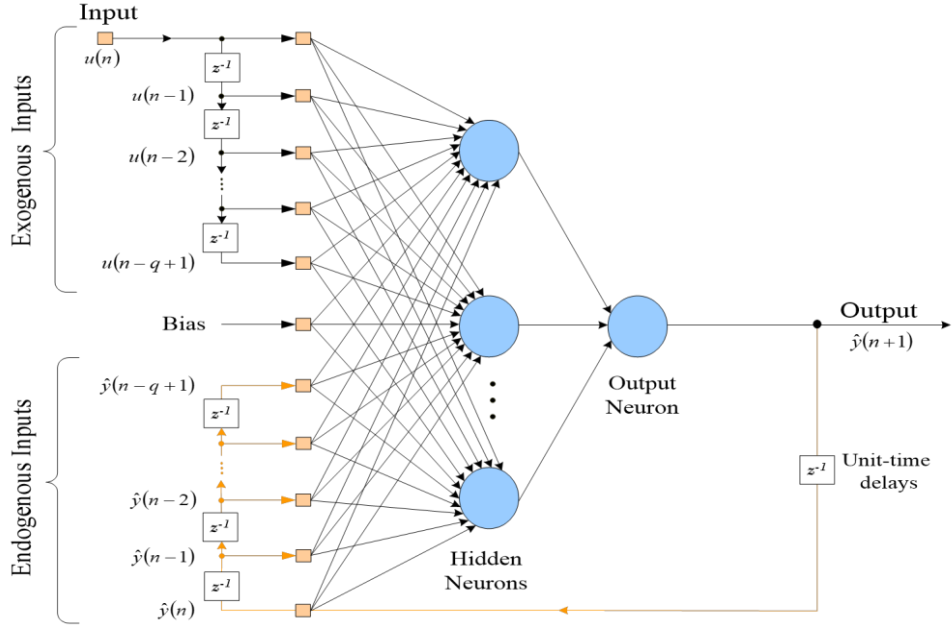

**Figure 2.7.** NARX network with *q* delayed inputs (Publication IV).

- Present and past input values, $u(n)$, $u(n-1)$,..., $u(n-q+1)$, which denote exogenous inputs. This input values comes from other sources (outside the network), and
- Delayed output values, $\hat{y}(n)$, $\hat{y}(n-1)$,..., $\hat{y}(n-q+1)$, on which the network output $\hat{y}(n+1)$ is regressed and represents the value of the endogenous variables.

The dynamic behaviour of the NARX network is expressed by Equation (2.16).

$$\hat{y}(n+1) = F\big(\hat{y}(n), \ldots, \hat{y}(n-q+1); u(n), \ldots, u(n-q+1)\big). \qquad (2.16)$$

### 2.4.2 Decision tree

Decision tree is a nonparametric hierarchical data structure which implements the divide-and-conquer strategy. Decision tree model comprises nodes, branches, and leaves. Nodes correspond to domain regions that need to be decompounded into smaller regions by splitting. Leaves represent domain regions where additional splits cannot be implemented. Branches connect to descendant nodes or leaves in proportion to specific split outcomes. Splits are determined by some relational circumstances on the basis of the selected instances that may have two or more outcomes. A split can be formally characterized by a test function $t : X \rightarrow R_t$ that maps instances into split outcomes. A separate outgoing branch is associated with each possible outcome of a node's split. The relationship between the parent node and its descendant nodes, theoretically characterized by the branches linking the former to the latter, does not often have to be unambiguously delineated in the data structure of the decision tree. If a split's outcome can be explicitly ascertained for any attainable instance, then it does partition the domain into disjoint subsets, in proportion to the outgoing branches. Hence, it is easy to realize that each node $n$ of a decision tree complies with a region (subset) of the domain (Equation 2.17) determined by the sequence of splits $t_1, t_2, ..., t_k$ and their outcomes $r_1, r_2, ..., r_k$ occurring on the path from the root to the node [108].

$$X_n = \{x \in X | t_1(x) = r_1 \wedge t_2(x) = r_2 \wedge \cdots \wedge t_k(x) = r_k\}. \tag{2.17}$$

Decision tree is fast learner with high degree of interpretability and handles complex nonlinear problems with a large number of observations and input variables by reducing them into manageable levels and recursively applies the same approach to the sub problems [109–111]. The power of this procedure arises from the potential to divide the instance space into subspaces and each subspace is fitted with varied models [77,110]. A decision tree that is applied for examining regression problems can be referred as a regression tree. The basic structure of a regression tree is shown in Figure 2.8. The left subfigure denotes the data points and their partitions while the right subfigure illustrates structure of the corresponding regression tree. As it can be observed from Figure 2.8, regression tree is composed of decision and leaf nodes. A test function is applied at each decision node and the branches are labelled with discrete outcomes of the function. This test procedure starts at the root and recursively carried out until leaf nodes are found. The value at the leaf node is the output [77]. Regression tree is one of the integrated learning

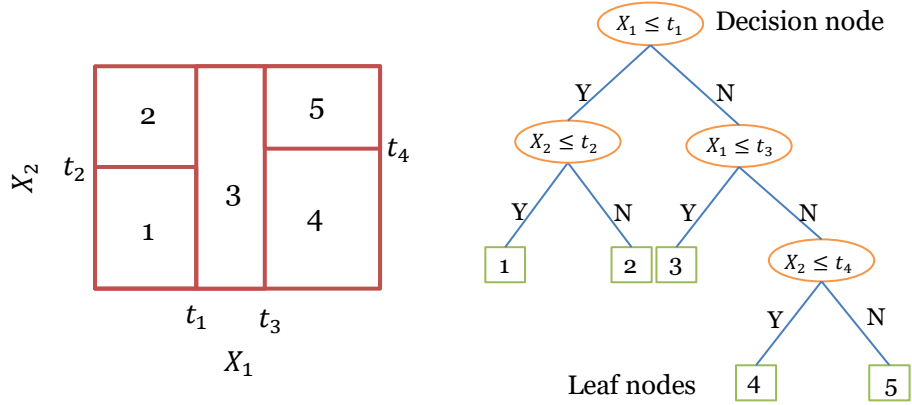

**Figure 2.8.** Example of a dataset and its corresponding regression tree (Publication III).

algorithms that are developed in Publication II to predict carbonation depth.

### 2.4.3 Ensemble method

The idea of an ensemble method is to build a powerful predictive model by aggregating multiple machine learning models, each of which solves the same original task [76,77,109,112,113]. Normally, generating multiple models utilizing different datasets without integrating them into an ensemble and directly choosing one of them that performs best does not deliver a functional solution. Ensemble methods are applicable to the two foremost predictive modelling tasks, classification and regression. In both cases, ensemble method provides significant improvement over a single model at the expense of investing more computation time due to building multiple models. To utilize this potential for superior predictive power, suitable techniques for building of base models (models used as inputs for ensemble methods) and aggregation are needed [108]. The base models often generated using machine learning algorithms. Model aggregation comprises integration of base models $m_1, m_2, \ldots, m_k$ into an ensemble model $\acute{M}$ by building a prediction combination strategy that compute $\acute{M}(x)$ based on $m_1(x), m_2(x), \ldots, m_k(x)$ for arbitrary $x \in X$. The amalgamated model $\acute{M}$ is represented by all of its base models and the strategy applied for integrating their predictions. Depending on the types of base models as well as the applied integration methods, various ensemble methods can be formed.

Though there are several ensemble models in the machine learning literature, there are two models that use decision tree learners as a base model and have proven to be effective on solving regression problems for a wide range of datasets. These ensemble models are: bagging and boosting decision trees [76,90,108]. Both types of models are integrated in the developed carbonation depth prediction model presented in Publication II of the dissertation. Boosting decision tree is adopted in Publication III to answer the research question three.

*Bagging decision tree*

In bagging decision tree, the base models are created using the training datasets. Multiple randomly drawn bootstrapped samples from the original data form the training datasets. This procedure is performed a number of times until a large subset of training datasets are formed and the same samples can be extracted more than once. On average, every formed bootstrapped training dataset hold $N\left(1-\frac{1}{e}\right) \approx 0.63N$ instances, where $N$ is the total number of samples in the original training dataset. The left-out instances in the training dataset are known as out-of-bag observations. The final output of the bagging ensemble model is the average of the projected output of the individual base models, thereby reducing its variance (tendency to learn random things) and provide higher stability [77,90,109,114]. The performance of the ensemble model is evaluated using the out-of-bag observations. In bagging decision tree, the base model fits the training data $D = \{(x_1, y_1), (x_2, y_2), \dots, (x_N, y_N)\}$, obtaining the tree's prediction $\hat{f}(x)$ at input vector $x$. Bagging averages this prediction over a wide array of bootstrap samples. For each bootstrap sample $D^{*t}, t = 1,2, \dots, T$, the model provides prediction $\hat{f}^{*t}(x)$. The bagged estimate is the mean prediction at $x$ from $T$ trees as described by Equation (2.18).

$$\hat{f}_{bag}(x) = \frac{1}{T}\sum_{t=1}^{T} \hat{f}^{*t}(x). \tag{2.18}$$

Bagging decision tree, with several base models, provides better prediction over a single model. Unlike the base models, overfitting will not occur in this ensemble method since the integration of the base models cancel it out effectively. Bagging decision tree often builds deep trees. Due to this bagging decision tree is time consuming and memory intensive, which in turn leads to slow predictions.

*Boosting decision tree*

Boosting can be described as an improvement of bagging that comprises multiple base models by shifting the focus toward instances that experience difficulty in prediction [90,108,114]. The shift of focus is mainly addressed by instance weighting. In contrast to bagging, boosting decision tree build simple tree models in a serial manner with advancement from one tree model to the other and combining them to boost the model accuracy. Each tree is grown from a training dataset $D = \{(x_1, y_1), (x_2, y_2), \ldots, (x_N, y_N)\}$, using information from previously grown trees. A relevant algorithm can be applied to fit the sample training datasets $D^{(t)}, t = 1,2,\ldots,T$, utilizing a sequence of varying weights $w^{(1)}, w^{(2)}, \ldots, w^{(T)}$, yielding the tree's predictions $\hat{f}^{(1)}(x), \hat{f}^{(2)}(x), \ldots, \hat{f}^{(T)}(x)$ for each input vector $x$ and their corresponding weight vector $w$. The weight vector is normally started by applying an initial weight $w^{(1)}$ and continually adjusted in every generated base model depending on observed residuals. The weight is increased for observations in which the base model yields high residuals (poor predictions) and decreased for cases in which the model provides low residuals (good predictions). The output of the boosted decision tree can be expressed by Equation (2.19), where $\{\propto_t\}_{t=1}^{T}$ are the linear combination coefficients.

$$\hat{f}_{boost}(x) = \sum_{t=1}^{T} \propto_t \hat{f}^{(t)}(x). \tag{2.19}$$

Various types of strategies for adjustment instance weight and model weighting can be applied to create boosting ensemble method. For instance, it is possible to apply residuals of previous models as target function values for generation of subsequent base model rather than instance weighting. This allows the regression algorithm to counteract limitation of the previous models instead of optimizing its training performance.

In boosting decision tree, each tree model can only deliver reasonably good predictions on some instances, and therefore more and more trees are built to enhance the performance iteratively. Unlike bagging, boosted ensemble models utilize shallow trees which make them usually smaller in terms of memory, and making prediction faster.

## 2.5 Machine learning based variable importance measure

Variable importance measure allows insight into the importance of the variables employed in the training dataset. Generally, there are three major categories of variable selection methods: filter, wrapper and embedded [111,115]. Filter methods are independent of the learning algorithms and relies only on the intrinsic characteristic of the data. These methods, compared to wrapper methods, are less computationally intensive. Wrapper methods demand a prespecified learning algorithm and based on the selected algorithm the performance of each variable is used as the measure for determining the final subset of variables. These methods are computationally intensive but yield better accuracy compared to filter methods. Embedded methods include the process of variable selection as part of the model development process. These methods combine the advantages of the filter and the wrapper methods, in terms of low computational costs and an adequate accuracy.

The ensemble methods (bagging and boosting decision trees) discussed above can perform an embedded variable selection. Both bagging and boosting decision trees are applied in Publication II. Both models were used to evaluate the importance of the input variables in predicting the carbonation depth. Ensemble method based on bagging decision tree is also implemented in Publication III in order to examine the importance of variables that describe the chloride penetration into concrete.

Variable importance measure based on permutation is one of the advanced and reliable embedded selection methods. As the name indicated, the variable importance ($VI$) measure based on this method is obtained by randomly permute the $j^{th}$ predictor variable $x_j$ (on each decision tree in the ensemble) with some permutation $\varphi_j$ among the training dataset. Then evaluate the out-of-bag error on this perturbed dataset. The importance score for the $j^{th}$ variable is computed by averaging the difference in out-of-bag error before and after the permutation over all the trees. The score is normalized by the standard deviation of these differences. Variables which induce large values for this score are ranked as more critical than variables which generate small values. This process is expressed mathematically in Equations (2.20) to (2.22) [116].

Let $\bar{\beta}^{(t)}$ be the out-of-bag instance for a tree $t$, with $t \in \{1, 2, \dots, T\}$. Then the importance of variable $x_j$ in tree $t$ is described by Equation (2.20).

$$VI^{(t)}(x_j) = \frac{\sum_{i \in \bar{\beta}^{(t)}} I(y_i = \hat{y}_i^{(t)})}{|\bar{\beta}^{(t)}|} - \frac{\sum_{i \in \bar{\beta}^{(t)}} I(y_i = \hat{y}_{i,\varphi_j}^{(t)})}{|\bar{\beta}^{(t)}|}, \qquad (2.20)$$

where $\hat{y}_i^{(t)} = f^{(t)}(x_i)$ and $\hat{y}_{i,\varphi_j}^{(t)} = f^{(t)}(x_{i,\varphi_j})$ is predicted value for $i^{th}$ instance before and after permuting its value of variable $x_j$, respectively. By definition, $VI^{(t)}(x_j) = 0$, if variable $x_j$ is not in tree $t$.

The measure of variable importance for each variable is determined as the average importance over all the trees, Equation (2.21).

$$VI(x_j) = \frac{\sum_{t=1}^{T} VI^{(t)}(x_j)}{T}. \qquad (2.21)$$

The standardized variable importance is evaluated by applying Equation (2.22). As expressed in Equation (2.22), the individual importance measures $VI^{(t)}(x_j)$ are computed from $T$ samples which are extracted from the original dataset. Thus, if every individual measure of variable importance $VI^{(t)}(x_j)$ has standard deviation $\sigma$, the average importance measure from $T$ replications has standard error $\sigma/\sqrt{T}$.

$$\widetilde{VI}(x_j) = \frac{VI(x_j)}{\frac{\sigma}{\sqrt{T}}}. \qquad (2.22)$$

Permutation based variable importance measure is applied in this dissertation to answer the research question three. Another technique has also been applied to evaluate the importance of variables in predicting the carbonation depth in order to answer second part of the research question two. This is due to the fact that variable importance measure based on permutation is impracticable for boosting decision tree, which is one of the ensemble methods employed in answering research question two. Details of the implemented method for measuring variable importance in case of boosting decision tree are presented in Section 3.4.4.

# Chapter 3

# Materials and methods

In this chapter, the utilized materials and the developed methods for assessing corrosion causing and controlling factors are discussed. The materials used in Publications II, III, and IV are presented since Publication I is based on a thorough literature review. In Publication I, materials from secondary sources, which include books, conference, and journal articles are used. The purpose of this publication is to examine the capability of machine learning methods in addressing the limitations of classical corrosion assessment models. As the focus of this dissertation is on data-driven corrosion assessment methods, the materials for Publications II, III, and IV are experimental data obtained from three different case structures.

All the methods developed in this dissertation rely on machine learning approach and implemented using MATLAB programming language. All the proposed methods to answer the research questions two, three and four of the dissertation are discussed in detail. The methods are utilized for developing data-driven models to assess corrosion causing and aggravating factors. The assessment includes predictions of carbonation depth, chloride profile and hygrothermal performance. In addition, the methods are used to discover influential predictors of carbonation depth and chloride concentration. As any data-driven models, the development process of these models primarily consists of data, data preprocessing and training. The experimental data and all the involved activities associated with model development process are presented in this chapter.

## 3.1 Concrete specimens for carbonation field test

The experimental data for carbonation study were obtained from concrete specimens that were prepared for Finnish DuraInt-project. The specimens have diverse mix compositions that represent the current prevalent industrial concrete mixes in Finland. This project was carried out jointly by Aalto University and VTT Technical Research Centre of Finland. The data were based on 23 concrete specimens. All the specimens were casted in steel moulds of size 100 x 100 x 500 mm³ and demoulded after 24 hours. Then they were immersed into water for seven days and cured in a controlled environment (21 °C temperature and 60% relative humidity). The specimens were sheltered and kept on wooden racks at about the age of 28 days in Espoo, southern Finland in order to simulate as sheltered concrete structures that are exposed to natural conditions (exposure class, XC3 (moderate humidity)) as shown in Figure 3.1. The annual average $CO_2$ concentration, temperature, and relative humidity at the storage of specimens are 375 ppm, 6 °C and 79%, respectively.

The carbonation front depths of the concrete specimens, from all sides, in a freshly broken surface of 100 x 100 mm² were measured at the age of 268, 770, 1825 and 2585 days. The carbonation depths were examined by spraying a pH indicator solution of phenolphthalein. The average of the carbonation depths measured from the four sides of each concrete specimen was considered as the representative value.

Accelerated carbonation tests were also carried out at the age of 28 and 56 days for the same concrete mixes. It was executed by exposing the concrete specimens to be carbonated in a climatic control test chamber which was filled with 1% of $CO_2$ and kept in a room with RH 60% and temperature 21 °C. The carbonation depths were measured by applying 1% phenolphthalein in ethanol solution. Carbonation fronts of two groups of concrete specimens after 56 days in the accelerated carbonation chamber is shown in Figure 3.2. Surface areas with a pink colour indicate the pH is above nine and are the non-carbonated part. The carbonated part of the specimens is the area where the colour of the concrete is unchanged.

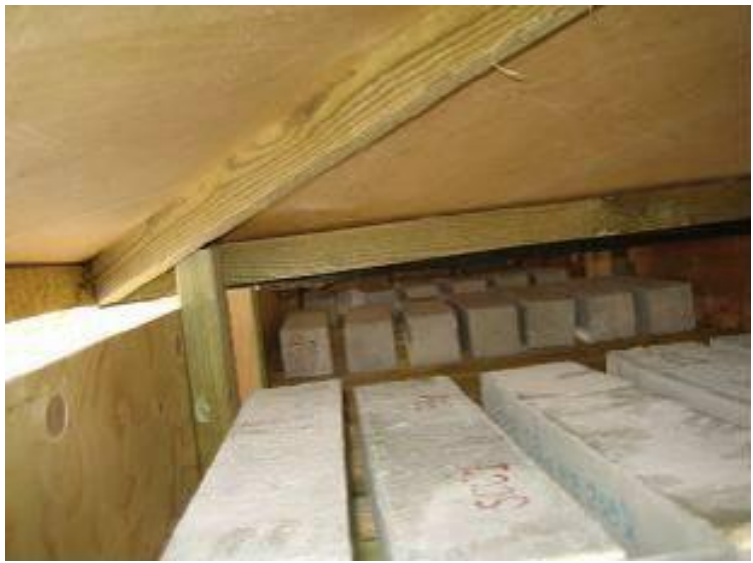

**Figure 3.1.** Sheltered concrete specimens for carbonation field test (Publication II).

### 3.2  Concrete specimens for chloride field test

The experimental data for chloride assessment were also acquired from concrete specimens of Finnish DuraInt-project. Though the data for both carbonation and chloride were obtained from the same project, the specimens for each case have dissimilar concrete mixes. The two cases were exposed to different field environments. Assessment of chloride penetration was performed using data acquired from concrete specimens with 18 dissimilar mix proportions. The specimens were casted in wooden moulds of size 300 x 300 x 500 mm³ in upright position to perform chloride test in the field. Surface treatments (impregnation, form lining, copper mortar) were applied on some of the specimens of DuraInt-project to examine their effect on chloride ingression. The concrete specimens were placed on the roadside at Kotka, Finland and Borås, Sweden. In this dissertation, the experimental data employed for chloride penetration assessment were taken from concrete specimens without surface treatment that were placed on the side of highway 7 (HW 7) at Kotka, representing exposure class XD3 (cyclic wet and dry). The reason for this is that the number of surface-treated concrete specimens with dissimilar concrete mix was few. In addition, all of them were not situated on the side of the thaw-salted road of HW 7 for long period of time. The geographical location of Kotka is illustrated in the map in Figure 3.3.

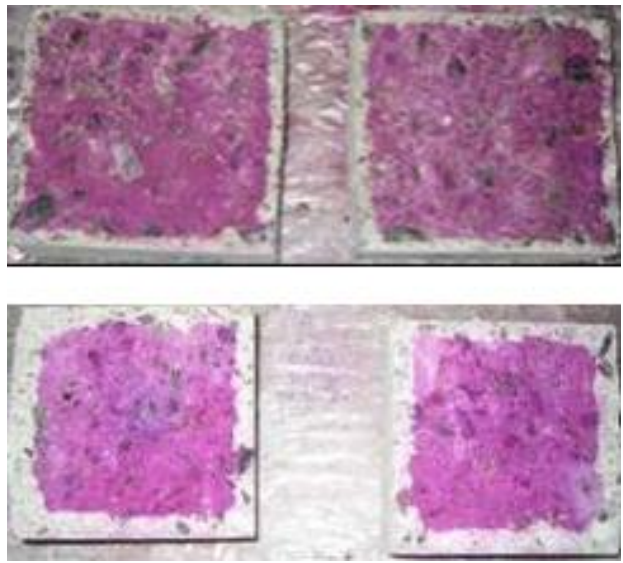

**Figure 3.2.** Carbonation fronts of two groups of concrete specimens after 56 days in a climatic control chamber (Publication II).

The amount of deicing salt (NaCl) that spread on HW7 from 2007 – 2013 (the considered period of experimental data) was about 0.99 kg/m² with an average of 102 salting instances. The daily average number of vehicles riding on HW 7 was estimated about 27,000 of which about 13% are heavy vehicles. The concrete specimens were placed in an array at a distance of 4.5 m, 6 m, 8 m, and 10 m from the HW 7 lane. All the specimens were placed on wooden stands which were installed on a gravel bed in order to avoid the probable water suction through the lower surfaces of the specimens. Field maintenance was performed in regular manner in order to assure that the surfaces of the specimens were exposed to splash water and water vapour. The chloride concentrations in concrete specimens were evaluated after one, three, and six years of exposure in the field environment. The chloride profile analyses were executed by extracting cylinder cores with size of (ø 100 mm, height >100 mm) from the concrete specimens. Dust samples were collected from the cored cylinders using a profile grinding technique at different depths. The examined depths of chloride range from 0.5 mm to 26 mm with increments of diverse orders. In addition, the nonsteady-state chloride diffusion coefficient ($D_{nssm}$) of all the specimens was examined by the rapid chloride migration (RCM) test in laboratory. In order to conduct this lab test, concrete cylinder specimens (three for each mix categories) with size of (ø 98 mm, height = 250 mm) were produced. The specimens were sliced at a thickness of 50 mm to form specimens size of (ø 98 mm, height = 50 mm) in accordance with NT Build 492 [117].

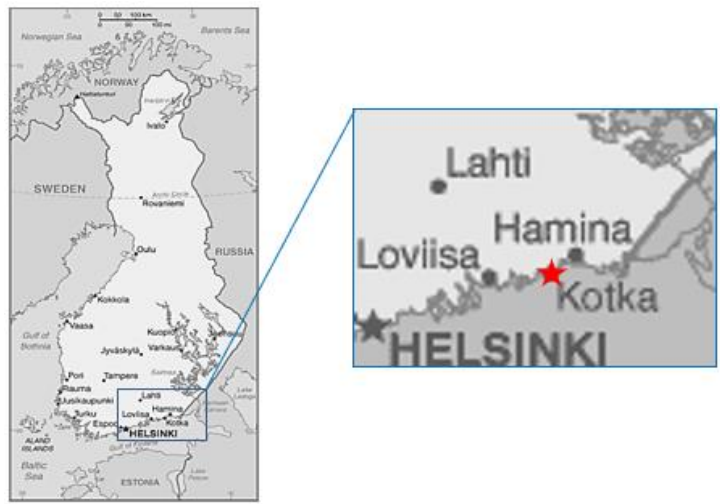

**Figure 3.3.** Map of Finland and Kotka where the concrete specimens for chloride field tests are located (Publication III).

## 3.3 Case structure for hygrothermal measurement

The experimental data for hygrothermal performance assessment were gathered from a six-storey building with surface-protected prefabricated RC sandwich panels. The building was constructed in 1972 and is located in the city of Vantaa, Finland. The exterior wall of the building is sandwich-type panels where the thermal insulation lies between two RC panels. They are connected to each other by steel trusses. The finishing type of the concrete façade members were brushed and coated. The average thickness of the outermost layers of the concrete panels is 53 mm with surface area of 7.84 m² (2.82 m width and 2.78 m height). This type of prefabricated RC sandwich panels were, and still are, predominantly utilized in Finnish multi-storey residential buildings [19,118–120].

Previously coated six concrete façade members from the southeast side of the case building were selected for hygrothermal behaviour investigation. The old coating from the outermost layer of the façade elements was removed using sand-water blasting method. Among the designated six concrete façade members, five of them were repaired with surface-protection systems after performing all the essential surface preparations. The applied surface treatments are labelled as S1, S2, S4, S5, and S6 as illustrated in Figure 3.4. The cleaned but the uncoated façade element is labelled as S3 and used as a reference. The surface treatments that were applied on the outermost layer of the façade elements can be categorized into two: cementitious and organic coatings. The S1 and S2 façade elements were treated with cementitious materials, whereas S4, S5, and S6 were coated with organic coating materials obtained from different Finnish companies. Cementitious coatings form a broad class that ranges from true cement-based coatings of a few to 10 mm thick. While, organic coatings form a continuous polymeric film on the concrete surface with a thickness ranging from 100 to 300 µm [11]. According to European Standard EN 1504, all the applied surface treatments in the concrete façade elements are able to limit the moisture content and to increase the concrete resistivity under rehabilitation principles P2 and P8, respectively. The types of the surface treatments applied in the five concrete façade members with the application methods are listed in Table 3.1.

The hygrothermal conditions of the ambient and inside the outermost surface-protected RC panels were measured using relative humidity/temperature probes. One probe for each concrete façade element was installed to measure the inner relative humidity and temperature, whereas one probe was mounted on the surface of a façade for ambient measurement. In order to mount the inner probes, holes were bored to a depth of about 40 mm at an approximate angle of 45° at the central area of the concrete façade members. Schematic representation of the mounted probe is illustrated in Figure 3.5. The cables of the probes were connected to a data logger to record the hygrothermal measurements. The ambient and the inner hygrothermal conditions were recorded with a regular time interval of half an hour for 719 days. Before installing the probes, they were calibrated using two-point calibration technique in accordance with the manufacturer's guide. The same calibration technique was also followed every six months after the installation of the probes.

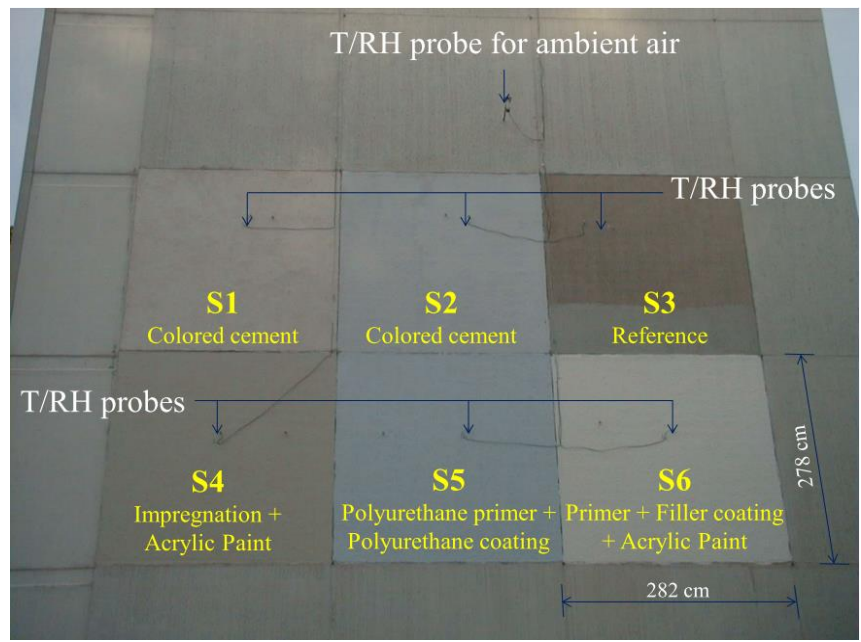

**Figure 3.4.** Concrete façade elements of the case structure for hygrothermal behaviour measurement (Publication IV).

**Table 3.1.** Concrete façade elements: treatment types and application methods (Publication IV).

| Façade labels | Treatment types | Application methods |
|---|---|---|
| S1 | Coloured cement coating | 1 x trowel |
| S2 | Coloured cement coating | 2 x brush |
| S4 | Impregnation | 1 x roller |
|  | Acrylic Paint | 2 x roller |
| S5 | Polyurethane primer | 1 x brush |
|  | Polyurethane coating | 2 x brush |
| S6 | Primer | 1 x roller |
|  | Filler coating | 1 x roller |
|  | Acrylic Paint | 2 x roller |

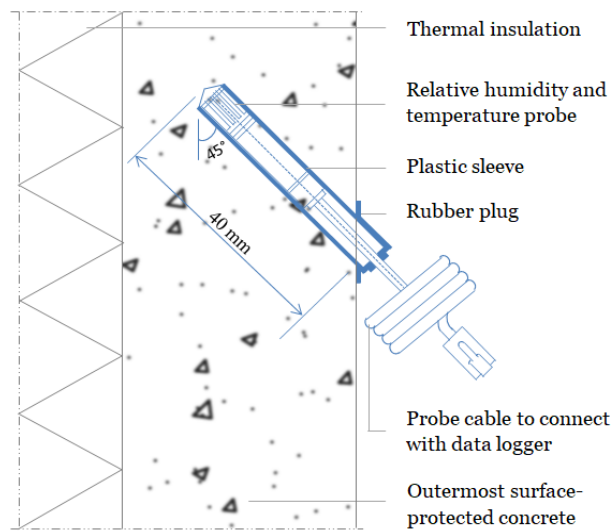

**Figure 3.5.** Schematic representation of the installed probe on concrete façade elements (Publication IV).

## 3.4 Model development for carbonation depth prediction

The development process of the carbonation depth prediction model, CaPrM, is presented in this section. CaPrM is designed by integrating four machine learning algorithms: neural network, decision tree, bagged and boosted decision trees. The overall CaPrM development process is shown in Figure 3.6. The initial task of the model development process is importing gathered experimental raw data that comprises parameters which describe the concrete mix ingredients, the concrete properties, the curing and field exposure conditions as well as the depth of carbonation measured at different ages. Then the execution of basic data exploration and data preprocessing follows. In any machine learning based methods, data often needs to be cleaned during preprocessing stage before they are processed further since unclean data may produce misleading results. For example, there may be incorrect or missed values in the training dataset and these values need to be rectified so that the model can analyse the data appropriately. Preprocessing phase could also include other tasks, such as data encoding and normalization.

After successful data preparation, the next stage is splitting the data into training, validation, and test subsets. The training data comprise the values of the input and the target parameters in order to learn a general rule that maps inputs to the desired output. The employed learning algorithms to train the data in the CaPrM were neural network, decision tree, bagged and boosted decision trees. These learning algorithms are commonly utilized to solve complex nonlinear regression problems efficiently, and thus making it suitable for carbonation depth prediction. The novelty of the proposed method lies in its ability to integrate the above four powerful learning methods and advanced optimization techniques. Each integrated algorithm learns from the training data and make predictions. Some of the employed machine learning methods requires the user to adjust certain controlling parameters in order to optimize their performance. The adjustment is carried out by evaluating the performance using validation dataset in case of neural network and cross-validation (subset of the training set) for decision tree based learning methods. After performing successful training and the necessary parameter adjustment, the prediction ability of the proposed carbonation depth prediction method is evaluated using a test dataset that are different from the training dataset. The integration of the four learning algorithms provides the opportunity to select the best performing learning methods among the four choices, by comparing the test error matrix, since there is no a single machine learning technique that performs optimal all the time. The incorporated ensemble methods enable selection of the influential carbonation predictors among the considered parameters. The major tasks involved in the development process are represented by grey coloured rectangular boxes.

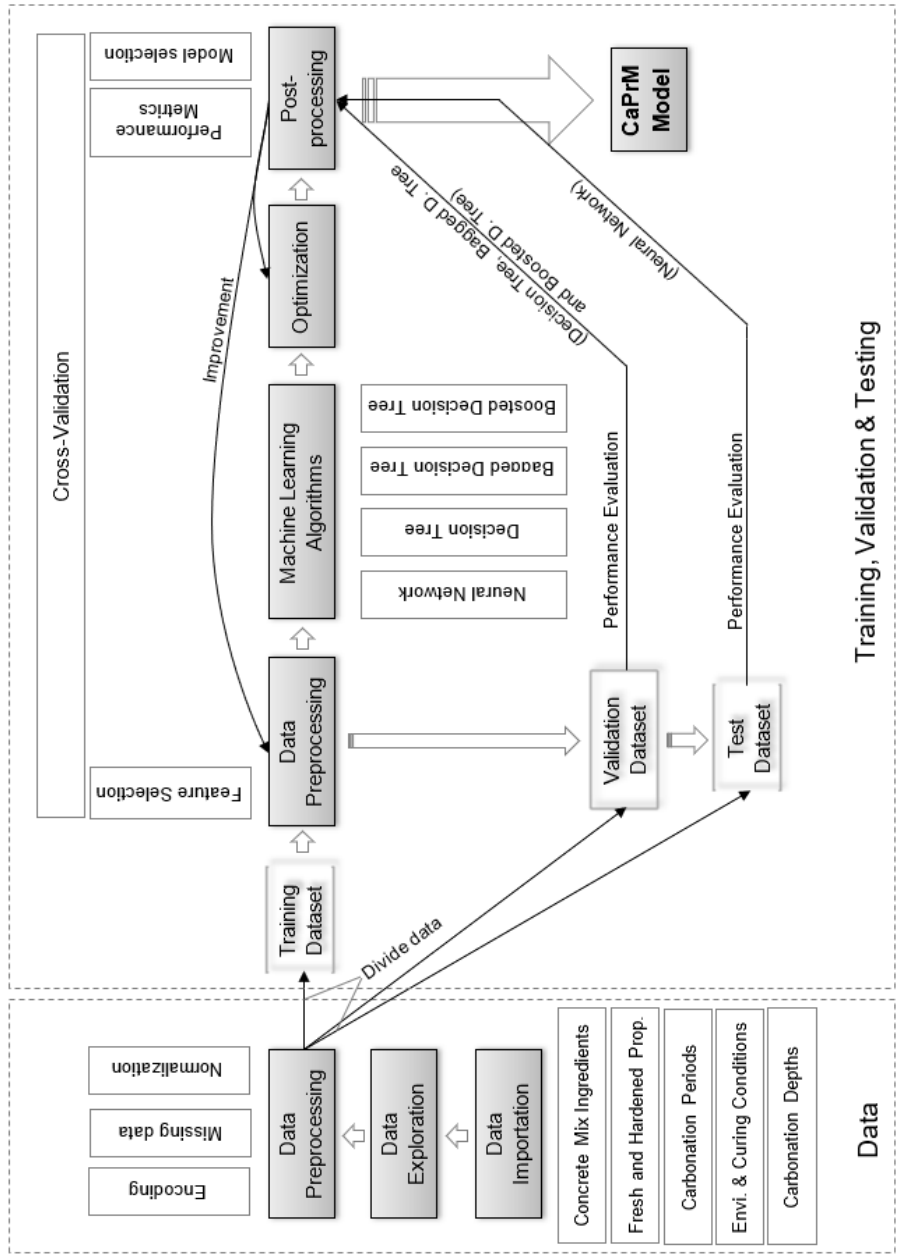

**Figure 3.6.** Development process of the CaPrM (Publication II).

### 3.4.1 Data for CaPrM

The experimental data employed in the CaPrM entail information regarding the concrete mix ingredients, the fresh and hardened concrete properties, the carbonation periods, the environmental and curing conditions, as well as the carbonation depths. The details of the data are given in Table 3.2. A total of six kinds of cements, based on the categorization of EN 197-1 [121], were utilized. These are Portland cement (CEM I 42,5 N-SR, CEM I 52,5 N and CEM I 52,5 R), Portland limestone cement (CEM II/A-LL 42,5 R), Portland composite cement (CEM II/A-M(S-LL) 42,5 N) and Portland slag cement (CEM II/B-S 42,5 N). Portland limestone cement (partially replaced with pulverized blast-furnace slag (BFS) or fly ash (FA)) was also utilized to produce few of the concrete specimens. The water-to-binder ratio (w/b) of the data ranges from 0.40 to 0.60. Every concrete mixture employs one type of plasticizer obtained from three producers, *VB-Parmix*, *Glenium G 51* or *Teho-Parmix*. An air-entraining agent (called either *Ilma-Parmix* or *Mischöl)* was introduced in each concrete mix type. Normally concrete admixtures are broadly classified into subcategories based on their chemical nature. Today, diverse type of plasticizers and air-entraining agents are available on the market from different sources. The performance of the chemical admixtures obtained from different sources are less uniform even if they are under the same subcategory or even with identical chemical compositions [122,123], causing difficulty to standardized them [28]. Due to this fact, the type of plasticizers and air-entraining agents applied for producing the concrete specimens are classified based on their brand names. It can be seen from Table 3.2 that the fresh concrete properties consist of test results of slump, slump flow of self-compacting concrete and air content. Tests that define the hardened concrete properties contain compressive strength and accelerated carbonation depth. The compressive strength of the concrete specimens was performed at the age of 28 days. The accelerated carbonation depth was tested at the age of 28 and 56 days.

It can be observed from Table 3.2 that there are 26 variables representing the employed data. Parameters numbered from 1 to 25 were used as input parameters and the last parameter, natural carbonation depth, which was measured at various age was designated as a target variable. The input variables comprise continuous and nominal data types, whereas the target variable entails only continuous data type. In Table 3.2, continuous variables are represented as C and nominal data types as N. Continuous variables are real numbers, such as results of quantitative measurements. Nominal variables are non-numeric and descriptive data types.

### 3.4.2 Data preprocessing for CaPrM

Data preprocessing is an essential step in the development process of any machine learning based methods. It often includes data encoding, missing data processing, data normalization and data partitioning. In the CaPrM development process the following data preprocessing tasks were executed.

*Data encoding*
Not all machine learning methods process heterogeneous data types (continuous and nominal). For instance, decision trees support both continuous and nominal data types without any problem, but neural network lacks the ability to process nominal data types. Hence, in the CaPrM all the non-numeric variables need to be encoded as numerical variables since one of the integrated learning methods is neural network. To do this, the most commonly applied encoding technique "1-of-N" was applied. The encoded variables are listed in Table 3.3.

**Table 3.2.** Description of variables employed in the carbonation dataset (C: continuous and N: nominal) adopted from Publication II.

| Variables category | No. | Variable subcategory | Description | Unit | Type and range | Short name |
|---|---|---|---|---|---|---|
| Concrete mix ingredients | 1 | Binder types | CEM I 42,5 N – SR | - | N:*** | Cem. types |
| | | | CEM I 52,5 N | - | | |
| | | | CEM I 52,5 R | - | | |
| | | | CEM II/A-LL 42,5 R | - | | |
| | | | CEM II/A-M(S-LL) 42,5 N | - | | |
| | | | CEM II/B-S 42,5 N | - | | |
| | | | CEM II/A-LL 42,5 R & blast-furnace slag | | | |
| | | | CEM II/A-LL 42,5 R & fly ash | | | |
| | 2 | Water-to-binder ratio | | - | C:0.37 to 0.60 | w/b |
| | 3 | Cement content | | [kg/m$^3$] | C:217.22 to 451 | Cement |
| | 4 | Blast-furnace slag content | | [kg/m$^3$] | C: 0 to 217.22 | BFS |
| | 5 | Fly ash content | | [kg/m$^3$] | C: 0 to 106 | FA |
| | 6 | Silica fume content | | [kg/m$^3$] | C: 0 to 0 | SF |
| | 7 | Total effective water | | [kg/m$^3$] | C: 153.60 to 185 | Total eff. water |
| | 8 | Aggregate content | Total aggregate | [kg/m$^3$] | C: 1706 to 1895 | Total Agg. |
| | 9 | | Aggregate < 0.125mm | [%]* | C: 2.40 to 6.70 | Agg. <0.125mm |
| | 10 | | Aggregate < 0.250mm | [%]* | C: 6.60 to 15.80 | Agg. <0.250mm |
| | 11 | | Aggregate < 4mm | [%]* | C: 36.30 to 53.20 | Agg. <4mm |
| | 12 | Product name of plasticizers | Glenium G 51 | - | N:*** | Plas. pro. name |
| | | | Teho-Parmix | - | | |
| | | | VB-Parmix | - | | |
| | 13 | Plasticizers content | | [%]** | C: 0 to 3.05 | Plasticizers |
| | 14 | Product name of air-entraining agents | Ilma-Parmix | - | N:*** | AEA pro. name |
| | | | Mischöl | - | | |
| | 15 | Air-entraining agents content | | [%]** | C: 0 to 0.06 | Air-ent. agents |
| Fresh concrete properties | 16 | Basic properties | Slump value | [mm] | C: 0 to 180 | Slump |
| | 17 | | SCC Slump-flow/T$_{50}$ | [mm/s] | C: 0 to 750 | Slump-flow/T$_{50}$ |

| | 18 | | Air content | [%] | C: 2.60 to 7.30 | Air cont. |
|---|---|---|---|---|---|---|
| Curing and field conditions | 19 | Curing conditions | uncontrolled | - | N:*** | Curing cond. |
| | | | controlled | - | | |
| | | | wet | - | | |
| | 20 | Field conditions | Temperature | [°C] | C: 6**** | Temp. |
| | 21 | | Relative humidity | [%] | C: 79**** | RH |
| | 22 | | $CO_2$ concentration | [ppm] | C: 375**** | $CO_2$ conc. |
| Hardened concrete properties | 23 | Mechanical property | Compressive strength | [MPa] | C: 32.80 to 58.50 | Comp. str. |
| | 24 | Durability properties | Accelerated carbonation depth | [mm] | C: 1.58 to 7.90 | Acc. carb. dep. |
| | 25 | | Age of the concrete at carbonation testing | [days] | C: 268 to 2585 | Carb. period |
| Carbonation depth | 26 | Natural Carbonation depth | | [mm] | C: 0.10 to 6.40 | Nat. carb. dep. |

*compared with the total aggregate, ** compared with the total binder materials, ***described in Table 3.3, ****yearly average

**Table 3.3.** 1-of-N encoding for non-numeric variables of the carbonation data (Publication II).

| Binder materials | | Curing conditions, product names of plasticizers and air-entraining agents | | |
|---|---|---|---|---|
| Nominal input variables | Encoded output | Nominal input variables | | Encoded output |
| CEM I 42,5 N – SR | [10000000] | Curing cond. | Uncontrolled | [100] |
| CEM I 52,5 N | [01000000] | | Controlled | [010] |
| CEM I 52,5 R | [00100000] | | Wet | [001] |
| CEM II/A-LL 42,5 R | [00010000] | Plasticizers | Glenium G 51 | [100] |
| CEM II/A-M(S-LL) 42,5 N | [00001000] | | Teho-Parmix | [010] |
| CEM II/B-S 42,5 N | [00000100] | | VB-Parmix | [001] |
| CEM II/A-LL 42,5 R & BFS | [00000010] | Air-ent. agents | Ilma-Parmix | [10] |
| CEM II/A-LL 42,5 R & FA | [00000001] | | Mischöl | [01] |

*Data normalization*

Normalization of the data before processing them in the neural network is a standard practice. It puts different variables on a common scale and is highly essential especially if the variables are in divergent scales. All the input and target variables are normalized using the formula presented in Equation (3.1) [124].

$$y = (y_{max} - y_{min}) * \frac{(x - x_{min})}{(x_{max} - x_{min})} + y_{min}, \qquad (3.1)$$

where $y$ is the normalized value of the variable; $y_{max}$ is the maximum value of the normalization range, (+1); $y_{min}$ is the minimum value of the normalization range, (-1); $x$ is the original inputs or target variables; $x_{max}$ is the maximum value for variable $x$; and $x_{min}$ is the minimum value for variable $x$. If $x_{max} = x_{min}$ or if either $x_{max}$ or $x_{min}$ are non-finite, then $y = x$ and no change occurs. After normalization, the values of the inputs and target fall in the interval [-1, 1].

*Missing data*

In any data driven based models, data quality plays a vital role in controlling the performance of the model. The amount of missing data less than 1% is generally considered trivial and 1–5% is manageable. Nevertheless, 5–15% requires advanced methods to correct it and more than 15% may severely impact any kind of interpretation [125]. Fortunately, all the gathered experimental data in the CaPrM are complete, though the Finnish DuraInt-project has missing values for some of the variables.

*Data partitioning*

The data employed in the CaPrM (92 instances and 26 features) were divided into training, validation, and test subsets for the neural network learning algorithm. The training dataset are utilized for computing the gradient and updating the network weights and biases. The validation dataset is used to halt the training when the generalization process stops improving, and thus avoiding overfitting. The purpose of the test dataset is to evaluate the predictive performance of the developed model. The training, validation, and test dataset represented 60%, 20% and 20% of the original data, respectively. Unlike neural network, the data for the decision tree was partitioned into training and test subsets by applying *K*-fold cross-validation technique. In case of limited data, *K*-fold cross-validation method is the best alternative in order to attain an unbiased estimate of the system performance, which in turn enhance the generalization ability of the model without overfitting [77]. In *K*-fold cross-validation, the training data is arbitrarily divided into *K* subsets with roughly identical sizes. One of the *K* subsets is applied as a test dataset for evaluating the model and the remaining (*K-1*) subsets as a training dataset. In total, *K* models are fit and *K* validation statistics are obtained. The predictive accuracy evaluations from the *K*-folds are averaged to provide a measure of the overall predictive performance of the model. Algorithm 1 presents the procedure of *K*-fold cross validation. In case of bagging decision tree, the training and test subset was formed based on the embedded sampling

---

**Algorithm 1:** *K-fold cross validation*

**Input:** Training dataset $D = \{(x_i, y_i), i = 1, \ldots, N\}$, where $y \in \mathbb{R}$

**Output:** Cross-validation estimate of prediction error, $CV(\hat{f})$

1: let randomly partition $D$ to $K$ roughly equal-sized parts
2: for the $k^{th}$ par$t$ $k = 1, \ldots, K$, fit the model to $K - 1$ parts of the training data $D$
3: do the above for the $k^{th}$ part and combine the $K$ estimates of the prediction error

Let $k: \{1, \ldots, n\} \to \{1, \ldots, K\}$ denote the indexing function that reveals the partition to which observation $i$ is assigned by the randomization. Then the prediction error of the cross-validation estimate is given by:

$$CV(\hat{f}) = \frac{1}{K} \sum_{k=1}^{K} L\left(y_i, \hat{f}^{-k(i)}(x_i)\right)$$

where $\hat{f}^{-k}(x)$ denote the function fitted with the $k^{th}$ part of the data removed.

**Table 3.4.** Type of algorithms, data size, and data partitioning applied in the CaPrM modelling process.

| Applied algorithms | Instances | Features | Data partitioning |
|---|---|---|---|
| Neural network | 92 | 26 | 60/20/20 |
| Decision tree | 92 | 26 | 10-fold cross validation |
| Bagging decision tree | 92 | 26 | 63/37 |
| Boosting decision tree | 92 | 26 | 70/30 |

procedure that underline in the method as presented in Section 2.4.3. In case of boosting decision tree, the original data were partitioned into training and test subset, covering 70% and 30% respectively. Using the training dataset, decision trees were grown sequentially until optimal size of the boosting ensemble is defined by cross validation. The applied data partitioning in the CaPrM modelling process is presented in Table 3.4 along with data size and type of algorithms.

### 3.4.3 Training for CaPrM

The adopted training algorithm in case of neural network is the fastest backpropagation algorithm that updates weight and bias values according to Levenberg-Marquardt optimization [126]. This algorithm computes the error contribution of each neuron after a batch of training data is processed. The error computed at the output is distributed back through the network layers in order to adjust the weight of each neuron. So, the network can learn the internal representations that allow mapping of the 25 input variables to the output (carbonation depth measured at different exposure times).

The Levenberg-Marquardt algorithm was formulated to approach second-order training speed without computing the Hessian matrix [127,128]. When the performance function has the form of a sum of squares, then the Hessian matrix can be estimated and described by Equation (3.2).

$$H = J^T J. \tag{3.2}$$

The gradient can be computed and expressed by Equation (3.3).

$$g = J^T e, \tag{3.3}$$

where $J$ is the Jacobian matrix that holds first derivatives of the network errors with respect to the weights and biases, and $e$ is a vector of network errors.

The Levenberg–Marquardt algorithm uses this approximation to the Hessian matrix in the Newton-like update, Equation (3.4).

$$x_{k+1} = x_k - [J^T J + \mu I]^{-1} J^T e, \tag{3.4}$$

where scalar parameter $\mu$ will ensure that matrix inversion will always yield a result.

Levenberg–Marquardt algorithm is effective and firmly suggested for neural network training [127,128]. This algorithm is fast when training neural networks measured on sum-of-squared error since it is tailored for such type of functions. Indeed, for large networks, it requires a huge memory as the Hessian matrix inversion needs to be computed every time for weight updating and there may be several updates in each iteration. So, the speed gained by second-order approximation may be completely lost [128].

The fundamental architecture of the neural network model integrated in the CaPrM is identical with Figure 2.5. It has three layers: an input, a hidden and an output layer. The optimal number of neurons in the hidden layer was determined based on the generalization error after performing a number of trainings. The activation functions allocated for the hidden layer was hyperbolic tangent transfer function. This function generates outputs between -1 and 1 as the input of the neuron goes from negative to positive infinity. Linear transfer activation function was assigned to the output layer of the network since nonlinear activation function may distort the predicted output. It transfers the neuron's output by simply returning the value passed to it. The input layer did not have an activation function as their role is to transfer the inputs to the hidden layer. Detail of the applied transfer functions are presented in Section 2.4.1. The learning rate of 0.1 was applied during the model training to update the weights and biases. The updates are obtained by multiplying the learning rate with the negative gradient. The larger the learning rate is the bigger the step, and thus the algorithm becomes unstable. A learning rate that is too small require more training to converge since steps towards finding optimal parameter values which minimize the loss function are tiny. The applied learning rate yields small generalization error with reasonable computational time.

Validation dataset is utilized to stop the network training early if its performance on this dataset fails to improve. Generally, the validation error (the error on the validation dataset) will decrease during the initial training phase. But it will typically begin to increase when the network begins to overfit the data as illustrated in Figure 3.7. When the validation error rises, the training is halted and the weights that were generated at the minimum validation error are utilized in the network. This approach usually gives the best generalization. The test dataset does not have any effect on network training, but it is used to evaluate the generalization of the network further. The training and the test performance of the network is discussed in Section 4.2. A MATLAB built-in function, `trainlm`, was applied to train the neural network part of the CaPrM using the training dataset presented in Section 3.4.1.

Decision tree is one of the integrated learning methods in the CaPrM. A MATLAB function, `fitrtree`, was applied to grow a regression tree. In a similar way presented in Section 2.4.2, this function builds a tree that yields the best prediction of the outcome for the training dataset. It partitioned the feature space into a set of rectangles recursively, and then fit a simple model in each one. This process recursively repeated until it fulfils stopping criteria. The applied function is able to grow deep decision trees by assigning the following three default values as stopping criteria: i) $N-1$ for maximal number of decision split, where $N$ is the sample size of the training dataset, ii) one for minimum number of leaf node observations, and iii) ten for minimum number of branch node observations. The model developer can change the default values when building a regression tree in order to control its depth. In the CaPrM, the default three stopping criteria were implemented for fitting the tree using a training dataset which entails information regarding the concrete mix ingredients, the fresh and hardened concrete properties, the carbonation periods, the environmental and curing conditions, as well as the carbonation depths.

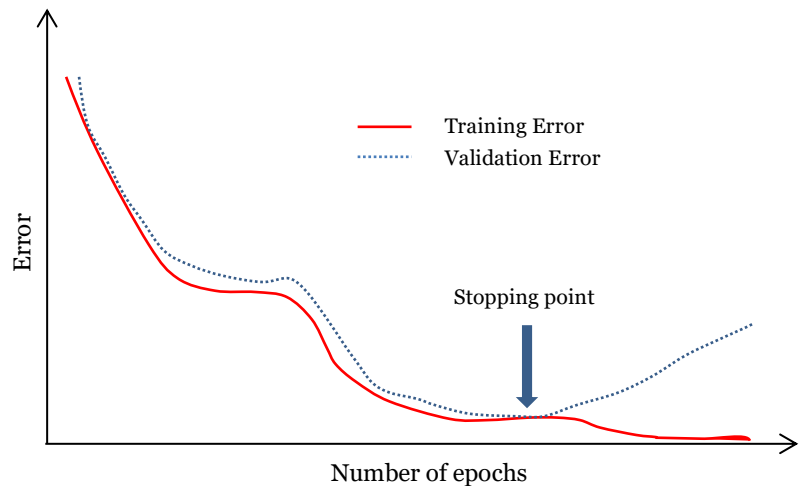

**Figure 3.7.** Illustration of training and validation errors as a function of epochs (training cycles).

The developed decision tree was cross validated to yield accurate prediction for new dataset since this method mitigates overfitting by testing for out-of-sample performance as part of tree building. Ten-fold cross validation was applied to evaluate the tree model as it has been the most common in machine learning based modelling practices. In fact, studies show that ten is the optimal number of folds that optimizes the time it takes to finalize the test while minimizing the bias and variance associated with the validation process [129]. Ten-fold cross validation randomly partitioned the training dataset into ten parts and trains ten new trees, each one on nine parts of the data. It then tests the predictive accuracy of each new tree on the data excluded in training of the corresponding tree. The mean square of the validation error was considered to evaluate the prediction performance of the developed decision tree. The generalization ability of the model was also examined using a test (previously unseen) dataset. The overall performance of the model is presented in Section 4.2.

Bagging and boosting decision trees are the two ensemble methods that are integrated in the CaPrM. As discussed in Section 2.4.3, the fundamental principle of an ensemble method is to aggregate multiple base models in order to enhance the prediction performance of a model. A MATLAB function, `fitensemble`, was used to build both bagging and boosting decision trees. This function returns a trained ensemble tree model that comprises the results of an ensemble of multiple decision tree based models. The development procedure of the base trees for both ensemble methods is the same as of the decision tree presented above. The main differences between the two methods are on sampling of the training dataset and the aggregation method of the trees. The pseudocode of the bagging decision tree is shown in Algorithm 2. It starts by defining the number of regression trees $T$ to be built. It then randomly draws multiple bootstrap samples from the original training dataset to create new training datasets $D^{*t}$ in order to build $T$ number of regression trees. Generally, a large number of trees will result in better accuracy but making it computationally expensive. In order to identify the optimal tree size, the bagging decision tree was trained using 300 trees and its MSE was computed using the out-of-bag instances. Then the out-of-bag error was evaluated with respect to the number of trees. The optimal number of trees that yield the lowest MSE was 150 trees. Using the newly formed training dataset, the bagging decision tree built an ensemble of 150 trees for predicting carbonation depth as a function of the multidimensional input variables. The performance of the developed bagging decision tree in predicting the carbonation depth was evaluated using the test dataset. The test outcomes of the model are presented in Section 4.2.

The boosted decision tree generally learned the training dataset $D = \{(x_i, y_i), i = 1, \ldots, N\}$ in sequence with improvement from one model to the next. As discussed in Section 2.4.3, there are several types of boosting ensemble methods. CaPrM adopted LSBoost (least-squares boosting) algorithm to minimize the cross-validated mean-square error of the ensemble. The applied LSBoost algorithm is summarized in Algorithm 3. It begins from the null model with residuals $r_i = y_i$ for all $i$ in the training dataset. Then it fit a decision tree to the residuals from the model instead

---

**Algorithm 2:** *Bagging*

**Input:** Training dataset $D = \{(x_i, y_i), i = 1, \ldots, N\}$, where $y \in \mathbb{R}$

**Output:** Bagged decision tree

1:    let $T$ be a total number of regression trees
2:    **for** $t = 1, 2, \ldots, T$ **do**
3:       create bootstrap samples $D^{*t}$ with equal number of intsances from a dataset $D$
4:       fit a tree $\hat{f}^{*t}(x)$ to the bootstrap sample $D^{*t}$
5:    **end for**

   **Output the bagged model**

6:    $\hat{f}_{bag}(x) = \frac{1}{T}\sum_{t=1}^{T}\hat{f}^{*t}(x)$

---

**Algorithm 3:** *LSBoost*

**Input:** Training dataset $D = \{(x_i, y_i), i = 1, ..., N\}$, where $y \in \mathbb{R}$

**Output:** Boosted decision tree, $\hat{f}_{LS\,Boost}(x)$

1:    let $T$ be a total number of regression trees and $\lambda$ is the learning rate
2:    initialize $\hat{f}(x) = 0$ and $r_i = y_i$ for all $i$ in the training data set
3:    **for** $t = 1, 2, ..., T$ **do**
4:       fit a tree $\hat{f}^t$ with $d$ splits ($d + 1$ terminal nodels) to the training data $(x, r)$
5:       update $\hat{f}$ by adding in a shrunken version of the new tree: $\hat{f}(x) \leftarrow \hat{f}(x) + \lambda \hat{f}^t(x)$
6:       update the residuals: $r_i \leftarrow r_i - \lambda \hat{f}^t(x_i)$
7:    **end for**

   **Output the boosted model**

8:    $\hat{f}_{LS\,Boost}(x) = \hat{f}(x) = \sum_{t=1}^{t} \lambda \hat{f}^t(x)$

---

of the outcome $y$. Sequentially, the algorithm updates the residuals by adding the newly generated decision tree into the fitted function. Each of these trees can be small to a certain extent by controlling the parameter $d$ in the algorithm. By fitting small trees to the residuals, the $\hat{f}$ will be slowly boosted in areas where the performance is weak. The shrinkage parameter (learning rate) $\lambda$ slows down the learning process further. For a small value of $\lambda$, the iteration number needed to attain a certain training error increase. In the CaPrM, the assigned learning rate was 0.1 and the number of trees was 150. Finally, the 150 tree models are combined to form a strong ensemble model. The prediction ability of the boosted decision tree was evaluated using a test dataset. The test results are presented in detail in Chapter Four.

### 3.4.4 Measuring importance of carbonation predictors

Besides predicting the depth of carbonation, the developed ensemble methods were applied to evaluate the importance of the input variables in estimating the carbonation depth. There are several ways to measure the importance of variables. Variable importance measurement by permuting out-of-bag observations (discussed in Section 2.4.3) is one of the commonly used methods. Nevertheless, in CaPrM another technique was applied since the described method is impracticable for boosting decision tree model. To measure the variable importance, `predictorImportance` function in MATLAB was applied for both bagging and boosting ensemble methods. This function averages the predictive measure of association for all input variables (predictors) over all trees in the ensemble. The predictive association measure is a value that shows the resemblance between decision rules that split observations. Among all viable decision splits that are compared to the best split (identified by growing the tree), the optimal surrogate decision split (that uses a correlated predictor variable and split criterion) provides the maximum predictive association measure. The second-best surrogate split yields the second-largest predictive association measure.

Assume $x_j$ and $x_k$ are input variables $j$ and $k$, respectively, and $j \neq k$. At node $t$, the predictive association measure, $\xi_{jk}$, between the best split $x_j < u$ and a surrogate split $x_k < v$ is described by Equation (3.5).

$$\xi_{jk} = \frac{\min(P_L, P_R) - \left(1 - P_{L_j L_k} - P_{R_j R_k}\right)}{\min(P_L, P_R)}, \tag{3.5}$$

where $P_L$ is the proportion of observations in the left child of node t, such that $x_j < u$; $P_R$ is the proportion of observations in the right child of node t, such that $x_j \geq u$; $P_{L_j L_k}$ is the proportion of observations at the left child node t, such that $x_j < u$ and $x_k < v$; $P_{R_j R_k}$ is the proportion of observations at right child node t, such that $x_j \geq u$ and $x_k \geq u$; $\xi_{jk}$ is a value in $(-\infty, 1]$. If $\xi_{jk} > 0$, then $x_k < v$ is a worthwhile surrogate split for $x_j < u$. Note: observations with missing values for $x_j$ or $x_k$ do not provide to the proportion computation.

The computed estimates of variable importance for the bagging and boosting decision trees are presented in Section 4.2. Every input variable employed in the training dataset to foresee the carbonation depth has one value. Variables that have obtained high value mean that they are important for the ensemble.

## 3.5 Model development for chloride profile prediction

In this section, the development process of chloride profile prediction model based on ensemble method is discussed. The ultimate purpose of the model is to determine the variables that are describe best the penetration of chloride into concrete. The model development process has four main steps: i) data, ii) bootstrap samples, iii) building multiple models, and iv) aggregation as shown in Figure 3.8. As any machine learning based model, the first step of the model development process is importing the experimental data. Though executing data preprocessing tasks is a standard practice, they were not performed during this model development process. This is due to the fact that the adopted ensemble method does not require data encoding and normalization. In addition, there are no missed values in the employed data that call for handling of missing values. The next step of the model development process is data partitioning to train and validate the model. The training sets are formed by drawing multiple bootstrap instances randomly from the dataset using bagging method. On average, each training set consists of about 63% of the original dataset. Any remaining samples (out-of-bag observations) after bootstrapping from the dataset are applied to evaluate the performance of the model training. The third step is building multiple chloride profile prediction models by utilizing the bootstrapped samples. The final step is aggregation of the models output in order to form an ensemble model. This is carried out by combining the predicted output of each model as discussed in Section 2.4.3. Once the ensemble model is created, the variable importance (VI) measures were carried out in order to determine the significance of chloride penetration controlling parameters in concrete. The whole process was iterated $i$ times to attain reliable results and then the average of these results become the VI measures.

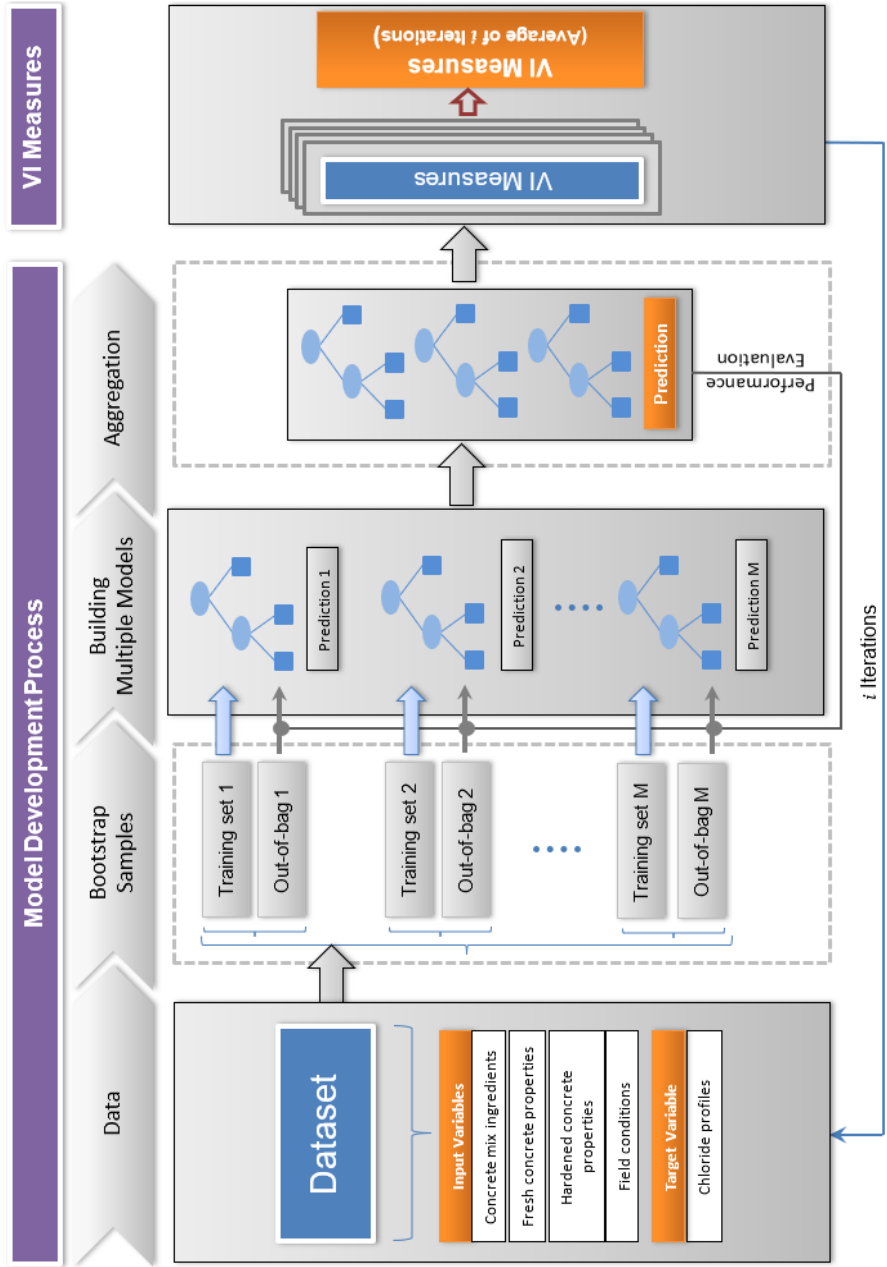

**Figure 3.8.** Chloride ingress prediction model development process and VI measures (Publication III).

### 3.5.1 Data for chloride prediction model

The data employed in the model entail information regarding the concrete mix ingredients, the fresh and hardened concrete properties, the field conditions and the chloride profiles measured at various exposure times. The details of the data are given in Table 3.5. A total of five types of cements, in accordance with the classification of EN 197-1 [56], were utilized to produce the concrete specimens. These are Portland cement (CEM I 42,5 N-SR and CEM I 52,5 R), Portland limestone cement (CEM II/A-LL 42,5 R), Portland composite cement (CEM II/A-M(S-LL) 42,5 N) and Portland slag cement (CEM II/B-S 42,5 N). Similar to the data utilized for carbonation depth prediction, few of the specimens adopted here were also produced using Portland limestone cement partially substituted with BFS or FA. The same types of plasticizers and air-entraining agents as in carbonation were also employed and classified based on their brand names for the same reason. The w/b ratio of the data spans from 0.38 to 0.51.

The data for the fresh concrete properties comprise test results of slump, density, and air content. Hardened concrete properties contain laboratory test results of pore volume, density (wet and dry), air void, compressive strength, carbonation diffusion coefficient and chloride migration coefficient. The test results of the pore volume provide information about the porosity of the concrete at early hardening phase. This property was tested at the same age with the wet and dry density which is at about the age of two days. The air void of the hardened concrete specimens was evaluated by thin-section analysis. The compressive strength was tested at the age of 28 days. The carbonation diffusion coefficient was computed after determining the carbonation depth of the concrete specimens at the age of 28 days. The chloride diffusion coefficient was examined at the age of three months.

The total number of variables employed in the data is 33. Variables numbered from 1 to 32 were assigned as input variables and the last variable (chloride profile of each specimen) was allocated as a target variable. The input variables encompass continuous and nominal data types whereas the target variable comprises only continuous data type. The continuous and the nominal data types are designated as C and N in Table 3.5, respectively.

### 3.5.2 Training of chloride prediction model

Bagging decision tree was developed to examine the importance of chloride penetration predicting parameters in concrete. The fundamental building process of bagging decision tree is described in Algorithm 2 (Section 3.4.3). Similar to CaPrM, the ensemble model was built in MATLAB but using a function known as `TreeBagger`. Ensembles created using `TreeBagger` algorithm have more functionality than those constructed with `fitensemble`. Both functions grow decision trees in the ensemble using bootstrap samples of the data, but the former selects a random subset of variables at each decision split and every tree involves several splits. Due to this, the ensemble method generated using `TreeBagger` function some-

**Table 3.5.** Description of variables employed in the chloride dataset (C: continuous and N: nominal, Publication III).

| Variables category | No. | Variable subcategories | Description | Units | Types and range | Short name |
|---|---|---|---|---|---|---|
| Concrete mix ingredients | 1 | Binder types | CEM I 42,5 N – SR | - | N: (1 = CEM I 42,5 N – SR, 2 = CEM I 52,5 R, 3 = CEM II/A-LL 42,5 R, 4 = CEM II/A-M(S-LL) 42,5 N, 5 = CEM II/B-S 42,5 N, 6 = CEM II/A-LL 42,5 R & blast-furnace slag, 7 = CEM II/A-LL 42,5 R & fly ash) | Bind. types |
| | | | CEM I 52,5 R | - | | |
| | | | CEM II/A-LL 42,5 R | - | | |
| | | | CEM II/A-M(S-LL) 42,5 N | - | | |
| | | | CEM II/B-S 42,5 N | - | | |
| | | | CEM II/A-LL 42,5 R & blast-furnace slag | - | | |
| | | | CEM II/A-LL 42,5 R & fly ash | - | | |
| | 2 | Water to binder ratio | | - | C: 0.37 to 0.51 | w/b |
| | 3 | Cement content | | [kg/m$^3$] | C: 217.22 to 451 | Cement |
| | 4 | Blast-furnace slag content | | [kg/m$^3$] | C: 0 to 217.22 | BFS |
| | 5 | Fly ash content | | [kg/m$^3$] | C: 0 to 106 | FA |
| | 6 | Total effective water | | [kg/m$^3$] | C: 159.50 to 180.40 | Total eff. water |
| | 7 | Aggregate content | Total aggregate | [kg/m$^3$] | C: 1706 to 1895 | Total Agg. |
| | 8 | | Aggregate < 0.125 mm | [%]$^*$ | C: 2.40 to 4.50 | Agg. <0.125 mm |
| | 9 | | Aggregate < 0.250 mm | [%]$^*$ | C: 6.60 to 11.40 | Agg. <0.250 mm |
| | 10 | | Aggregate < 4 mm | [%]$^*$ | C: 36.30 to 52.50 | Agg. <4 mm |
| | 11 | Product name of plasticizers | Glenium G 51 | - | N: (1 = Glenium G 51, 2 = Teho-Parmix, 3 = VB-Parmix) | Plas. pro. name |
| | | | Teho-Parmix | - | | |
| | | | VB-Parmix | - | | |
| | 12 | Plasticizers content | | [%]$^{**}$ | C: 0.60 to 2.54 | Plasticizers |
| | 13 | Product name of air-entraining agents | Ilma-Parmix | - | N: (1 = Ilma-Parmix, 2 = Mischöl) | AEA pro. name |
| | | | Mischöl | - | | |
| | 14 | Air-entraining agents content | | [%]$^{**}$ | C: 0.01 to 0.06 | Air-ent. agents |
| Fresh concrete properties | 15 | Basic properties | Slump | [mm] | C: 40 to 180 | Slump |
| | 16 | | Density | [kg/m$^3$] | C: 2287 to 2395 | Density |
| | 17 | | Air content | [%] | C: 3.40 to 6.90 | Air cont. |
| | 18 | Pore volumes and density | Air pores | [%] | C: 3.55 to 6.99 | Air pores |
| Hardened concrete properties | 19 | | Total porosity | [%] | C: 17.52 to 20.39 | T. porosity |
| | 20 | | Capillary + gel porosity | [%] | C: 12.87 to 14.68 | C+G porosity |
| | 21 | | Density (wet) | [kg/m$^3$] | C: 2502 to 2581 | Density (w) |
| | 22 | | Density (dry) | [kg/m$^3$] | C: 2354 to 2427 | Density (d) |
| | 23 | Thin section results | Total air pores | [%] | C: 1.90 to 5.90 | T. air pores |
| | 24 | | Air pores <0.800 mm | [%] | C: 0.80 to 4.60 | AP <0.800 mm |
| | 25 | | Air pores <0.300 mm | [%] | C: 0.60 to 3.50 | AP <0.300 mm |
| | 26 | | Specific surface | [mm$^2$/mm$^3$] | C: 12.80 to 36.50 | S. surface |
| | 27 | | Spacing factor (< 0.800 mm pores) | [mm] | C: 0.18 to 0.51 | SF < 0.800 mm |
| | 28 | Mechanical property | Compressive strength | [MPa] | C: 38 to 58.50 | Comp. str. |
| | 29 | Durability properties | Accelerated carbonation coefficient | [mm/d$^{0.5}$] | C: 1.58 to 3.96 | $k_{acc}$ |
| | 30 | | Chloride migration coefficient | [m$^2$/s] | C: 1.40 to 15.09x10$^{-12}$ | $D_{nssm}$ |
| Field conditions | 31 | Field conditions | Exposure time | [year] | C: 1 to 6 | Expo. time |
| | 32 | | Distance from highway lane | [m] | C: 4.50 to 10 | Dis. from HW |
| Chloride profiles | 33 | Chloride profiles | Chloride concentration at various depth | [%]*** | C: 0 to 0.10 | Chloride profile |

*compared with the total aggregate, **compared with the total binder materials, *** by weight of concrete times referred as random forest algorithm. The basic development step of random forest algorithm is summarized in Algorithm 4. In this dissertation, the developed ensemble method referred as bagging decision tree instead of random forest.

The first step for development of a powerful bagging decision tree is determining a suitable leaf size for each decision tree in the ensemble. In fact, the default minimal leaf size of the adopted algorithm to build bagging decision trees is five. Trees grown with this value are usually very deep and optimal for determining the predictive power of an ensemble. Bagging decision trees grown with larger leaves may not lose their predictive power while reducing training and prediction time as well as memory usage. Due to these facts, it is necessary to find the optimal leaf size. This can be attained by building ensemble trees employing the training dataset with dissimilar leaf sizes and rational number of trees. Then assess which of the tree configuration option offers the least mean-square error (MSE). To determine the suitable leaf size, ensemble trees with tree size of 100 and leaf sizes of 5, 10, 20, 50 and 100 were built. For each ensemble trees, the out-of-bag predictions were computed by averaging over predictions from

---

**Algorithm 4:** *Random forest*

**Input:** Data $D = \{(x_i, y_i), i = 1, \ldots, N\}$, where $y \in \mathbb{R}$

**Output:** Random forest (Bagged decision tree)

1:   let $T$ and $B$ be a total number of regression trees and nodes, respectively
2:   **for** $t = 1, 2, \ldots, T$ **do**
3:      create a bootstrap sample $D^{*t}$ with equal number of intsances from a dataset $D$
4:      grow a random-frost tree $T_t$ to the bootstrapped data until the minimum node size is reached
5:      **for** $b = 1, \ldots, B$ **do**
6:         select $m$ variables at random from the $p$ variables
7:         choose the best split among the $m$ variables
8:         split the node into two daughter nodes
9:      **end for**
10:  **end for**

    **Output the random forest**

11:  $\hat{f}_{rf}(x) = \frac{1}{T}\sum_{t=1}^{T} T_t(x)$

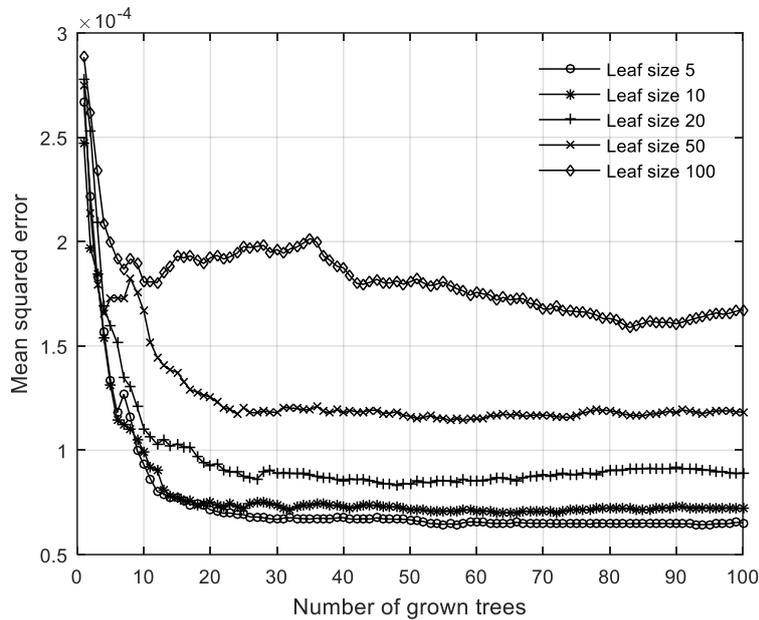

**Figure 3.9.** Mean-square errors to determine optimal tree and leaf sizes for chloride profile prediction model (Publication III).

all trees in the corresponding ensembles. Then the MSEs of each ensemble trees were computed by averaging the squared difference between the predicted responses of the out-of-bag and the target responses. The MSEs obtained by the ensemble methods for the examined leaf and tree sizes are shown in Figure 3.9. Even if the errors for leaf size five and ten are comparable as can be seen from Figure 3.9, leaf size five provides the lowest MSE. Therefore, to carryout effective model training the leaf and the tree sizes were designated as five and 100, respectively. Using the identified configuration, trees were grown for each bootstrap replica and train each tree in the ensemble. The bootstrap replicas comprise information about the concrete's mix ingredients, the fresh and hardened concrete properties, the field conditions as well as the chloride profiles measured at different ages.

### 3.5.3 Measuring importance of chloride predictors

The objective of the variable importance analysis is to determine the degree of significance of each variable which are embedded in the dataset in predicting the chloride concentration in concrete. It was discussed above that the ensemble decision trees of CaPrM built using `fitensemble` algorithm had applied a function, `predictorImportance,` to examine how influential the input variables in predicting the carbonation depth. Its equivalent for `TreeBagger` algorithm is a function called `OOBPermutedVarDeltaError`. This function gives a numeric array of size (1-by-number of variables) consisting of significance measure for each input variable. This was executed by arbitrarily permuting out-of-bag data across a single variable at a time and predicting the increase in the out-of-bag error due to the permutation. This measure was computed for each tree in the ensemble then averaged them and the averaged value was divided by the standard deviation over the whole ensemble. This process was reiterated ten times since this number of iterations offered stable outcome with rational computational time. The ultimate variable importance measure for each variable was evaluated by averaging the results of the ten iterations. The higher the value of the variable importance measure the greater the significance of the variable in predicting the chloride concentration in concrete. The process of out-of-bag variable importance measurement by permutation is already discussed in detail in Section 2.4.3 and summarized by Algorithm 5.

In order to evaluate the significance of variables from various perspectives, ten bagged decision tree based chloride profile prediction models were developed by following the same procedure presented above. Based on the parameters in their input dataset, the ten models were categorised into two groups: Model A and Model B. The details of the classifications of the ten models are presented in Table 3.6. Model A utilized all parameters presented in Table 3.5 except chloride profiles as input parameters. Model B employed input parameters representing only concrete mix ingredients, field exposure conditions, and chloride migration coefficients. The purpose of Model B was to study the importance of fresh and hardened concrete tests in predicting the chloride profile. In both model categories, the target dataset was the chloride profile. Each group was further divided into three scenarios to analyse the parameter's significance by excluding the influence of the exposure time and distance from highway. The number of instances and features

---

**Algorithm 5:** *Out-of-bag VI measure by permutation*

**Input:** Data $D = \{(x_i, y_i), i = 1, \ldots, N\}$, where $y \in \mathbb{R}$

**Output:** Out-of-bag VI measure

Let $D^{*t}$ the bootstrap samples and $p$ is the number of predictors in the training dataset.

1: **for** $t = 1, 2, \ldots, T$ **do**
2:    identify the out-of-bag instances for a tree $t$, $\bar{\beta}^{(t)} \subseteq \{1, \ldots, p\}$
3:    estimate the out-of-bag error $\hat{y}^{(t)}$
4:    **for** each predictor variable $x_j$, $j \in \bar{\beta}^{(t)}$
5:      randomly permute the instances of $x_j$
6:      estimate the model error, $\hat{y}_{\varphi j}^{(t)}$, using the out-of-bag instances containing the permuted values of $x_j$
7:      take the difference $\hat{y}_j^{(t)} = \hat{y}_{\varphi j}^{(t)} - \hat{y}^{(t)}$ // Predictor variables not split when growing tree $t$ are attributed a difference of 0.//
8:      **for** each predictor variable in the training dataset, $D^{*t}$
9:         compute the mean, $\bar{y}_j$ of the differences over the learners, $j = 1, \ldots, p$
10:         standard deviations, $\sigma_j$ of the differences over the learners, $j = 1, \ldots, p$
11:      **end for**
12:    **end for**
13: **end for**

   **Output out-of-bag VI measure for $x_j$**

14: $\widehat{VI}(x_j) = \dfrac{\bar{y}_j}{\sigma_j}$

---

considered in each model is different and presented in Table 3.7 along with the adopted type of algorithm and data partitioning.

The first scenario (A.1 and B.1) takes into account all the respective variables of Models A and B as described above. The second scenario (A.2 and B.2) is identical with scenario one except the employed chloride profiles comes only from the specimens located at 4.5 m. The intention of this scenario is to avoid the distance effect on predicting the chloride profile and concentrate on the influence of other parameters. The focus of the third scenario is to eliminate the influence of the exposure time by considering the chloride profile measured at a specific exposure time. Under this scenario, there are three models in each model group (A.3(i), A.3(ii), A.3(iii) and B.3(i), B.3(ii), B.3(iii)) since the chloride concentration in the concrete specimens were analysed at three different exposure times (one, three, and six years). All these models employed the chloride profiles measured at the depth of 0.5 mm, 1.5 mm, 3 mm, and 5 mm since the amount of chloride was examined in all concrete specimens at these depths. Each model was run ten iterations and the final variable importance measure was evaluated by averaging the results of the ten iterations. The findings are presented in Section 4.3 in detail.

**Table 3.6.** Details of the ten models developed for chloride profile prediction (Publication III).

| Scenario | Model A | | | Model B | | | Description |
|---|---|---|---|---|---|---|---|
| | Model name | Input variables category | No. of variables | Model name | Input variables category | No. of variables | |
| 1 | **A.1** | All variables* | 33 | **B.1** | All concrete mix ingredients*, field conditions* and $D_{nssm}$ | 18 | Entails data at which the chloride profiles measured on all specimens at different ages. |
| 2 | **A.2** | All variables* except dis. from HW | 32 | **B.2** | All concrete mix ingredients*, expo. time and $D_{nssm}$ | 17 | Consists of data at which the chloride profiles measured on specimens only located at 4.5 m away from HW 7. |
| 3 | **A.3(i)** | All variables* except expo. time | 32 | **B.3(i)** | All concrete mix ingredients*, dis. from HW and $D_{nssm}$ | 17 | Entails data of specimens at which the chloride profiles measured at one year of exposure. |
| | **A.3(ii)** | | | **B.3(ii)** | | | Ditto but at three years of exposure. |
| | **A.3(iii)** | | | **B.3(iii)** | | | Ditto but at six years of exposure. |

**Table 3.7.** Type of algorithm, data size, and data partitioning used to build chloride profile prediction models.

| Applied algorithms | Model types | Instances | Features | Data partitioning |
|---|---|---|---|---|
| Bagged decision tree | Model A.1 and B.1 | 522 | 33 | 63/37 |
| | Model A.2 and B.2 | 345 | 32 | 63/37 |
| | Model A.3(i) and B.1(i) | 215 | 32 | 63/37 |
| | Model A.3(ii) and B.2(ii) | 120 | 32 | 63/37 |
| | Model A.3(iii) and B.2(iii) | 189 | 32 | 63/37 |

## 3.6 Model development for hygrothermal prediction

The development process of the hygrothermal prediction model is discussed in this section and illustrated in Figure 3.10. The rectangular boxes coloured in grey in the figure represent the major tasks of the modelling procedure. As in any other data-driven models, the initial step was importing the monitored hygrothermal data. The data comprise the hygrothermal measurements of the ambient and the inner surface-protected concrete façade elements. After importing the data, data exploration was executed in order to comprehend and visualize the principal characteristics of the dataset. This activity is usually carried out by utilizing customized visual analytical tools. Data preprocessing was the next essential step and this task needs to be carried out before the data being processed further. Data preprocessing, especially in neural network based models, could also involve other tasks as discussed in Section 3.4.2. The next important step after data preprocessing was data partitioning in which the data were divided into training, validation, and test sets. The training dataset is a set of hygrothermal data that are utilized to train the adopted neural network algorithm (Nonlinear autoregressive with external input, NARX). The purpose of the validation dataset is to assist the network to halt training when the generalization process stops improving and thus preventing overfitting. The test dataset was utilized to evaluate the performance of the developed neural network based hygrothermal prediction model.

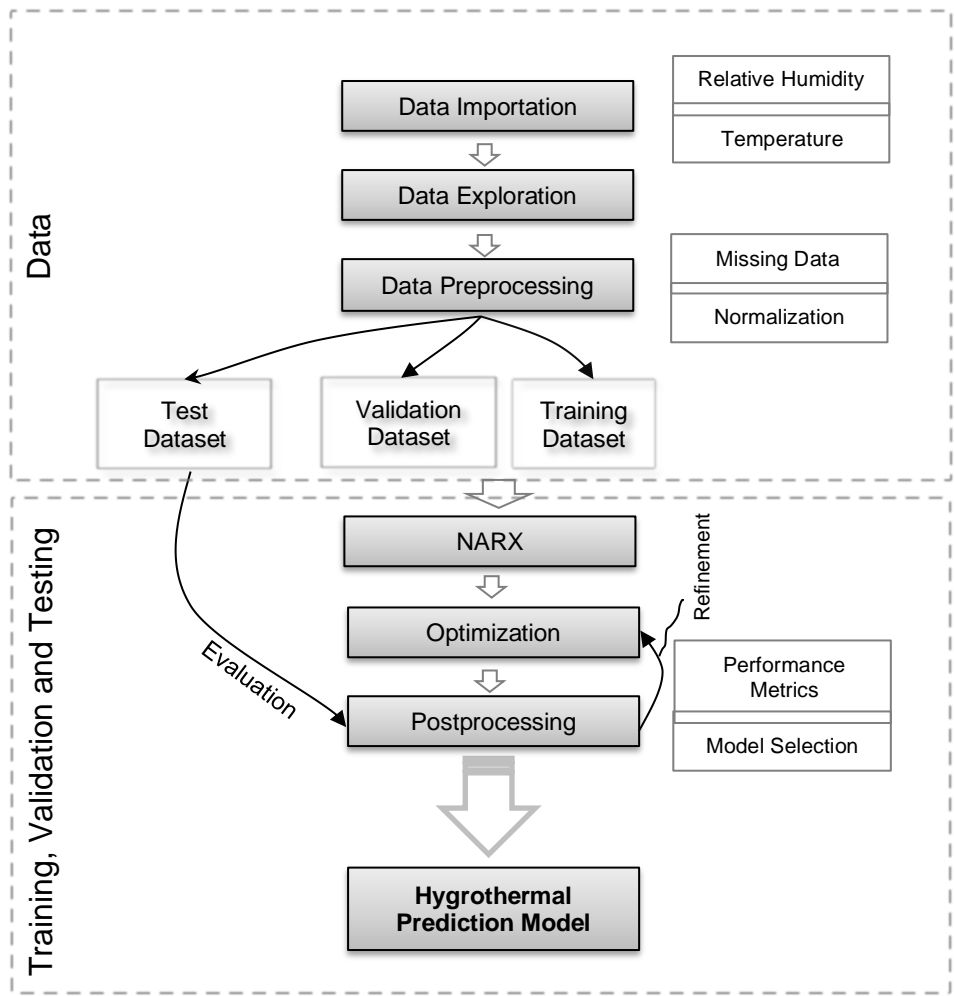

**Figure 3.10.** Hygrothermal prediction model development process (Publication IV).

### 3.6.1 Data and its preprocessing for hygrothermal model

The data employed in the hygrothermal prediction model entail four variables: ambient relative humidity, ambient temperature, inner relative humidity, and inner temperature. The data were gathered from two years of in-service monitoring of the concrete façade elements in the case structure. The data were recorded with a regular time interval of half an hour. The ambient relative humidity and temperature were designated as the input variables. The target variables were the values of the relative humidity and the temperature measured inside the concrete façade members. Both temperature and relative humidity values are numeric, and units used for them were °C and %, respectively. The size of the data utilized to predict the hygrothermal behaviour inside each concrete façade element is given in Table 3.8.

**Table 3.8.** Type of algorithm, data size, and data partitioning applied to build hygrothermal prediction model.

| Applied algorithm | Façade element | Algorithms | Instances | Features | Data partitioning |
|---|---|---|---|---|---|
| NARX | S1 | Relative Humidity | 719 | 2 | 75/15/10 |
| | | Temperature | 719 | 2 | 75/15/10 |
| | S2 | Relative Humidity | 719 | 2 | 75/15/10 |
| | | Temperature | 719 | 2 | 75/15/10 |
| | S4 | Relative Humidity | 719 | 2 | 75/15/10 |
| | | Temperature | 719 | 2 | 75/15/10 |
| | S5 | Relative Humidity | 719 | 2 | 75/15/10 |
| | | Temperature | 719 | 2 | 75/15/10 |
| | S6 | Relative Humidity | 719 | 2 | 75/15/10 |
| | | Temperature | 719 | 2 | 75/15/10 |

During the hygrothermal modelling process, data encoding was unrequired since the utilized data entail only numerical variables. The applied data normalization and data partitioning procedure were exactly the same as the procedure presented in Section 3.4.2. All the data were normalized by transforming the values of the input and the target variables in the interval [-1, 1]. The data (size of 719 instances and two features) were also randomly divided into three clusters: training, validation, and test datasets which hold 75%, 15% and 10% of the dataset, respectively. Missing data processing was carried out since about 6% of the monitored hygrothermal data were missed for successive days at particular times from every surface-protected façade element. To substitute the missing data and eliminate the noise from the monitored hygrothermal data, a moving average filter technique was applied. This method smooths and replaces the missing data with the average of the neighbouring data points defined within the span [130] and is expressed by Equation (3.6).

$$x_s(i) = \frac{1}{2M+1}\big(x(i+M) + x(i+M-1) + \cdots + x(i-M)\big), \tag{3.6}$$

where $x_s(i)$ is the smoothed value for the $i^{th}$ data point, $M$ is the number of neighboring data points on either side of $x_s(i)$, and $2M + 1$ is the span.

### 3.6.2 Training of hygrothermal prediction model

The model training process was carried out using a function `narxnet` in MATLAB. This function builds a NARX network with the default hyperbolic and linear transfer functions in the hidden and output layers, respectively. The hyperbolic activation function produces outputs in the range between -1 and 1 whereas the linear transfer function computes the neurons output by simply transferring the value given to it. As discussed in Section 2.4.1, the NARX network has generally two inputs. In case of the considered NARX network, one of the inputs is from the monitored hygrothermal data (e.g. ambient temperature/relative humidity), and the other input is a feedback connection obtained from the output of the model (e.g. predicted inner temperature/relative humidity). For each of these inputs, there is a tapped-delay-line memory to store previous values. In order to complete the architecture of NARX network, the number of tapped-delay-line memories and hidden neurons should also be assigned. The default number of the tapped-delay-line memories and hidden neurons for `narxnet` function is 2 and 10, respectively. These default values were employed for creation of the initial NARX network that predicts the hygrothermal performance inside surface-treated concrete element. Then the network was tested with other values and all training performances were compared. The comparison is necessary to identify the optimal number of hidden neurons and the tapped-delay-line memories that yield the best performance. The determined optimal values were utilized in the model as the final configuration of the model. The basic graphical representation of the designed NARX network for modelling the hygrothermal behaviour of the case structure is identical with Figure 2.7.

After the configuration of the NARX network was completed, the next task was inputting the preprocessed hygrothermal data for model training. Since the network contains tapped-delay-lines, it is vital to fill initial values on the delay-line memories of the inputs and the outputs of the network. This was carried out by applying a command, prepa­rets, which automatically shifts the input and the target time series as many steps as required to fill the initial delay states. This function also reformats the input and the target data whenever the network is redesigned with different numbers of delays. After the training data is ready, the network was trained using a Levenberg-Marquardt algorithm. It is a fast, powerful, and widely applied algorithm to solve several types of problems. Details of this algorithm are already discussed in Section 3.4.3.

The performance of the network training was evaluated using the validation dataset for different networks with varying number of tapped-delay-line memories and hidden layer neurons. After performing several trainings, tapped-delay-line memories of two and hidden layer neurons of ten were identified as the optimal ones that provide the least generalization errors. Tapped-delay-line memories of two units mean that the output of the designed network, $\hat{y}(n+1)$, is fed back to the input of the network through delays, $\hat{y}(n)$, and $\hat{y}(n-1)$ as the output of the network is a function of these delays. The training performance of the developed NARX model to predict the relative humidity in concrete façade element S4 is illustrated, as an example, in Figure 3.11. It can be clearly observed that the network has been trained smoothly and any overfitting/underfitting has not been occurred. The best validation performance is obtained at nine epochs (full training cycles on the training dataset) with MSE of 1.05. The MSE of test dataset (which was randomly generated) is 1.16. After successful training, the performance of the model in evaluating the hygrothermal behaviour inside all surface-treated concrete façade elements were tested using the last 90 days of the data, which were not utilized during the model trainings as well as validation processes. The details of the training performance and the prediction ability of the developed hygrothermal prediction models on the last 90 days for all façade elements are presented in Section 4.4.

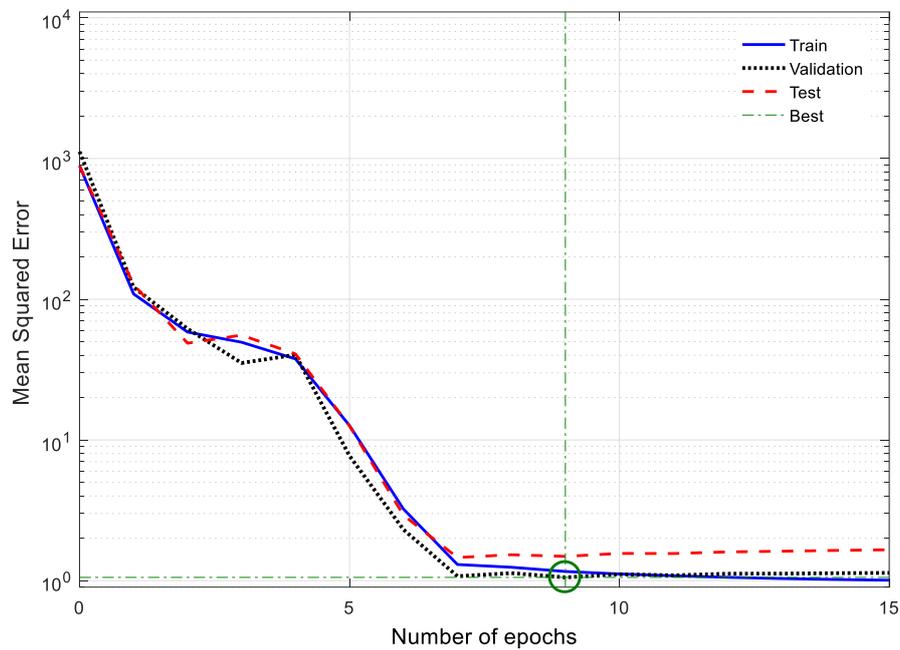

**Figure 3.11.** Validation performance of the trained NARX model that employs relative humidity data obtained from the ambient and inside S4.

## 3.7 Exploratory data analysis development for visualization

In this section, the development process of exploratory data analysis (EDA) technique is presented. Its primary aim is to visualize the condition of corrosion and other deterioration mechanisms caused unintentionally by the implemented surface-protection systems in a more realistic manner. EDA is an essential step in any data-driven modelling process, and it is often applied after data collection and preprocessing, where the data is merely visualized, plotted, and manipulated without any presumption. It assists to assess the data quality and build optimal models. EDA is not only guide for building a useful model but also assist to understand the output of the model.

The data image technique was applied for visualizing the status of corrosion, frost and chemical attacks in surface-protected concrete façade elements using the hygrothermal data. MATLAB programing language was used for displaying these images employing `imagesc` function. This function uses x and y values to specify the data size in respective directions, and then generates an image with dataset to scaled values with direct or indexed colours to indicate the magnitude of each variable for each observation. As the hygrothermal data were monitored with a regular interval of short time and comprise missed values, producing and visualization of the status of corrosion, frost and chemical attacks in the concrete elements is not a straightforward task. The following fundamental steps were carried out.

- Step 1: smooth the data to reduce irregularities of the monitored/predicted time-series hygrothermal data to deliver a clearer view of the true underlying behaviour.
- Step 2: translate the data to status of corrosion, frost and chemical attacks by utilizing predefined conventions.
- Step 3: display the image by applying the `imagesc` function and add a colour scale to the image.
- Step 4: define the coordinates of the missed values in concrete façade elements which comprise the missed data.
- Step 5: represent the missed data using the coordinates defined in step 4.

The visualized conditions of corrosion, frost, and chemical attacks of surface-protected concrete façade elements of the case structure are presented and analysed in Section 4.4.

# Chapter 4

# Results

This chapter presents the main results of the dissertation. The presentation of the results is organised into four sections based on the four research questions. Each section presents the core results of one of the research questions that are discussed in Chapter One.

## 4.1 Need for data-driven approaches

Publication I answers the research question of "*how to eliminate or mitigate the uncertainties observed in the traditional corrosion assessment methods?*". To answer this question, Publication I concentrates on: i) examining the limitations of the conventional corrosion assessment methods, and ii) reviewing the recent advances and future directions on the concrete durability assessment. The focus in this section is on the findings of (ii) as the first task has already been discussed in Section 2.3.

The findings of Publication I revealed that there had been few attempts to predict the carbonation depth and the chloride concentration in concrete using machine learning methods. The few proposed carbonation depth prediction modes are listed in Table 4.1. The majority of these works were based on short-term tests which were aimed to characterize the carbonation resistance of concrete at laboratory. They also consider few parameters, missing some of the important ones that describe the microstructure of concrete. The model proposed in [131] take into account 39 input parameters, but some of them do not describe the condition well. This is due to the fact that the data were collected from concrete specimens exposed to natural environment located in different exposure conditions. The data were acquired from 88 literatures. All works, except those executed by the author of this dissertation [132–134], failed to perform a certain crucial data optimization steps during the model development process. Without following these steps, some parameters may become

**Table 4.1.** Data-driven models that are proposed for carbonation depth prediction.

| Work | Main learning algorithm type | Exposure environment | | Exposure duration | | No. of input parameters | Data optimization |
|---|---|---|---|---|---|---|---|
| | | Lab | Field | Long | Short | | |
| [161] | Neural Network | X | ✓ | ✓ | X | 6 | X |
| [131] | Neural network | X | ✓ | ✓ | X | 39 | X |
| [162] | Neural network | ✓ | X | X | ✓ | 5 | X |
| [163] | Adaptive neuro-fuzzy inference system | X | ✓ | ✓ | X | 6 | X |
| [132] | Neural network | ✓ | X | X | ✓ | 15 | ✓ |
| [164] | Neural network | ✓ | X | X | ✓ | 3 | X |
| [133] | Decision tree Bagged decision tree | ✓ | X | X | ✓ | 15 | ✓ |
| This work [134] | Neural network Decision tree Bagged decision tree Boosted decision tree | X | ✓ | ✓ | X | 25 | ✓ |

X = not applicable, ✓ = applicable, Short < 52 weeks, Long ≥ 52 weeks irrelevant and/or redundant for representing the carbonation process, which ultimately reduces the performance of the model.

Similar to the carbonation depth prediction models, the existing machine learning based chloride concentration prediction models largely employ data generated from accelerated laboratory tests as seen in Table 4.2. These models were applied to characterize the chloride permeability of concrete. The purpose of the models was to reduce or fully substitute the rapid chloride penetration test since its experimental determination in laboratory is usually resource and time consuming. Except [135], all the related work failed to perform data optimization techniques. Unlike the existing works, the model developed in this dissertation is based on long-term field tests, taken into account 32 input parameters and performed data optimization technique. The main purpose of the developed model is to determine the influential parameters that predict the chloride profile of concrete besides making chloride profile prediction.

In both carbonation and chloride cases, the majority of the existing models have adopted neural network algorithm. Unlike these works, the developed models in this dissertation employs neural network and other learning algorithms including decision trees and ensemble methods. Examining the ability of various machine learning algorithms for prediction of the corrosion causing factors is essential in order to identify

**Table 4.2.** Data-driven models proposed for evaluating chloride penetration (directly or indirectly).

| Work | Problem type | Main learning algorithm type | Exposure environment | | Exposure duration | | No. of input parameter | Data optimization |
|---|---|---|---|---|---|---|---|---|
| | | | Lab | Field | Long | Short | | |
| [165] | Permeability | Neural network | ✓ | X | ✓ | X | 5 | X |
| [166] | Permeability | Neural network | ✓ | X | X | ✓ | 6 | X |
| [135] | Permeability | Support vector regression | ✓ | X | X | ✓ | 7 | ✓ |
| [45] | Permeability | Neural network | ✓ | X | X | ✓ | 2 - 6 | X |
| [167] | Permeability | Neural network Adaptive neuro-fuzzy inference system | ✓ | X | X | ✓ | 4 | X |
| [168] | Permeability | Neural network | ✓ | ✓ | X | ✓ | 6 | X |
| [169] | Diffusion coefficient | Neural network | ✓ | X | X | ✓ | 7 | X |
| [170] | Diffusion coefficient | Neural network | ✓ | X | X | ✓ | 4 | X |
| [171] | Diffusion coefficient | Neural network | ✓ | X | X | ✓ | 8 | X |
| [54] | Diffusion coefficient | Neural network | ✓ | X | X | ✓ | 8 | X |
| This work [172] | Influential predictors | Bagged decision tree | X | ✓ | ✓ | X | 32 | ✓ |

X = not applicable, ✓= applicable, Short < 52 weeks, Long ≥ 52 weeks the algorithm that performs best. This is due to the fact that the relative prediction power of any machine learning algorithms primarily depends on the details of the considered problems. Without experimenting, it is impossible to identify the powerful algorithms that excel a given problem.

The previous attempts to predict carbonation depth and chloride concentration using data-driven models are encouraging but a lot of improvements have to be carried out. The shortcomings of the previous works can be summarized as follows. First of all, the majority of the models have employed few data to train the models which are acquired from short-term tests. Secondly, the models missed some important input parameters that describe the carbonation and the chloride ingress process. Thirdly, most of them adopted only one type of learning algorithm. Fourthly, the few models that considered several parameters failed to perform certain crucial data optimization steps during the model development process. The performance of the proposed machine learning based carbonation depth and chloride profile prediction models can be improved considerably by addressing the above shortcomings, which is the contribution of Publications II and III of the dissertation.

The results of Publication I also demonstrated that machine learning will be a core to the next generation corrosion assessment method due to the emerging use of wireless sensors for monitoring RC structures. Today, there are several studies which show the applicability of wireless sensors for monitoring RC structures spanning from earlier-age parameters to environmental situations that can initiate or accelerate corrosion of reinforcement bar [136–140]. Monitoring of various parameters without extracting knowledge or inference from the monitored data is pointless. So, the integration of machine learning and wireless monitoring will form a principal component in the inspection, assessment, and management of RC structures. Eventually, it will bring a paradigm shift in durability assessment of RC structures. The future recommended layout of aging management method for RC structures is shown in Figure 4.1. As seen in the figure, the sensors that are integrated in the structure will continuously deliver real-time data regarding the temporal and spatial changes of the monitored parameters that control corrosion of reinforcement bar or other deterioration mechanisms. The sensors data will be transferred to a cloud storage that gives substantial benefit because data from various streams can be accessed and shared with Internet connectivity from anywhere. Such a monitoring system can be seen as a reliable nondestructive technique that provides valuable in-service data without the participation of inspection crews on the field. This approach will be more cost effective than performing periodic field testing in the long term, considering the cost of labour, the costs to the users, and their safety [141,142]. Condition assessment and prediction of the structure can be executed remotely and rapidly without the need for empirical models. The prediction enables a more realistic condition assessment of a structure and accurately timed maintenance measures, which in turn reduces the associated costs considerably. Moreover, the proposed system can learn the synergic effect of different deterioration mechanisms and discover new useful knowledge using data of several parameters monitored by various sensors. The discovered knowledge will assist engineers to come up with optimal solutions that improve the durability of RC structures as well as to define proactive maintenance plan.

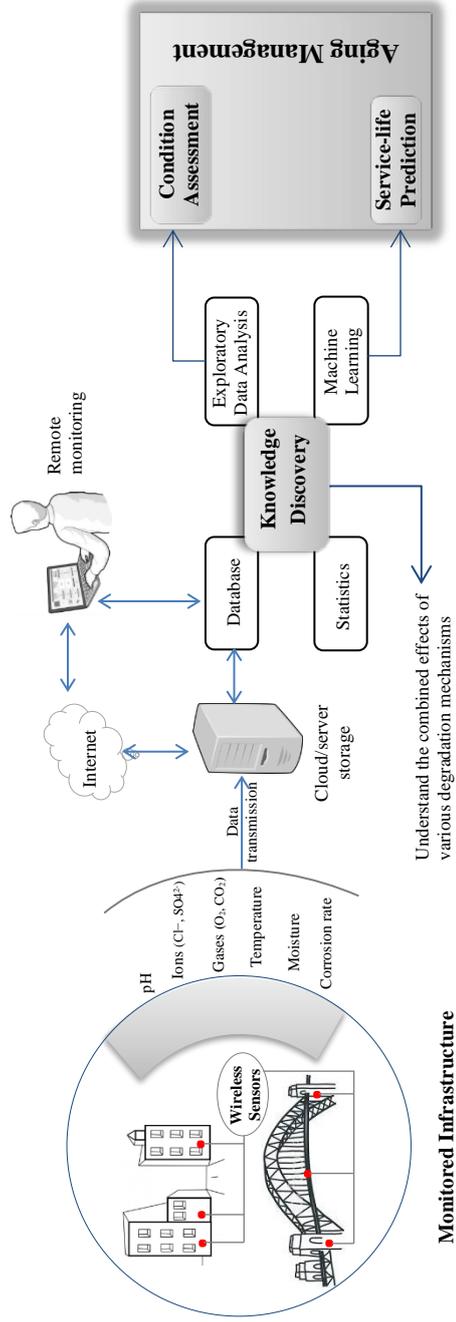

**Figure 4.1.** Overview of next generation preferred condition assessment method for RC structures (Publication I).

## 4.2 Carbonation depth prediction

The development process of the carbonation depth prediction model (CaPrM) has been discussed in Section 3.4 and in this section the performance result of CaPrM is presented. This section is arranged into three subsections. The first subsection elaborates the performance of the developed carbonation prediction model and its validity. The second subsection presents the set of influential variables that describe the carbonation depth. The comparison of CaPrM versus traditional carbonation prediction model is presented in the last subsection. These three subsections answer the research question two of the dissertation, *"how to develop accurate carbonation depth prediction model that considers the complex parameter interactions? What are the predominant carbonation depth predictors?"*.

### 4.2.1 CaPrM performance

CaPrM was trained using a dataset generated from concrete specimens that are exposed to natural carbonation for long term at field environment. The training dataset comprise information regarding the concrete mix ingredients, the fresh and hardened concrete properties, the carbonation periods, the environmental and curing conditions, as well as the carbonation depths. The training performance of CaPrM is illustrated in Figure 4.2, demonstrating the measured versus the predicted carbonation depth of all the integrated learning algorithms. The coefficient of correlation (R-values) were applied to examine the training accuracy of each learning algorithms. This parameter indicates how well the integrated learning algorithms regress the carbonation depth on the input variables. It can be observed from Figure 4.2 that the R-values of all of the four utilized learning algorithms surpass 0.90. This confirms that all the integrated machine learning methods track the real carbonation depth competently during the model training phase. Observably, neural network has attained the best learning performance (R=0.97), followed by decision tree (R=95), boosted decision tree (R=0.94), and bagged decision tree (R=0.91). It can also be noticed that, except neural network, all the learning algorithms exhibit a slight tendency to underestimate the depth of carbonation when the carbonation depth is large. The reason is the available number of large carbonation depths in the training dataset was few. For example, measurements of carbonation depth above 3 mm represent only about 9% of the total observation. The lack of sufficient number of data in the training dataset causes for the underestimation in this range.

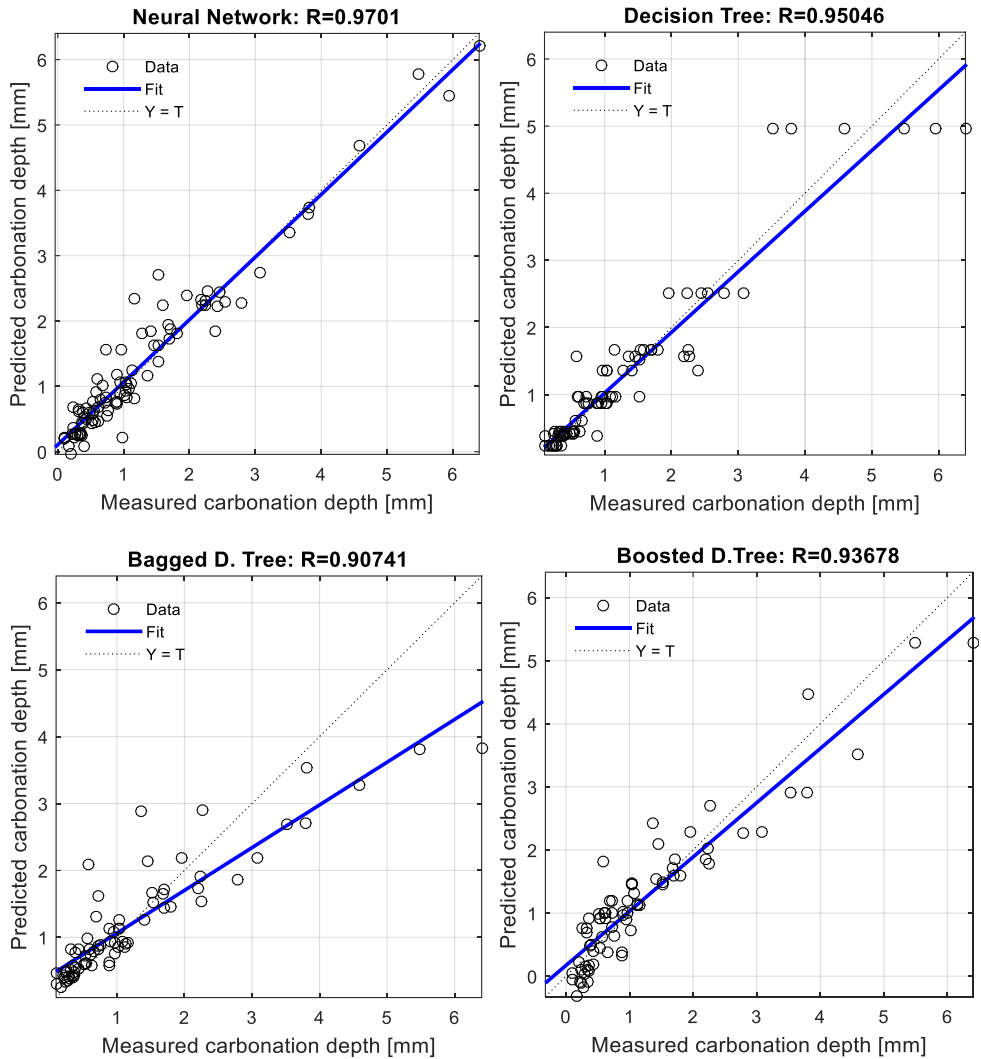

**Figure 4.2.** Training performance of the CaPrM. Y and T are predicted and measured, respectively (Publication II).

As any machine learning model, the validity of the developed model should be evaluated using a test dataset which is extracted from the original data but different from training dataset. The purpose is to test the real performance of the model since data-driven models may introduce error due to high bias and variance. High bias can cause the learning algorithm to miss the relevant association between the input and the target variables, resulting *underfitting*. High variance can cause the learning algorithm to model the random noise in the training dataset rather than the desired outputs and causing *overfitting* [82]. These can make a prediction model unstable and therefore unsuitable for solving real-world applications. Hence, the optimal data-driven models should have the best trade-off between the bias and the variance. This is often controlled by altering the model's capacity (ability to fit a wide variety of functions).

The training and test error of all the integrated learning algorithms in the CaPrM were checked to detect any overfitting and/or underfitting. Overfitting occurs when the difference between the error of training and test is far too high. Underfitting happens when the model is unable to acquire a sufficiently low error value on the training dataset. To detect overfitting and/or underfitting, the following error measurements were carried out: mean-square error (MSE), root-mean-square error (RMSE), and mean-absolute error (MAE). The MAE, also called the absolute loss, is an average of the absolute residuals/errors (the difference between the predicted and the actual value) and measured in the same units as the data. MSE is the mean of the squared difference between the target and the predicted value. It is the most widely employed loss function for regression models. RMSE is simply the square root of the MSE. Sometime RMSE is preferable than MSE because understanding of error values of MSE is difficult due to the squaring effect, particularly, if the target value represents quantities in unit of measurements. RMSE retains the original units as MAE. The MSE, RMSE, and MAE are calculated using Equations (4.1), (4.2), and (4.3), respectively.

$$MSE = \frac{1}{n}\sum_{i=1}^{n}(Y_i - \hat{Y}_i)^2, \tag{4.1}$$

$$RMSE = \sqrt{\frac{1}{N}\sum_{i=1}^{N}(Y_i - \hat{Y}_i)^2}, \tag{4.2}$$

$$MAE = \frac{1}{N}\sum_{i=1}^{N}|Y_i - \hat{Y}_i|, \tag{4.3}$$

where $\hat{Y}_i$ is the predicted output value, $Y_i$ is the measured target value, and $N$ is the number of observations.

The performance evaluation using the above statistical measures confirm that all the integrated learning algorithms predict the carbonation depth with rationally low error on previously unseen data. Average of ten round statistical performance indicators (MSE, MAE, and RMSE) of all the learning methods are given in Table 4.3. The difference between the training and the test errors of all the integrated learning algorithms is small, confirming the generalization ability of the CaPrM. The lower the statistical errors in training and testing phases are the superior the performance of the model. It can be easily noticed from Table 4.3 that the neural network algorithm outperformed all the other models. The MSE of the neural network algorithm is the lowest both in training and testing stages compared to the other integrated learning algorithms. For instance, the MSE of the boosted decision tree is increased by 9% in the model training phase while by 10% in the testing phase. This demonstrates that the neural network algorithm integrated in the CaPrM has a balanced trade-off between bias and variance errors, confirming its high learning and generalization abilities. Among decision tree based algorithms, boosted decision tree has a high generalization ability with errors at training phase (MSE=0.21, MAE=0.23 and RMSE=0.46) and testing phase (MSE=0.26, MAE=0.31, and RMSE=0.51). Though, MAE of the decision tree is slightly less than the bagged decision tree, its MSE value revealed that it has relatively weak generalization ability. Indeed, these results are valid only for the utilized particular dataset. The performance of each learning algorithms may vary if a different data were employed. This would not be a problem in CaPrM since it always provides the opportunity to select the best performing one by comparing the validation errors of the four integrated learning algorithms.

**Table 4.3.** Average of ten round statistical performance measurements for CaPrM.

| Dataset | Learning method | Training error | | | Test error | | |
|---|---|---|---|---|---|---|---|
| | | MSE | MAE | RMSE | MSE | MAE | RMSE |
| After variable selection | Neural network | 0.1895 | 0.1962 | 0.4353 | 0.2417 | 0.2860 | 0.4916 |
| | Decision tree | 0.3696 | 0.2436 | 0.6079 | 0.4189 | 0.3232 | 0.6473 |
| | Bagged decision tree | 0.2820 | 0.2498 | 0.5310 | 0.3770 | 0.3415 | 0.6140 |
| | Boosted decision tree | 0.2068 | 0.2326 | 0.4548 | 0.2649 | 0.3061 | 0.5147 |
| Before variable selection | Neural network | 0.3664 | 0.2624 | 0.6053 | 0.3522 | 0.3860 | 0.5935 |
| | Decision tree | 0.4106 | 0.3325 | 0.6408 | 0.5295 | 0.4491 | 0.7276 |
| | Bagged decision tree | 0.3998 | 0.3298 | 0.6323 | 0.4907 | 0.4391 | 0.7005 |
| | Boosted decision tree | 0.3371 | 0.3094 | 0.5806 | 0.3749 | 0.4116 | 0.6123 |

The performance evaluations discussed above are after integrating the variable selection technique. The statistical measures before incorporation of variable selection method in the model development process were also evaluated and presented in Table 4.3. The motive for evaluating results of both approaches was to demonstrate the importance of implementing variable selection methods in enhancing the prediction ability of the model. According to the statistical measures, incorporation of variable selection technique in modelling process improved the prediction ability of all the integrated learning algorithms in CaPrM considerably. It can be observed from Table 4.3 that after implementing variable selection the MSE of the neural network was reduced by a factor of about 1.5 followed by boosted decision tree with 1.4 reduction factor. Similarly, the MSE of the standalone decision tree and the bagged decision tree have decreased by 26% and 30%, respectively. This proofs that integration of variable selection methods in CaPrM modelling process leads to its prediction performance enhancement.

The test residuals (the difference between the actual and the predicted carbonation depths) of the four integrated models are computed and their distributions are visualized with boxplot in Figure 4.3. The median of the residuals is represented by a red line within the blue box that covers the middle 50% ($25^{th}$–$75^{th}$ percentiles) of the residuals. The whiskers go down to the smallest and up to the largest values. Residuals greater than 1.5 box lengths above the whiskers are outliers and designated by a red plus sign. It can be clearly seen from Figure 4.3 that the median of residuals of the neural network is about in the middle of the box and distributed around zero. In another word, the residuals have a constant variance pattern and normally distributed. This confirms that the neural network based carbonation prediction model is accurate on average for all tested data. The median of the residuals of decision tree and bagged decision tree are almost zero. The medians of bagged and boosted decision trees are closer to the first quartile than to the third quartile. This indicates that the distributions of the residuals are slightly skewed to the right. Generally, the boxplot of the residuals of a single run test validates that all the integrated models learn the nonlinear relation of the input variables and are able to predict the carbonation depth with high accuracy.

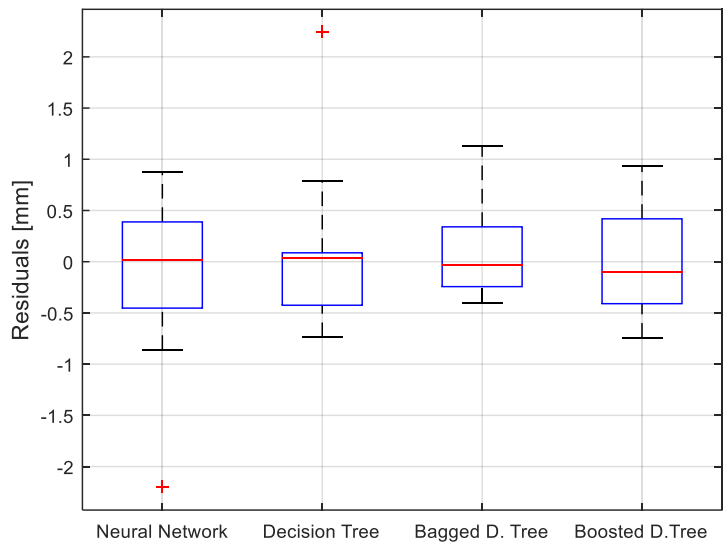

**Figure 4.3.** Boxplot of residuals of the CaPrM model (Publication II).

### 4.2.2 Carbonation depth predictors

The second part of the research question two aims to determine the predominant carbonation depth predictors. Discovering the most informative set of variables that describe the carbonation depth is vital in order to develop efficient and parsimonious carbonation depth prediction model. To determine the influential variables, the predictive power of all the utilized input parameters of the dataset was examined using the ensemble methods (bagged and boosted decision trees). These methods provide variable importance weight, which are computed by aggregating the weights over the trees in the ensemble. Variables with a higher importance score are indicative of their significance in predicting carbonation depth. The importance score of all the considered input parameters that were deduced from the bagged and boosted decision trees are shown in Figure 4.4. It can be clearly noticed from the figure that accelerated carbonation depth, w/b and compressive strength are the topmost three predicting parameters of carbonation depth for the considered data. These determined top three carbonation predictors are well-known parameters and they have already been considered in several empirical models. It is known that cement types affect the carbonation process, but it was not identified as influential parameter by the models.

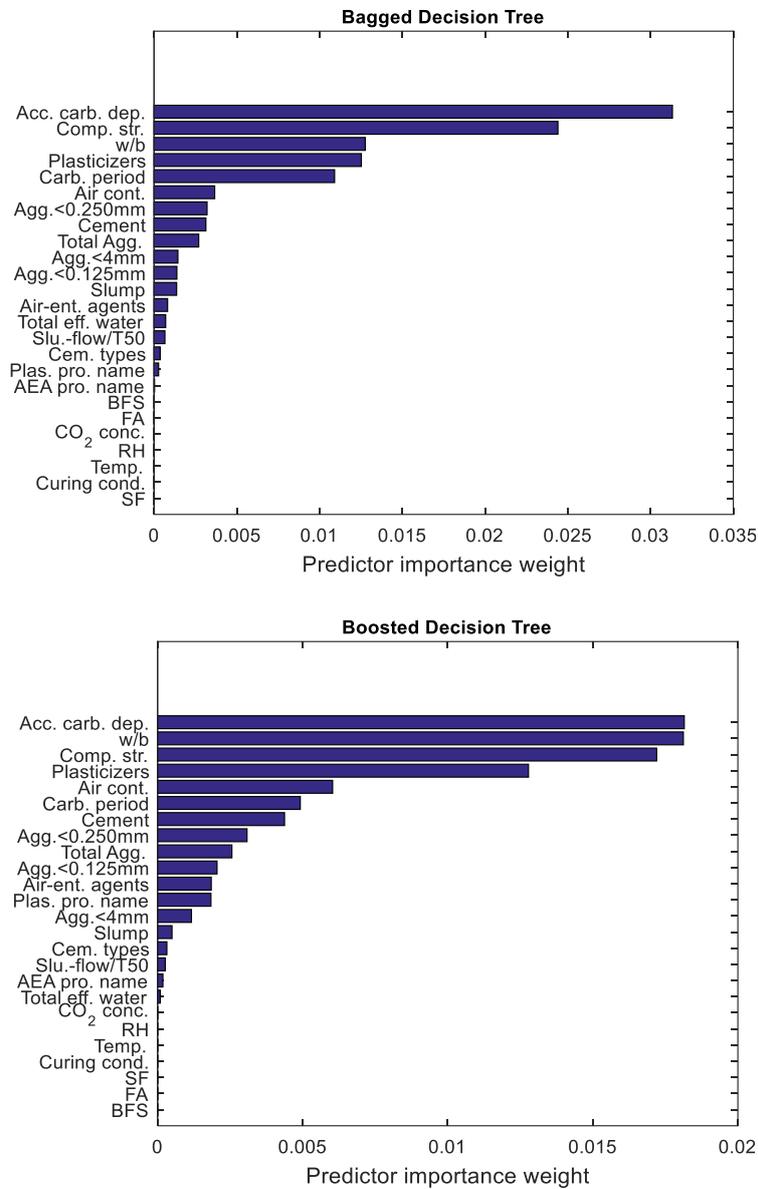

**Figure 4.4.** Measures of carbonation predictor parameters (Publication II).

This does not mean that this variable is uninformative. It only means that this variable was unpicked by the ensemble trees since other variables encode the same information. For instance, cement types are already described by both w/b and compressive strength. The compressive strength of concrete is not only affected by the w/b and cement types but also by other factors such as aggregate size distribution. Hence, compressive strength describes the carbonation process better than the combination of individual parameters that influence the strength of concrete.

Next to the top three parameters, the amount of plasticizers, the air content, and the carbonation period are the predominant carbonation depth predictors. This is an interesting finding since the identified parameters, except the carbonation period, are overlooked in various conventional models. In fact, there are few studies that demonstrate the effect of different types of plasticizers in improving the carbonation resistance of concrete through their influence on the pore structure and the morphology of the hydrated product [143,144]. Other studies show the impact of air content on carbonation process by conducting experiment on air-entrained concrete [145,146]. According to the studies, air-entraining admixtures increase the carbonation resistance of concrete. Those studies were made on concrete specimens of different mix compositions exposed to indoor and outdoor environments. Though air-entraining admixture is one of the influential factors that control the carbonation process, it was not identified as among the top predictors. This is due to the fact that the effect of the air-entraining admixtures is already explained by the air content of the concrete. So, this describes the reason for air content of the concrete being identified as among the top carbonation predictor. The significance of aggregate size distribution in predicting the carbonation depth is noticeable in Figure 4.4, although it is not among the top six descriptors. The reason is that it controls the air permeability of gases. The finding of aggregate size distribution as influential predictor is supported by earlier research [30,147]. This parameter is also missed in several conventional models.

The importance measure does not identify some known significant parameters, such as supplementary cementitious materials, curing, and environmental conditions. In case of supplementary cementitious materials, the reason is that they have already been expressed by w/b and the combination of other variables, for example, accelerated carbonation depth, comprehensive strength, and air content. The use of similar curing and environmental conditions for all concrete specimens was the reason for these parameters to be unidentified by the model. It can be observed from Figure 4.4 that the ranks provided by both learning methods, bagged and boosted decision trees, are fairly similar. Cumulatively, about 87% of the total influence is attributed by the top six significant parameters in both models.

### 4.2.3 Comparison of CaPrM and conventional models

The prediction performance of the CaPrM was compared with the conventional Fick's second law based carbonation depth prediction model. To execute a fair comparison, the natural carbonation coefficient for each concrete type was determined and employed in the conventional model given in Equation (2.6). The natural carbonation coefficient of each concrete type was computed using the carbonation depth measured at the age of 268 days. Using this coefficient in Equation (2.6), the carbonation depths at the ages of 770, 1825 and 2585 days were predicted. Then the performance of the model was evaluated by analysing the mean-square error difference between the computed and the measured carbonation depths. The MSE of the conventional model was 0.51, which is larger than the MSE of the CaPrM as can be seen from Table 4.3. The error difference is significant, about two-fold compared with the MSE of neural network and boosted decision tree. The MAE of the conventional model was 0.50, which is more than 1.7 times the MAE of the neural network of the CaPrM.

The residual distribution of the traditional model is presented in Figure 4.5 since the error statistics alone do not deliver sufficient information regarding their distribution. This figure comprises two plots (a boxplot and a histogram plot). These two plots are essential to check the normality of the residual distribution. It can be observed from the boxplot that the residuals median of the traditional carbonation prediction model is higher than any of the integrated models in the CaPrM (illustrated in Figures 4.3 and 4.5). The whisker of the conventional carbonation prediction model ranges from -1 to 1.76 which is greater than from all integrated models in the CaPrM. This demonstrated that the number of underestimate and overestimate predictions of the carbonation depth by the traditional model is considerably higher than the CaPrM. The histogram plot shows the histogram of the residuals with the best-fitting normal curve. It can be noticed that the shape of the histogram is asymmetric and skewed to the right direction. The estimated mean ($\mu$) and standard deviation ($\sigma$) of the fitted normal distribution curve is 0.30 and 0.64, respectively. Unlike the conventional model, the residual distribution of the neural network of the CaPrM is more closely resembles a symmetric normal distribution as can be seen in Figure 4.3. This shows that mean of the error is close to zero. All these facts confirmed the superiority of the CaPrM over the conventional model.

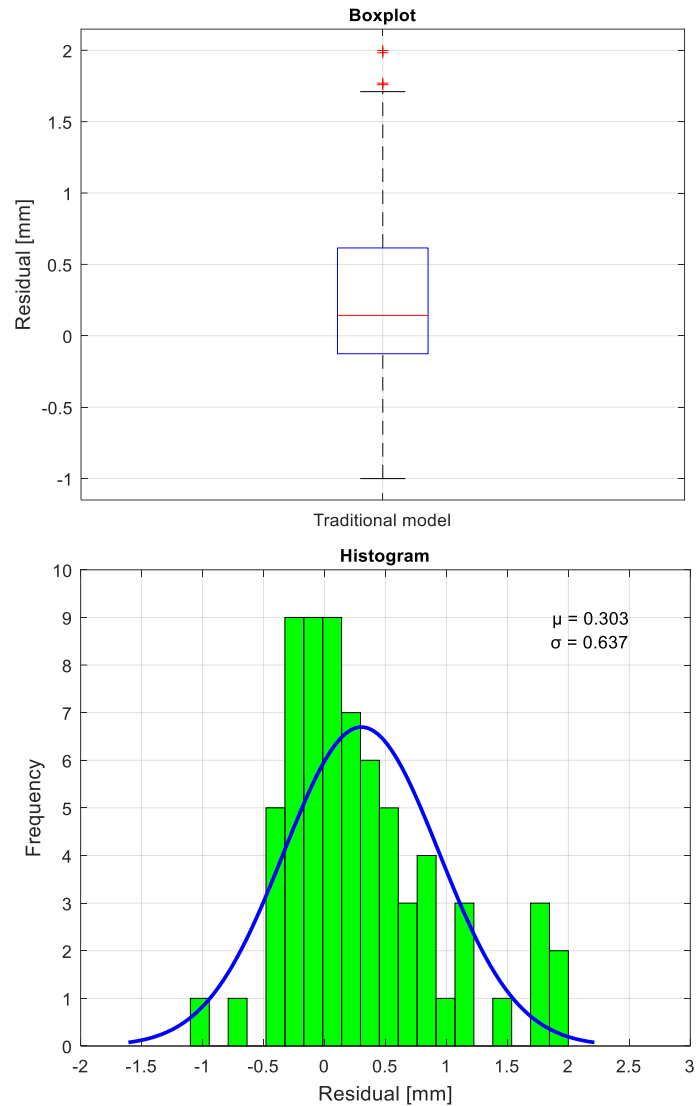

**Figure 4.5.** Residual of the conventional carbonation model: as a boxplot and histogram with a distribution fit.

Though the traditional carbonation depth prediction model based on Fick's second law of diffusion is commonly applied, it is unable to provide accurate predictions as presented above. The main reason for this is that the dependence between concrete carbonation and its conditioning factors is inherently complex and time dependent. So, it is impossible or too complex to describe them mathematically. In the presence of readily available data, machine learning algorithms can characterize the carbonation process very well since it has the ability to learn the complex interrelation among the governing factors. Unlike the conventional model, machine learning based models are able to discover patterns that never been observed before.

## 4.3 Significance of chloride penetration controlling parameters

The third publication answers the research question three of the thesis, *"what are the significant parameters that describe the chloride concentration into concrete?"*. The influential parameters that describe the chloride profile were determined and evaluated using the developed chloride profile prediction model that was discussed in Section 3.5. The model is developed using bagging decision trees and employed long-term field experimental data that are acquired from the Finnish DuraInt-project. Ten models were established by utilizing diverse input dataset. In order to examine the importance of fresh and hardened concrete tests in predicting the chloride profile, the ten models were categorized into two groups: Model A and Model B. That is, Model A utilizes all the input variables whereas Model B excludes the fresh and hardened concrete test variables except chloride migration coefficient, $D_{nssm}$. The reason for keeping $D_{nssm}$ in the dataset is that this parameter is considered as one of the best indicators of the resistance of the concrete to chloride ion ingression. Each group was further divided into three scenarios employing: i) all the input variables, ii) all the input variables except distance from highway, and iii) all the input variables except exposure time. The model classification details can be referred from Table 3.6.

### 4.3.1 Chloride profile predictors using all variables

In this scenario, the training dataset consists of all the respective input parameters of the Models A and B. The parameter importance measures for both Models (A.1 and B.1) were performed and shown in Figure 4.6. It can be clearly seen from Figure 4.6 that the distance from highway is the primary parameter which controls the chloride profile in both models. This is the anticipated outcome since the quantity of chlorides splashed on the concrete surfaces heavily depends on the distance between the highway lane and the specimens. The nearer the distance the higher is the amount of the splashed chlorides. Next to distance from highway, compressive strength, cement content, total effective water, binder types, and exposure time are the five influential predictors in Model A.1. Even if this model comprises numerous parameters from fresh and hardened concrete properties, the discovered influential chloride profile predictors are from concrete mix ingredients. This demonstrated that the influence of advanced laboratory tests performed at early age is impotent in predicting the chloride profile holistically.

In the case of Model B.1, the significant parameters next to distance from highway are cement content, total effective water, binder types, exposure time, and w/b ratio. In both models, supplementary cementitious materials have the lowest influence in predicting the chloride concentration in concrete. Indeed, it is renowned that supplementary cementitious materials are mainly utilized to boost the resistance of concrete against chloride permeability because they can reduce the size of large pores and capillaries. However, they were not recognized as a significant chloride profile predictor since their types and quantities are already described by the binder types and w/b, respectively. The type of air-entraining agents is also appeared to be powerless for predicting the chloride concentration in concrete as confirmed by the results of Model A.1 and B.1. This parameter was classified based on their production name (due to the same reason mentioned in Section 3.4.1), and thus it appeared as "AEA pro. name" in Figure 4.6. Though, the types of chemical admixtures generally influence the pore structure of concrete, the finding of this study confirmed that the air-entraining agent types (unlike type of plasticizers) do not affect the chloride permeability property of concrete. The utilized air-entraining agents in this dissertation were produced from either fatty acid soap or vinsol resin. So, it is worth to mention that this finding is only valid for the employed type of air-entraining agents, mix composition and exposure conditions. Other types of air-entraining agents may behave differently in concrete specimens exposed to the same or different environments.

In this scenario, the chloride migration coefficient ($D_{nssm}$) is recognized as a trivial parameter in predicting the chloride profile. The reason for this is that the chloride transport properties rely on the intrinsic permeability of the concrete, which is varying with time throughout the cement hydration process as well as based on the amount of chloride concentration in the pore solution. Additionally, the incorporation of chloride profiles quantified from specimens composed of various binder types and located at various distances with different exposure times diminish the significance of $D_{nssm}$ in predicting concentration of chloride in concrete. Though $D_{nssm}$ is the most widely adopted parameter to evaluate the resistance of concrete against chloride ingress, discovering it as an insignificant predictor is not a new phenomenon. There is a study based on long-term experiment that corroborates the inability of $D_{nssm}$ obtained from accelerated lab test in characterizing the long-term chloride permeability of concrete exposed to real environmental conditions [148]. The power of $D_{nssm}$ in describing the resistance of concrete against chloride ingress is evaluated in Section 4.3.2 and 4.3.3 by excluding the parameter distance from highway and exposure time, respectively.

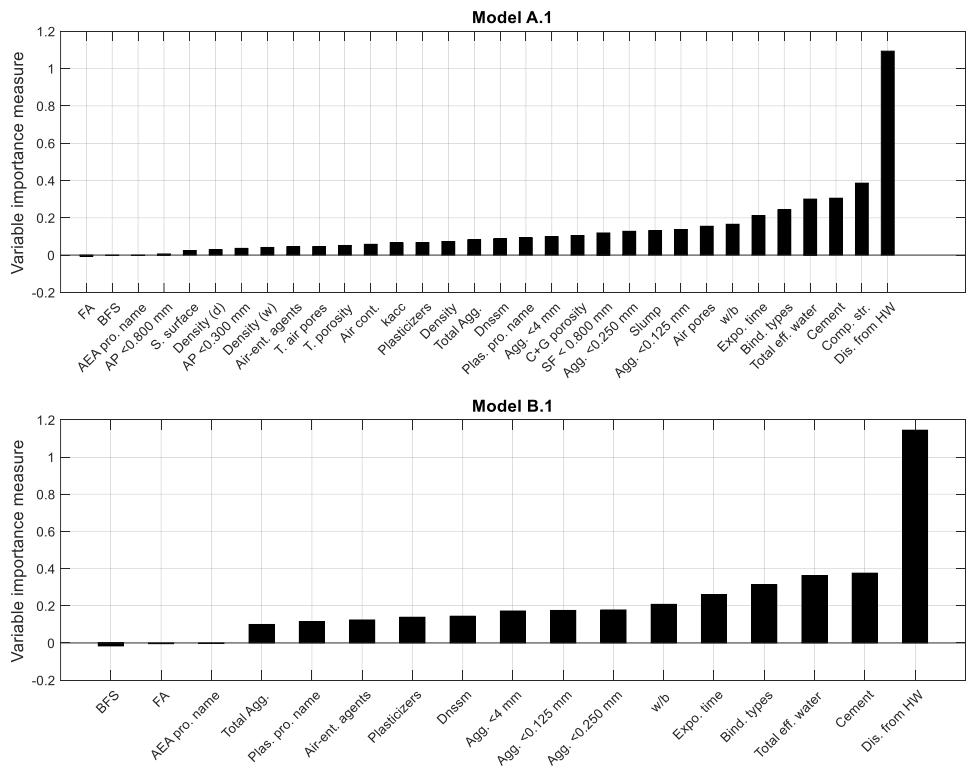

**Figure 4.6.** Variable importance measures of chloride profile for scenario one (Publication III).

### 4.3.2 Chloride profile predictors at fixed distance

In order to examine the significance of parameters independent of specimen's distance from highway, the chloride profile measured only at 4.5 m and all the respective input parameters of this scenario were employed in the dataset of Models A.2 and B.2. The choice of this distance is due to the presence of higher number of specimens at this distance, acquiring more chloride profile in the dataset. The top six most significant chloride profile predictors determined in this scenario are illustrated in Figure 4.7. It can be observed that the exposure time is the foremost powerful predictor of chloride concentration in concrete for Models A.2 and B.2 with importance measure of 0.43 and 0.44, respectively. The discovered six topmost significant chloride profile predictors have about 77% and 85% of the collective contributions to the ensemble Models A.2 and B.2, respectively. These are substantial percentage of contributions given the fact that the number of utilized input parameters for Model A.2 was 31 whereas for Model B.2 was 16. The variable importance measures of each corresponding parameters of the two models are comparable as can be seen from their rank in Figure 4.7. The importance variable measures of the top six significant chloride profile predictors with their percentile contribution for both models are presented in Table 4.4. As it can be noticed from Table 4.4, the exposure time has a relative contribution of 40.8% in Model B.2. The w/b parameter with a 12.3% contribution is the second leading variable. Aggregate <0.125 mm is the third predominant parameter that had 10.4% contributions to the chloride profile prediction. Cumulatively, 63.5% of the total influence is attributed by these three significant parameters. The $D_{nssm}$, plasticizers type, and aggregate <4 mm parameters are the next significant variables which represent 9.1%, 7.2% and 5.6% of contributions in predicting the amount of chloride concentration in concrete, respectively.

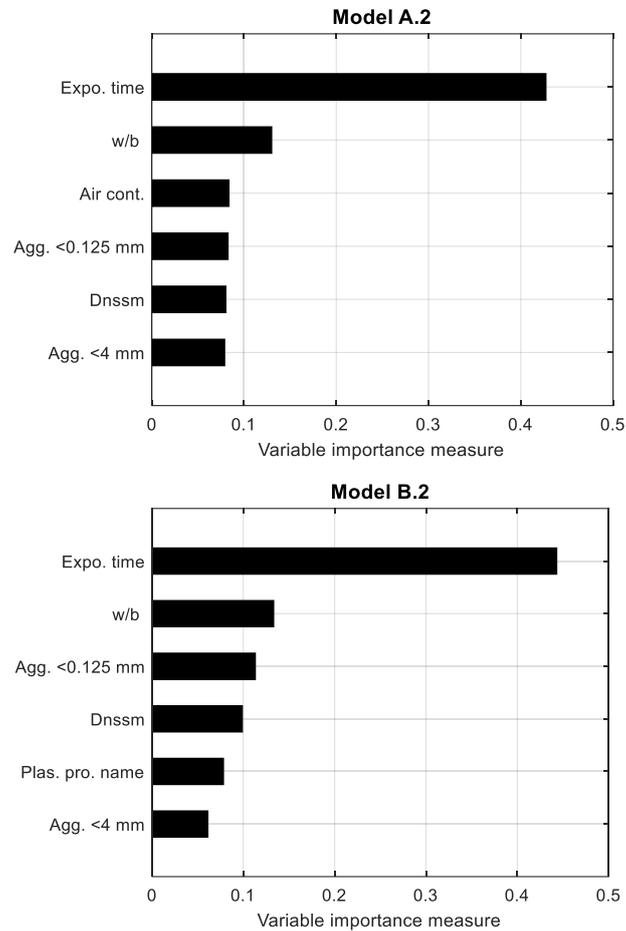

**Figure 4.7.** Variable importance measures of chloride profile for scenario two (Publication III).

Unlike scenario one, the variable importance analysis demonstrated that the chloride migration coefficient ($D_{nssm}$) is among the six significant

**Table 4.4.** Importance measure of influential variables and their percentile contribution for Models A.2 and B.2 (Publication III).

| Model A.2 | | | Model B.2 | | |
|---|---|---|---|---|---|
| Variable name | VI measures [-] | Contribution [%]* | Variable name | VI measures [-] | Contribution [%]* |
| Expo. Time | 0.4273 | 37.62 | Expo. time | 0.4431 | 40.78 |
| w/b | 0.1300 | 11.45 | w/b | 0.1333 | 12.27 |
| Air cont. | 0.0837 | 7.37 | Agg. <0.125 mm | 0.1133 | 10.43 |
| Agg. <0.125 mm | 0.0828 | 7.29 | $D_{nssm}$ | 0.0991 | 9.12 |
| $D_{nssm}$ | 0.0804 | 7.08 | Plas. pro. Name | 0.0784 | 7.22 |
| Agg. <4 mm | 0.0790 | 6.96 | Agg. <4 mm | 0.0613 | 5.64 |

*compared with the total contributions of all input variables utilized in respective models.

chloride concentration predictors in both models. The most probable reason for this is that the distance from highway lane is fixed, and thus the amount of the splashed surface chloride is identical for all the specimens included in the models. Despite the same amount of splashed surface chloride, the amount of chloride concentrations in the concrete specimens varies because of the continuous chemical reaction of chlorides with the dilute cement solution. Thus, discovering the $D_{nssm}$ as a significant predictor confirms that it is a function of the amount of chloride at the concrete surface. The other laboratory test which is identified as a powerful chloride profile predictor by Model A.2 is the air content. Discovering the air content as a potential chloride profile predictor by Model A.2 out of the considered 16 types of fresh and hardened concrete tests is an interesting finding. The air content test was carried out on fresh concrete using pressure method in accordance with EN 12350-7 [149]. All the concrete specimens considered in this study employ air-entraining admixtures. The discovery of the air content as a powerful predictor indirectly represents the air-entraining admixtures since the air content is largely controlled by this admixture. Though the purpose of this admixture is to improve the resistance of the concrete against frost attack, it also influences the chloride ingress into the concrete. Even if very little research has been performed on understanding the effect of air-entraining admixtures on chloride transport, there is a study which demonstrates their power in controlling the transport processes of gases and ions into the concrete [150]. The finding of only air content as influential predictor out of the considered 16 types of fresh and hardened concrete tests demonstrates that several advanced laboratory tests executed at early age are insignificant in predicting the chloride concentration in concrete in this scenario.

Among the parameters that describe the concrete mix ingredients, w/b and aggregate size distribution appeared to be the leading influential predictors of chloride penetration into the concrete pores. The reason for this is that w/b and aggregates govern the pore structure of the cement paste and thus influence the chloride ion transport properties. It is a well-known fact that chloride transportation in concrete through aggregates is trivial since aggregates that are used to produce concrete are generally dense. Nonetheless, at the interfacial transition zone (ITZ) where the cement paste in the vicinity of aggregate surface exhibits lower cement content and higher porosity compared with the cement paste in areas far away from the aggregate. The ITZ covers a considerable portion of the total cement paste volume and governed by the aggregate size distribution [16,151]. This may explain why aggregate size distribution influences the chloride transport property of concrete. Parameters that describe the concrete mix ingredients such as the type of cement, the supplementary cementitious materials, and the chemical admixtures are not discovered as powerful predictors by Model A.2. This is due to the fact that their effect is already explained by the air content since this test is predominantly governed by the type and amount of cement, the supplementary cementitious materials, and the chemical admixtures.

It can be observed that the type of plasticizers is determined as a significant predictor by Model B.2. This model considers only parameters from concrete mix ingredients and $D_{nssm}$. The identified parameter (type of plasticizers) is appeared as "plas. pro. name" in Figure 4.7 since product name of concrete admixtures were applied to classify them. The plasticizers applied in the concrete mix were based on Polycarboxylate, Melaminsulfonate or Polycarboxsylate Ether. The discovery of type of plasticizers as a predominant chloride profile predictor is an interesting finding. This parameter is not considered in the conventional chloride concertation prediction models. Indeed, there are several studies which demonstrated that plasticizers alter the pore characteristics of the hardened concrete. But the effect of type of plasticizers on chloride permeability is still insufficiently studied. There is a study based on short-term test that confirm the same finding [152].

### 4.3.3 Chloride profile predictors at three different ages

The intention of this scenario is to examine the effect of the input parameters in characterizing the chloride profile at particular exposure times. The determined top six parameters that describe well the penetration of chloride into concrete at various years of exposure are illustrated in Figure 4.8. Similar to scenario one, distance from highway is the leading significant parameter in both models, Model A.3 and B.3. In fact, this is the anticipated result as the chloride profile is highly dependent on the amount of chloride at the concrete surface. Next to distance from highway, the compressive strength, the total effective water and the cement content are the three powerful chloride profile predictors identified by Model A.3 at one, three, and six years of exposure. It is well understood that the chloride permeability of concrete depends predominantly on the porosity and interconnectivity of the pore system in the concrete. The above identified three parameters directly/indirectly govern the porosity and interconnectivity of the pore system, and thus their selection as chloride profile indicators are evident. The amount of the total effective water and cement are also identified as powerful predictors of chloride profile by Model B.3 at all age groups. The influence of these two parameters in describing the chloride permeability is recognized by numerous studies and used in conventional models. They are commonly represented as water-to-cement ratio (w/c). In all models, the w/b ratio was employed as one of the training input parameters but not as w/c since different type of supplementary cementitious materials, including blast-furnace slag and fly ash were utilized to produce some of the concrete specimens. Identifying the w/b as one of the influential predictors indirectly demonstrates that the supplementary cementitious materials are powerful in predicting the chloride concentration in concrete. As the number of occurrences of blast-furnace slag and fly ash are limited in the original dataset, the model recognized the cement amount and the total effective water as more powerful chloride profile predictors than the w/b.

It can be noticed from Model A.3 (i) that the amount of plasticizers and the binder types are among the influential chloride profile predictors for concrete specimens. When the exposure time is increased from one to three and six years, the power of plasticizers as a predictor to evaluate the chloride resistance of concrete has vanished. In case of Model A.3 (ii) and A.3 (iii), air pores as well as capillary and gel porosity from concrete properties are determined as significant predictors. Though these two models have considered 18 types of fresh and hardened concrete properties, only those that describe the strength and the air void of the hardened concrete were discovered as influential chloride profile predictors. The chloride migration coefficient which is commonly used as an indicator of chloride permeability of concrete is not among the identified influential predictors.

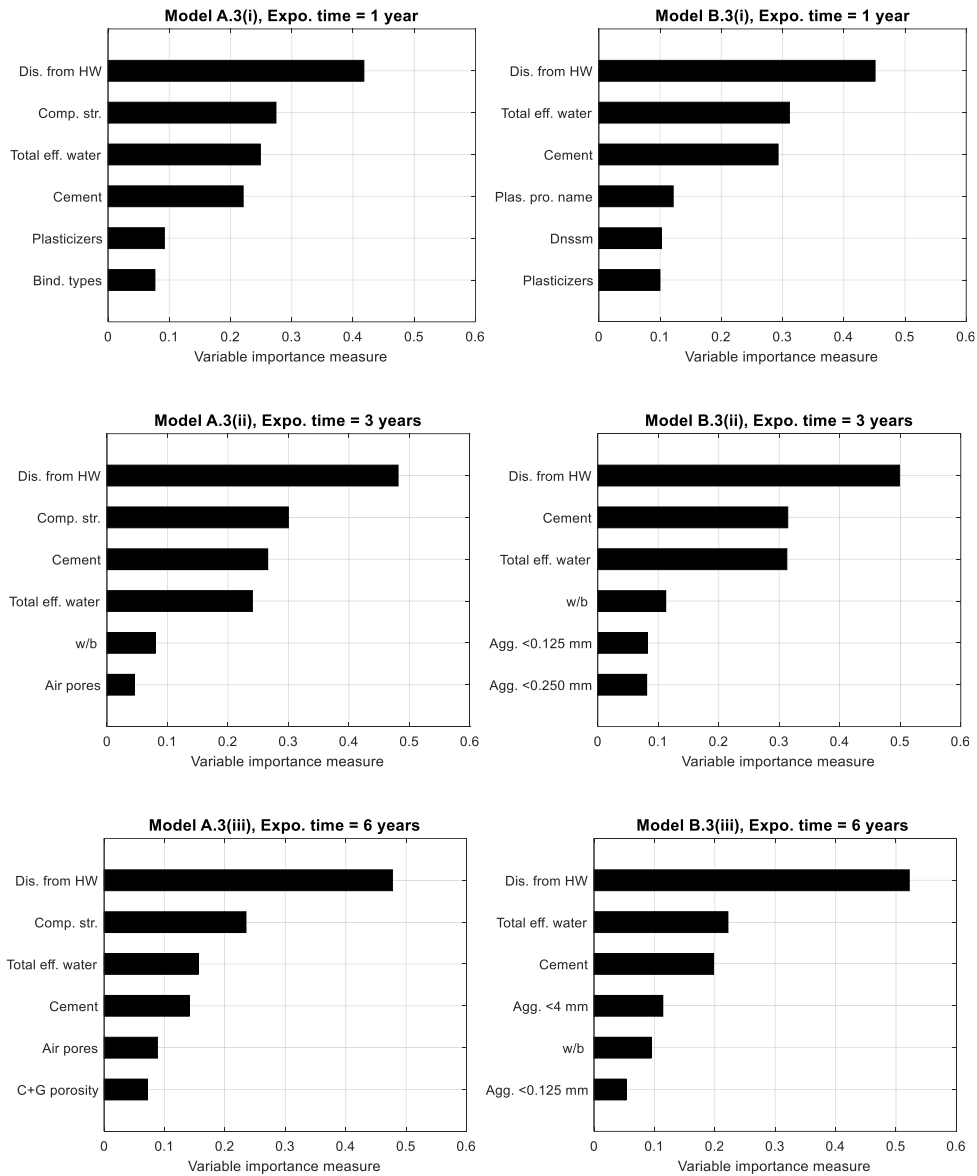

**Figure 4.8.** Variable importance measures of chloride profile for scenario three (Publication III).

The variable importance measure of Model B.3 revealed that the amount and type of plasticizers and the $D_{nssm}$ are among the influential descriptor of the chloride profile at one-year exposure than older ages. The supplementary cementitious materials appeared to be more informative for describing chloride ingress into concrete pores at three and six years of exposure but not at earlier age. The utilization of supplementary cementitious materials in concrete alters the kinetics of hydration. This change modifies the microstructure of the concrete, and thus changing its long-term durability properties [153–155]. This reason explains why the supplementary cementitious materials characterize the chloride concentration at older ages. As in supplementary cementitious materials, the importance of the aggregate size distribution in predicting the chloride profile also appeared to be less significant at the age of one year. For example, three predictors representing aggregate size distribution (Agg. <0.125 mm, Agg. <0.250 mm, and Agg. <4 mm) got importance ranks of $11^{th}$, $12^{th}$ and $13^{th}$, respectively in Model B.3(i). Whereas they got $5^{th}$, $6^{th}$ and $9^{th}$ in Model B.3(ii), and $4^{th}$, $6^{th}$ and $8^{th}$ in Model B.3(iii). Some of the parameters are not shown in Figure 4.8 since it consists only the top six predictors due to readability reason. These three aggregate size distributions contribute about 7%, 13%, and 15% in predicting the chloride penetration at the age of one, three, and six years, respectively. This can be described by the alteration of the ITZ properties over time.

The significance measure of the aggregate size distribution and the aggregate content on concrete exposed to deicing environment at specific exposure times that are determined by the developed model is illustrated in Figure 4.9. In Model A.3 (utilizes all input parameters), it can be observed that the contribution of the aggregate size distribution in describing the ingression of chloride ions into concrete has increased with exposure time. In Model B.3 (employs all parameters from concrete mix ingredients, distance from highway and $D_{nssm}$), the influence of the aggregate size distribution is insignificant at earlier age than exposure time of three and six years. However, the reverse phenomenon has been observed in case of aggregate content in both models with even negative importance measure at later age. This demonstrates that after some years of exposure, aggregate content does not have a predictive power to determine the chloride concentration in concrete. There are studies that concluded the aggregate content influences the chloride penetration into concrete but not the aggregate size distribution [16,17]. But they come up with this finding by validating the model with mortars exposed to chloride for short term, three weeks and 15 months. The evaluation at one-year exposure time in Figure 4.9 also shows aggregate content influences the chloride penetration more than the aggregate size distribution, but it is the opposite at three and six years of exposure. In addition to the aggregate size distribution, findings of Publication III revealed that utilization of supplementary cementitious materials in concrete production plays a role in characterizing the chloride penetration into concrete at later age. Hence, all these confirm that generalizing results that are acquired from short-term laboratory/field examination is inappropriate.

In another perspective, the findings of this scenario confirmed that $D_{nssm}$ derived from the chloride profile at early age is not always describing the chloride profile of concrete. Though this property is widely applied as a durability indicator, it appeared to be impotent to describe the chloride permeability of concrete in the case of Model A of this scenario where a range of fresh and hardened concrete properties were considered for training the model. But the $D_{nssm}$ becomes a significant descriptor at the age of one year in the case of Model B of this scenario where only the concrete mix ingredients and the $D_{nssm}$ were considered for the model training. This proofs that best indicators of chloride permeability of concrete determined from short-term experiment are not always powerful predictors. Similar to scenario two, the aggregate size distribution, the amount and type of plasticizers and the supplementary cementitious materials were discovered as best indicators of chloride profile of concrete. These parameters are missed in the conventional chloride concentration prediction models.

Overall, distance from highway, compressive strength, total effective water and cement content are the top four best predictors of chloride profile in all age groups identified by Model A.3. These parameters, except compressive strength, are also among the top three influential predictors of chloride ingress in all age groups in case of Model B.3. The remaining significant predictors identified by each model are varying at each exposure time. Indeed, it is well known that transport of chloride ions into concrete pores decreases with exposure time due to the change in microstructure. Even depending on the exposure condition, the chloride ions may leach out from the concrete pores, which in turn lessen the chloride concentration. For instance, 45% of the cases utilized in this scenario showed that the chloride profiles measured at one year of exposure is larger than those after three years of exposure. Similarly, 39% of the concretes exposed to six years had lower chloride concentration than those exposed only for three years. Even, 35% of the cases revealed that the chloride profile after six years of exposure is lower than those after only one year of exposure. This may explain why some of the discovered predictors are varying at different exposure time and the complexity of chloride profile predictions.

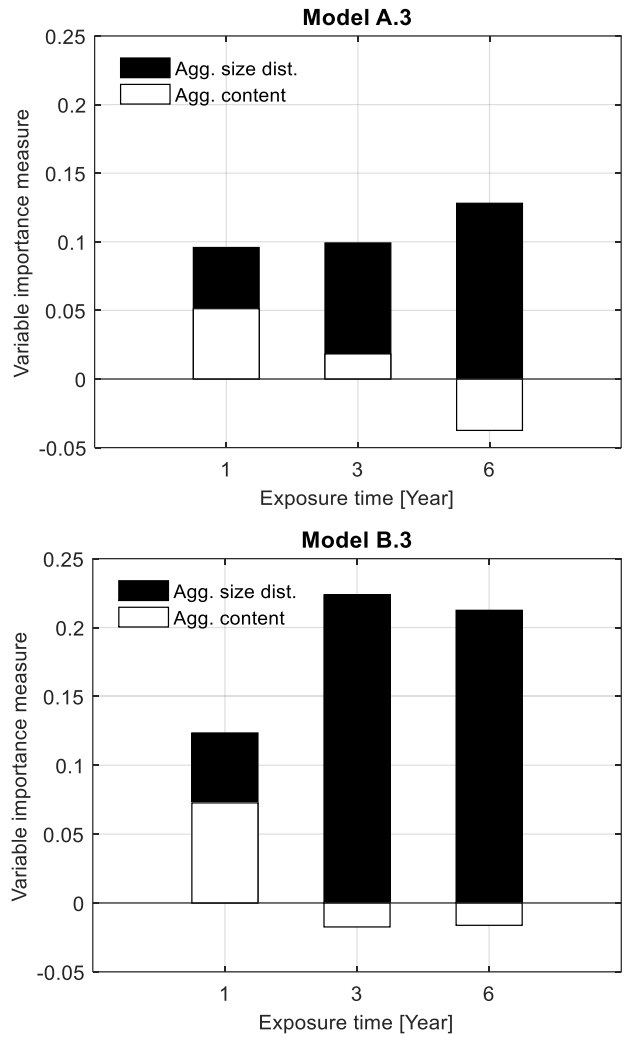

**Figure 4.9.** Significance measure of aggregate size distribution and aggregate content with respect to exposure time in Models A.3 and B.3 (Publication III).

### 4.3.4 Evaluating prediction power of the influential parameters

The contribution of the determined top six influential parameters in predicting the chloride penetration in all the above three scenarios was evaluated using unseen data. The evaluation was carried out using the respective models of the scenarios by employing three input data categories. In the first category, the inputs were all parameters presented in Table 3.5. The second category entails the top six significant parameters that were determined by the ensemble method in their respective models. In the third category, the most widely utilized parameters in the classical models (cement content, total effective water, chloride migration coefficient, exposure time and distance from highway lane) were employed. The contributions of the parameters considered in the three categories were evaluated using the MSE of the ten models. The MSEs of the models for all the three categories are given in Table 4.5. These MSE values are average of ten iterations. It can be recognized from Table 4.5 that the MSEs of all models which consider the top six significant parameters (second category) is smaller compared with the first category. The decrement in MSEs is noticeably large except in case of Models A.2 and B.2. This result demonstrates that the use of several parameters does not warrant accurate predictions. It is also recognized that the MSEs of all models with top six influential parameters is smaller than the MSEs of the third category except Models A.2 and A.3(i). For example, 19% reduction in MSE is perceived in Model B.3(iii) of the second category compared to that of the third category. In the same model, the second category achieved 34% less MSE than the first category. This proofs that data-driven models that take into account the influential parameters have the ability to provide a more accurate chloride-profile prediction.

**Table 4.5.** Chloride profile prediction models' error on the three categories (Publication III).

| Model names | Mean-square error | | |
|---|---|---|---|
| | Category 1 | Category 2 | Category 3 |
| A.1 | 1.43E-04 | 1.21E-04 | 1.37E-04 |
| B.1 | 1.37E-04 | 1.18E-04 | 1.35E-04 |
| A.2 | 1.70E-04 | 1.64E-04 | 1.58E-04 |
| B.2 | 1.60E-04 | 1.52E-04 | 1.59E-04 |
| A.3(i) | 8.66E-05 | 7.29E-05 | 6.93E-05 |
| A.3(ii) | 2.31E-04 | 1.62E-04 | 1.67E-04 |
| A.3(iii) | 4.10E-04 | 3.01E-04 | 3.36E-04 |
| B.3(i) | 8.66E-05 | 7.26E-05 | 7.80E-05 |
| B.3(ii) | 2.30E-04 | 1.62E-04 | 1.69E-04 |
| B.3(iii) | 3.75E-04 | 2.80E-04 | 3.33E-04 |

## 4.4 Hygrothermal behaviour prediction

In this section, the results of Publication IV which answers the research question four, *"how to predict the hygrothermal interaction inside surface-protected concrete while identifying the appropriate surface-protection system?"*, are discussed. As presented in Section 3.6, the surface-protection systems were applied on the five outermost layer of the façade elements of the case structure and labelled as S1, S2, S4, S5, and S6. S1 and S2 are coated with cementitious materials, whereas S4, S5, and S6 are treated with organic coating materials that are obtained from different manufacturers.

The hygrothermal prediction model was developed by utilizing two years of the monitored data. The training performance of the model for predicting the relative humidity and temperature inside all surface-protected concrete façade members are illustrated in Figure 4.10 and Figure 4.11, respectively. It is clearly seen from Figure 4.10 that the performance of the applied surface treatments in controlling the relative humidity is varying. The relative humidity in S1 and S2 is largely above 90% and 80% respectively, whereas in S5 and S6 it is above 70%. The relative humidity in S4 is mostly in the range between 60% and 100%. The temperature interaction in all façade members are almost in the same range as seen in Figure 4.11. It can also be observed that the correlation coefficients (R-values) for the prediction of the inner relative humidity and temperature for all façade elements are close to one. This demonstrates that the developed model, during the training phase, track effectively the actual measured relative humidity and temperature inside the surface-protected concrete façade members.

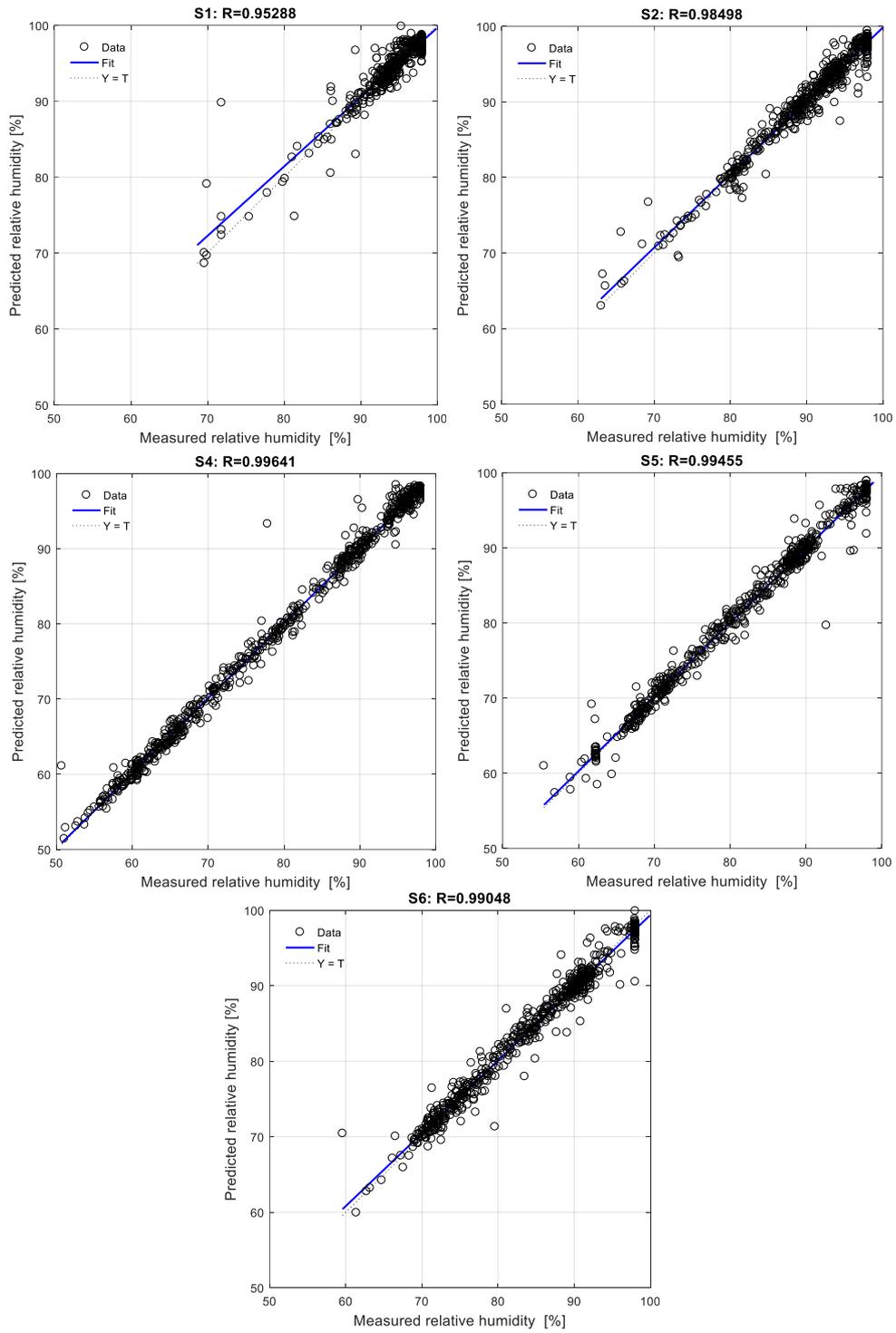

**Figure 4.10.** Training performance of hygrothermal model: inner relative humidity. Y and T are predicted and measured, respectively.

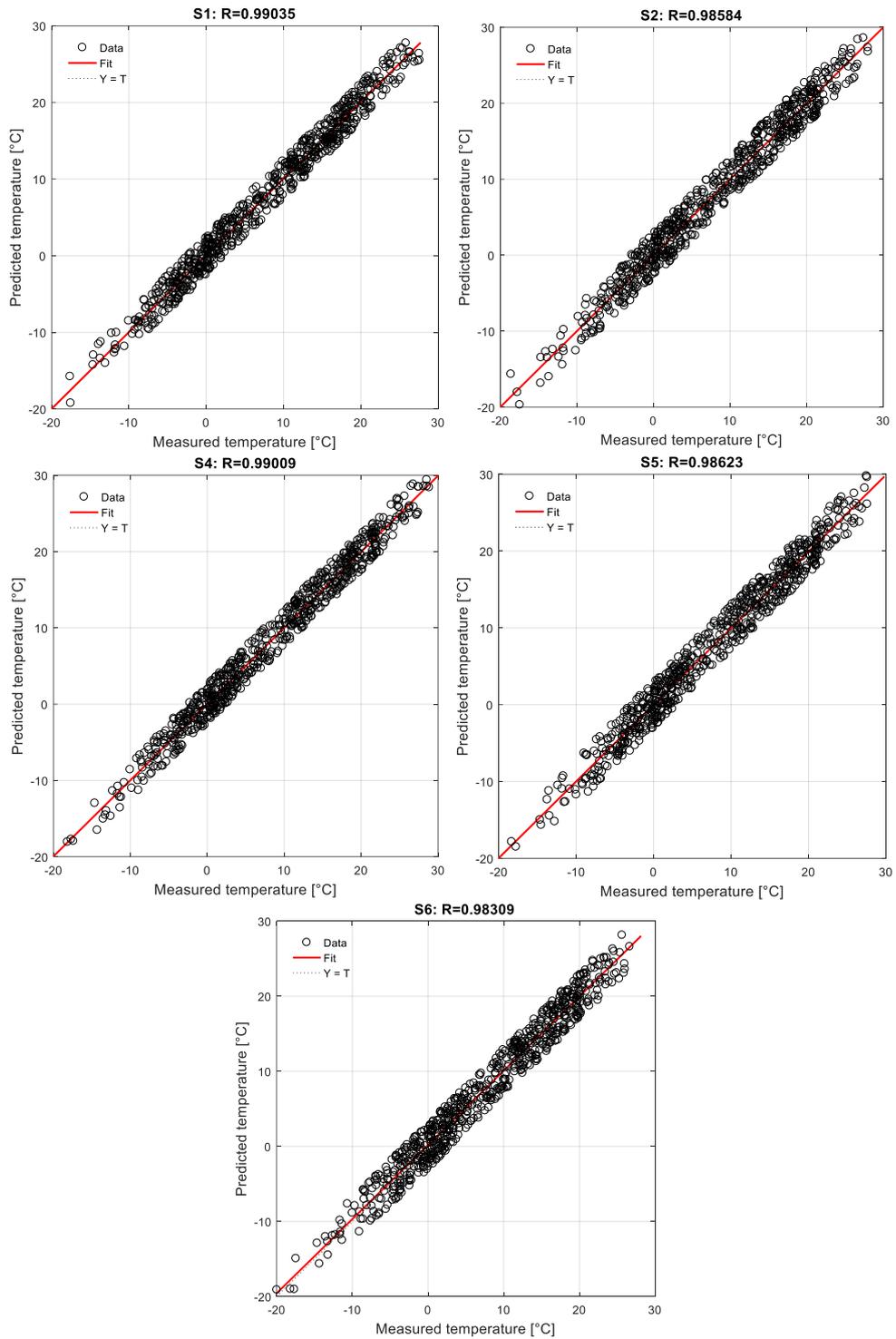

**Figure 4.11.** Training performance of hygrothermal model: inner temperature. Y and T are predicted and measured, respectively.

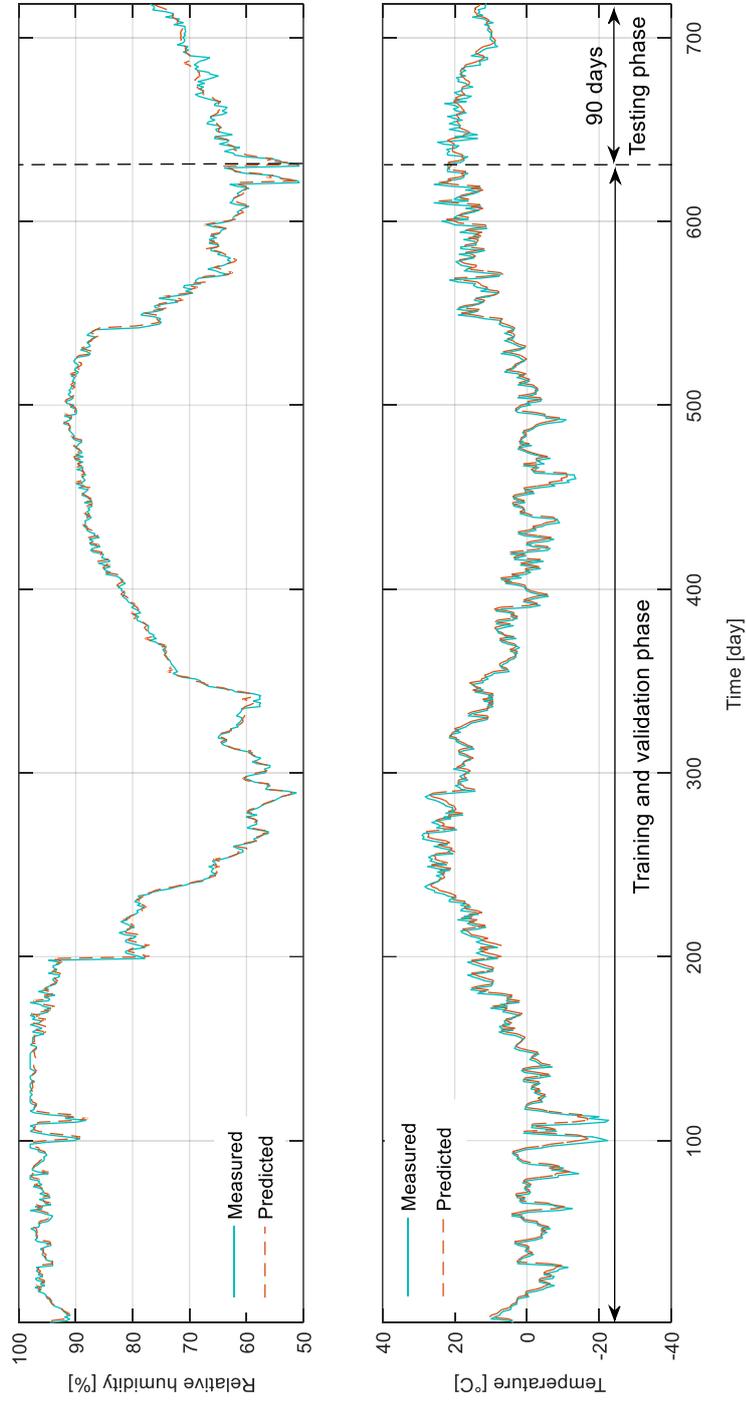

**Figure 4.12.** Predicted hygrothermal performance of the façade element S4 (Publication IV).

The validity of any successfully trained machine learning based model is evaluated based on their generalization capability. Such a model should be able to perform well when it is presented with unseen data within the range of the input parameters that are utilized during the model training phase. The performance of the developed hygrothermal models in predicting the inner hygrothermal behaviour were tested using the unseen datasets of the ambient relative humidity and temperature which covers the measurements of the last 90 days. The measured and the predicted relative humidity and temperature in concrete façade element S4 is illustrated in Figure 4.12. It can be seen that the predicted relative humidity and temperature are almost similar with the actual ones. The MSE of the predicted inner relative humidity and temperature were 7.89 and 2.44, respectively. The MAE of the predicted relative humidity and temperature were 4.04 and 0.89, respectively. As the MAE value has the same unit as the data, the average errors of the relative humidity and the temperature are about ±4% and ±0.89 °C, respectively. The MSEs and MAEs of the temperature are smaller than the relative humidity since the ambient temperature does not fluctuate extensively unlike the ambient relative humidity. Overall, the test results confirmed that the developed model predicts the hygrothermal behaviour inside surface-protected concrete façade element with rationally low error despite some missing data in the training dataset.

The test errors of the developed hygrothermal models for each concrete façade elements are presented in Table 4.6. These values are the average of ten statistical measures for each façade element. Although the model development principles are identical, each model was optimized for each façade elements because their hygrothermal behaviours are dissimilar due to the difference in the applied surface-protection materials and methods. The lower the value of the error statistics (MSE, RMSE, and MAE) is the better the prediction accuracy of the model. The small MSE values of the predicted relative humidity and temperature confirm that the developed model has high generalization capability. It can be observed from Table 4.6 that the MSE, MAE and RMSE of the temperature are smaller compared to their corresponding values of the relative humidity. The MSE and RMSE of the temperature are almost equal for all façade members. Though the error measures for each façade elements are different, the error values are generally low which proof the suitability of the developed model for examining the hygrothermal performance in surface-protected concrete façade elements.

**Table 4.6.** Statistical performance measurement of the developed hygrothermal prediction model on test dataset (Publication IV).

| Façade element | MSE | | MAE | | RMSE | |
|---|---|---|---|---|---|---|
| | Relative humidity | Temperature | Relative humidity | Temperature | Relative humidity | Temperature |
| S1 | 3.5074 | 2.8867 | 5.0236 | 0.7265 | 1.8728 | 1.6990 |
| S2 | 4.3062 | 2.7960 | 4.4686 | 0.7756 | 2.0751 | 1.6721 |
| S4 | 5.8039 | 2.8320 | 3.9995 | 0.8875 | 2.4091 | 1.6829 |
| S5 | 4.0053 | 2.5812 | 4.1923 | 0.8125 | 2.0013 | 1.6066 |
| S6 | 3.9371 | 2.5936 | 4.4386 | 0.6289 | 1.9842 | 1.6105 |

The developed data-driven hygrothermal prediction models were able to learn the hygrothermal interrelation inside surface-protected concrete façade elements using data obtained from sensors and perform fairly accurate prediction without the need for other mathematical solutions. The prediction allows to perform a pragmatic evaluation of the case structure in order to make rationally accurate schedule for maintenance measures, which in turn lessens the associated costs remarkably. With the availability of long period of measured hygrothermal data, the performance of the developed models adaptively enhances and thus able to generate more long-term predictions. In addition, increasing the number and types of the monitored structures that are situated in different locations results in getting more divers data that help to make the model results to be applicable for a wide range of cases.

### 4.4.1 Corrosion status of façade elements

In order to determine the protection performance of the applied surface-treatment systems against corrosion of reinforcement bar, the correlation between the hygrothermal and the corrosion rate is needed. The corrosion rate of the embedded reinforcement bar can be computed using Equation (4.4) [156–158].

$$r = C_T r_o, \quad (4.4)$$

where $r$ is the rate of corrosion [µm/year], $C_T$ is the temperature coefficient, and $r_o$ is the rate of corrosion at +20 °C [µm/year]. $C_T$ and $r_o$ are time-variant variables which are dependent on the temperature and relative humidity inside the concrete, respectively. For carbonated concrete, $C_T$ and $r_o$ can be described by Equations (4.5) to (4.7) [156,158].

$$C_T = 1.6 \cdot 10^{-7}(30 + T)^4, \quad (4.5)$$

$$r_o = 190 \cdot (RH)^{26} \text{ when } RH \leq 0.95, \quad (4.6)$$

$$r_o = 2000 \cdot (1 - RH)^2 \text{ when } RH > 0.95, \quad (4.7)$$

where $T$ is the inner temperature [°C] and $RH$ is the inner relative humidity [%] of the pore structure.

The corrosion rate was computed by employing the measured and the predicted hygrothermal data only for the last one year of the data. The computed corrosion rate values are translated to corrosion status based on the classification (passive, low, moderate, and high) as given in Table 4.7, [159,160]. The corrosion state of the last one year is visualized by utilizing the developed exploratory data analysis technique (Section 3.7) and it is illustrated in Figure 4.13. The developed data exploratory method helps to easily visualize the corrosion status of the façade elements throughout the year. The unidentified amount of the corrosion rate due to the malfunction of the hygrothermal probes in the reference concrete façade member is left blank in the figure (white colour). It can be noticed from Figure 4.13 that the state of corrosion analysed using the monitored and the predicted data is almost identical. In both situations, the corrosion status for the concrete façade members S4, S5, and S6 is low and passive, whereas for S1, S2, and S3 it ranges from passive to high. Among all the façade elements, S1 and S2 provide poor protection against the progression of corrosion. Even if the surface of both façade elements were coated by the same materials from different manufacturers, the coating applied on façade element S1 offers the poorest protection. This reveals that not only the type of the coating materials but also their source and/or application technique has influence on the protection performance of the coating. It can also be observed that the corrosion status is varying within a short period of time in façade elements S1 and S2. This demonstrates that corrosion rate prediction based on the traditional instantaneous electrochemical measurement may under or overestimates its value. Hence, implementation of a long-term hygrothermal monitoring and modelling strategy is essential.

**Table 4.7.** Classification of corrosion status.

| Corrosion rate [µm/year] | State of corrosion |
|---|---|
| < 1 | Passive |
| 1- 5 | Low |
| 5-10 | Moderate |
| >10 | High |

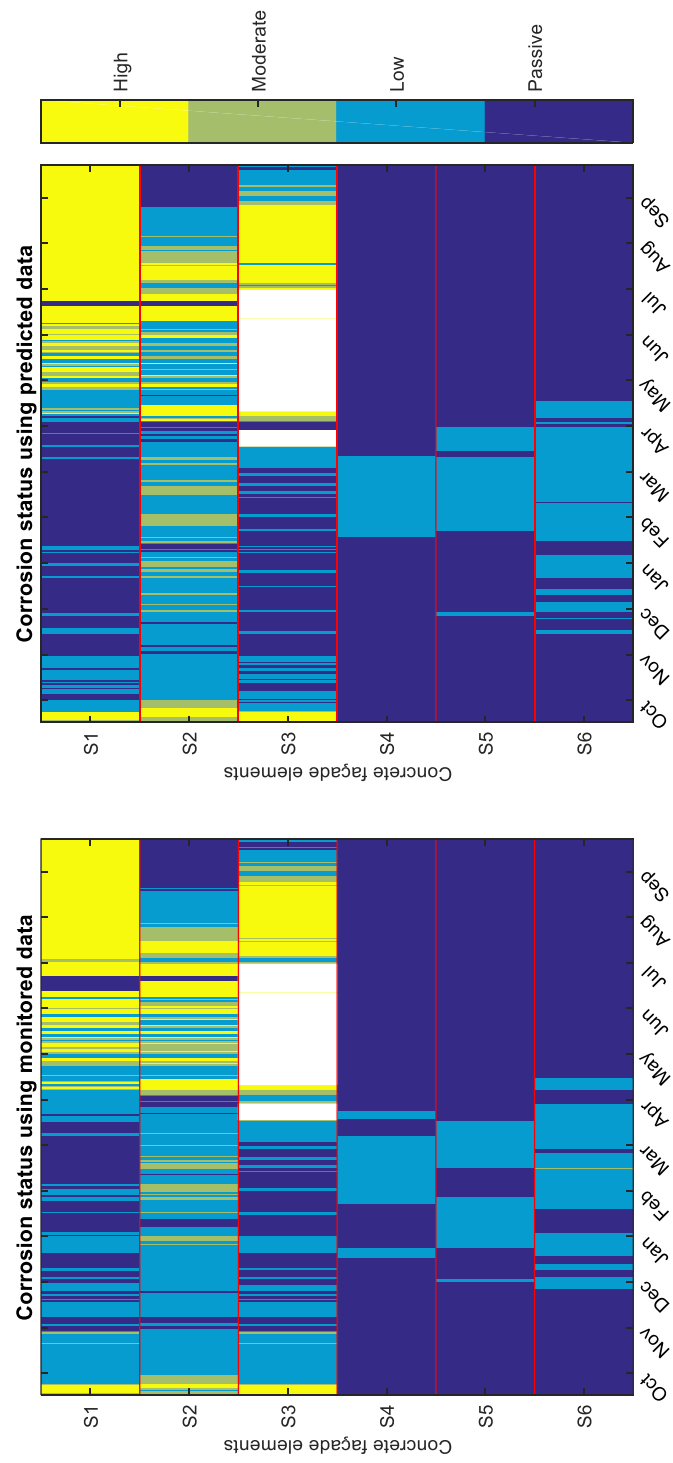

**Figure 4.13.** Corrosion statuses of the concrete façade elements (Publication IV).

### 4.4.2 Status of frost and chemical attacks

As discussed in Section 2.2.2, surface-protection systems are commonly applied on the surface of RC structures to protect the corrosion of reinforcement bar since they act as a physical barrier preventing the penetration of moisture. Nevertheless, some of the applied surface-protection systems may cause other type of deteriorations unintentionally. Due to this, the effectiveness of the surface-protection systems was evaluated by considering the effect of moisture on frost and chemical attacks. To analyse these attacks, four risk levels were defined (insignificant, slight, medium, and high) as in [28] and presented in Table 4.8.

The sensitivity of the surface-treated concrete façade elements for frost and chemical attacks on the second year was analysed using the monitored and the predicted data based on the defined risk levels. It is visualized using the developed exploratory data analysis method and illustrated in Figure 4.14. The unknown deterioration risk levels caused by the failure of the hygrothermal probes in the reference façade element is left blank and represented in the figure by white colour. It can be noticed from Figure 4.14 that the sensitivity differences between the monitored and the predicted data are inconsiderable for both type of attacks. Organic coatings outperform the cementitious ones in protecting the concrete elements against frost and chemical attacks while hindering the rate of corrosion. The cementitious coatings applied on the surface of S1 and S2 façade elements are ineffective in defending the concrete against frost and chemical attacks. Coating applied on surface of S1 is less protective than S2. This confirmed that the coating application techniques and their source influence the performance of the surface treatment in protecting against corrosion as well as frost and chemical attacks.

**Table 4.8.** Risk level classification of frost and chemical attacks.

| Effective relative humidity | Risk levels | |
|---|---|---|
| | Frost attack | Chmical attack |
| Very low (RH < 45%) | Insignificant | Insignificant |
| Low (RH = 45-65%) | Insignificant | Insignificant |
| Medium (RH = 65-85%) | Insignificant | Insignificant |
| High (RH = 85-98%) | Medium | Slight |
| Saturated (RH > 98%) | High | High |

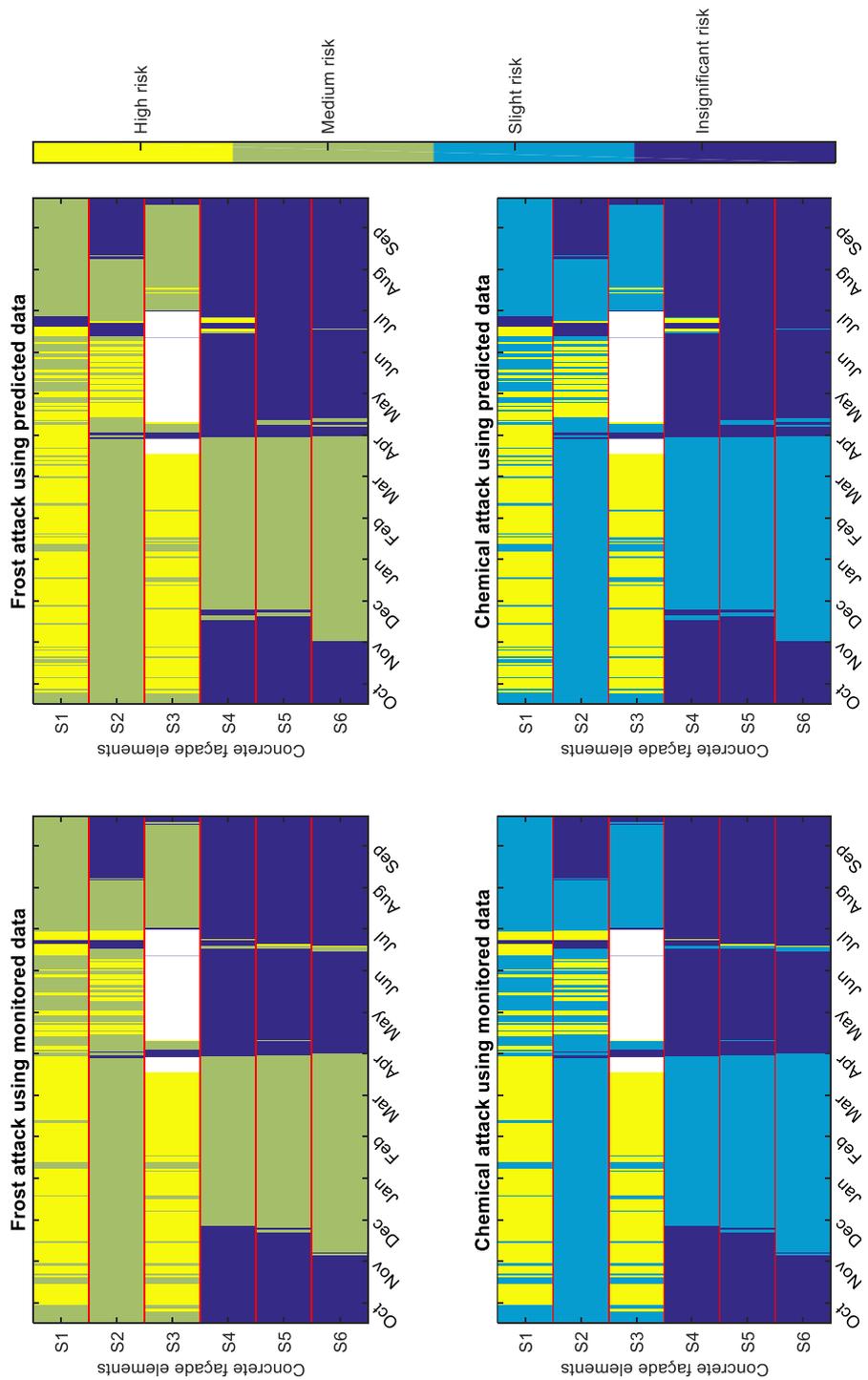

**Figure 4.14.** Statuses of frost and chemical attacks of concrete façade elements (Publication IV).

# Chapter 5

# Discussion

This chapter presents the implications of the dissertation findings as well as the reliability and validity of the undertaken research. It also discusses recommendation for future research directions. The discussion of theoretical and practical implications of the research to the scientific community, society and companies is necessary in order to facilitate the transfer of the discovered knowledge and methodologies. Assuring the reliability and validity of the research is vital in order to confirm the credibility of the findings of this dissertation.

## 5.1 Theoretical implications

As discussed in Chapter Two, the conventional models that predict the carbonation depth and the chloride concentration in concrete rely on limited parameters. Clear understanding of the significance of all involving parameters that describe the carbonation process and the penetration of chloride ions into concrete is critical in order to develop reliable and reasonably precise prediction models. The research results of this dissertation have established methods for prediction of carbonation depth and the ingression of chloride ions into concrete as well as measuring of parameters significance. These methods are based on machine learning techniques.

The findings based on the developed method of parameters significance measure revealed that accelerated carbonation depth, w/b and compressive strength are the three foremost powerful predictors of the carbonation depth. These are well recognized parameters and have been considered in the conventional models. The next three determined carbonation depth predictors are the amount of plasticizers, the carbonation period, and the air content. This is an interesting finding since these parameters, except the carbonation period, are overlooked in various conventional models. This finding is supported by the previous works [143,144]. These earlier researches claimed that different types of plasticizers govern the carbonation resistance of concrete through their influence on the pore structure and the morphology of the hydrated product. The research results of this dissertation also highlight the importance of air-entraining admixtures in characterising the carbonation process. Though the main purpose of air-entraining admixtures is to improve the resistance of the concrete against frost attack, it also affects the carbonation process. In fact, the discovered parameter by the CaPrM was the air content. This parameter can explain the effect of air-entraining admixtures as they heavily controlled it. The finding of air-entraining admixtures as influential descriptor is complemented by earlier researches [145,146]. These researches were made on concrete specimens of different mix compositions exposed to indoor and outdoor environments. They pointed out that the air-entraining admixtures increase the carbonation resistance of concrete.

The discovered topmost powerful predictors of chloride penetration representing the mix composition of the concrete are cement content, amount of total effective water, aggregate size distribution, supplementary cementitious materials, amount and type of plasticizers. The cement content and the amount of total effective water (usually designated in the form of water-to-cement ratio, w/c) are taken into consideration in numerous traditional chloride concentration prediction models. Nevertheless, the aggregate size distribution is not considered in the conventional models and its role in describing chloride penetration into concrete is often controversial. There are earlier researches that concluded the aggregate content influences the chloride penetration into concrete but not the aggregate size distribution [16,17]. But they come up with this finding by validating the model with mortars exposed to chloride environment for short term, three weeks and 15 months. Indeed, the findings of this dissertation agree with [16,17] for the case of one year exposure but contradicting them at older ages (three and six years). In addition to the aggregate size distribution, the findings of this research revealed that utilization of supplementary cementitious materials in concrete production plays a role in characterizing the chloride penetration into concrete at later age. Contrary to the aggregate size distribution, the amount and type of plasticizers describe the chloride profile of concrete at one-year exposure than older age. All these findings confirm that generalizing results that are acquired from short-term laboratory/field examination is inappropriate.

The dissertation discovered the amount and type of plasticizers as a predominant descriptor of the chloride profile. This parameter is not considered in the conventional chloride concertation prediction models. Indeed, there are several studies which demonstrated that plasticizers alter the pore characteristics of the hardened concrete. However, the effect of the type of plasticizers on chloride permeability is still insufficiently studied. The finding of the type of plasticizers as influential predictor of chloride profile is supported by previous study in [152]. The findings of this dissertation also showed that the air-entraining admixture influences the chloride ingress into the concrete. This parameter is not taken into an account by the conventional models. Even if very few researches have been performed on understanding the effect of air-entraining admixtures on chloride transport, the research in [150] is in line with the above finding.

Among several advanced fresh and hardened concrete properties, the compressive strength was determined as an influential descriptor of chloride concentration in concrete. This is not a new finding and it has been considered by several conventional models as a predictor. Next to this parameter, $D_{nssm}$ was discovered as a predominant chloride profile predictor. The significance of this predictor is higher at earlier age than exposure time of three and six years. In scenario where a range of fresh and hardened concrete properties were considered, the findings revealed that the prediction power of $D_{nssm}$ to describe the chloride profile has diminished. Other few properties that describe the pore volume of the concrete (air content, air pores as well as capillary and gel porosity) characterize the chloride concentration better than $D_{nssm}$. All these facts demonstrate that several advanced laboratory tests executed at early age are insignificant in predicting the chloride concentration in concrete. In addition, the nonsteady-state chloride diffusion coefficient derived from the chloride profile at early age is not always describing the chloride profile of concrete.

All the above findings have important theoretical implications. The results of the developed data-driven models proof their ability in isolating the effect of particular parameters out of several interdependent complex corrosion causing parameters. Understanding the contribution of parameters in predicting the carbonation depth and the chloride penetration could help researchers to focus their research on concrete materials that can resist well carbonation and chloride ingression. In addition, discovering the optimal influential parameters that best predict the carbonation depth and the chloride profile is critical in order to develop parsimonious and accurate model. In another perspective, the results of this dissertation serve as a show case for researchers to conduct similar scientific research on other concrete deterioration mechanisms by adopting data-driven models.

## 5.2 Practical implications

The penetrations of $CO_2$ and $Cl^-$ into concrete are the predominant factors that initiate corrosion of reinforcement bar. Accurate estimations of the carbonation depth and the chloride concentration in concrete are crucial to make realistic decisions regarding the maintenance plan of RC structures. Establishing analytical models that fully represent the carbonation process and the chloride ingression is challenging. This is due to the complex nature of the degradation mechanisms as well as the availability of a wide variety of concrete mix types and exposure conditions. In another perspective, once the corrosion initiates, the rate of corrosion is primarily governed by the hygrothermal interaction through their influence on the electrochemical reactions. In this dissertation, data-driven models that evaluate the carbonation depth, chloride profile and hygrothermal behaviour were developed.

The carbonation depth and chloride profile prediction models were able to discover new influential parameters which were missed in the conventional models. The discovered useful knowledge could assist companies to produce concrete that has the ability to resist against carbonation and/or chloride attack. Using the developed models, the material engineer/concrete designer can evaluate the performance of the newly designed concrete against carbonation and chloride penetration. Both models can be applied to evaluate the depth of carbonation and the chloride profile anywhere as far as there is a readily available input data to train them. All these facts have huge economical implication to the society since the models assist in designing optimal concrete mixes and defining proactive maintenance plan, which in turn minimize the lifecycle costs.

The developed data-driven hygrothermal model is a practical approach to evaluate the performance of surface-protection systems. This model along with the exploratory data analysis technique allows a more realistic evaluation of corrosion condition and other deterioration mechanisms caused unintentionally by the implemented surface-protection systems. In practice, this evaluation result helps the engineer/owner in choosing appropriate surface-protection system for the RC structure under consideration. As in any data-driven models, longer-term prediction becomes achievable when the amount of training data increases. This helps in determining the deterioration of the surface-protection materials well in advance, leading to proactive maintenance which ultimately saves the lifecycle costs substantially.

All the developed data-driven models are robust and reproducible with no or little effort by anyone who already has data or can acquire monitoring systems to collect data. The models with long-term data can be used to carryout corrosion assessment of structures accurately, which in turn enable to perform timed maintenance measures. Indeed, as presented in Section 4.1, with the advancement of sensing technologies, infrastructure of the next generation will integrate the physical infrastructures with cyber infrastructures which comprise wireless sensors, networks and computing devices. There is no doubt that data-driven based models presented in this dissertation will be the integral components of tomorrow's cyber-physical infrastructure systems. So, the developed models will play a substantial role in making proactive decisions and/or preparing short- and long-term plans to manage RC structures efficiently.

## 5.3  Reliability and validity

The reliability and validity of the research undertaken in this dissertation have been confirmed. All the findings of this research were deduced using the developed data-driven prediction models of carbonation depth, chloride penetration, and hygrothermal interaction. The validity of all the models was verified with unseen data and the verification outcomes have proofed their high accuracy prediction capability. The reliability of all the models was validated by using different unseen/test dataset and analysing the residuals. Detail verification procedures of all the developed models are presented in Chapter Four. The resulting residuals of all the models fell within the acceptable range and demonstrated the high generalization ability of the models. To ensure the stability of the models that are developed to determine the significance of parameters, the whole processes were reiterated ten times and the average of the ten outputs were taken as a final result. Though, the averaged values were taken into consideration, the difference among the ten outcomes were insignificant. All these facts confirm that the research outcomes of this dissertation are reliable and valid.

Lack of spatially dispersed experimental data can be seen as limitations of this dissertation. The models that were developed to examine the significance of parameters in predicting chloride profile employ data obtained from concrete specimens which were located at Kotka, Finland. These specimens were placed in identical altitude and experience similar multi-deterioration actions throughout the field experiment period. Due to this, factors describing the boundary conditions of the concrete specimens, the amount and frequency of sprinkled deicing salt in the highway, the climatic situations, the altitude where the concrete specimens placed, the amount of carbon dioxide in the environment and the traffic density, are the same. It is obvious that these parameters play a significant role on the chloride penetration, and thus incorporating data from different field experiments could help to measure the importance of these factors. The same is also true in the case of the CaPrM. All the concrete specimens employed as input data for the CaPrM were placed at Espoo, Finland. The curing conditions and methods as well as the environmental situations for all concrete specimens were identical. Due to this, it was impossible to measure the importance of these parameters in predicting the carbonation depth. The data utilized in the hygrothermal model were also gathered from surface-protected concrete façade of a building situated in Vantaa, Finland. Except the applied coating materials and their application methods, all the conditions that govern the hygrothermal performance such as the concrete mix composition, the depth of concrete cover, and the curing condition of the concrete panels were identical. This was the reason for considering only the hygrothermal data for model training. It would have been interesting if these parameters were different and able to evaluate their effect on predicting the hygrothermal performance of surface-protected concrete panels. In addition, the data come from a single structure, lacking consideration of different environmental conditions. This limits the application of the model results in a wide range of cases.

## 5.4 Recommendations for future research

In this thesis, the addressed degradation mechanisms that cause and control the corrosion of reinforcement bar are carbonation, chloride penetration and hygrothermal interaction. These degradation mechanisms were dealt separately using data obtained from three different case structures, one for each deterioration mechanism. In fact, the chloride penetration study was carried out using concrete specimens exposed to deicing salts as well as freezing and thawing. But the effect of freeze-thaw cycling was not separately studied due to lack of spatially disperse concrete specimens in the training dataset. In reality, all the described degradation mechanisms affect the corrosion of reinforcement bar simultaneously or consequently. It is a well-known fact that the influence of the synergic deterioration progresses faster and severe than the effect caused by any single degradation process. With the advancement of durability sensors, it will soon be practical to simultaneously monitor all the parameters which describe the degradation mechanisms in a single concrete element. In the future, it is highly recommended to monitor all corrosion causing and controlling mechanisms simultaneously in a single concrete element in both laboratory and field environments. In addition, data should be collected from spatially dispersed RC structures. Once data are available, it is possible to follow the same modelling approaches that were presented in this dissertation and the significance of various parameters involved in the synergic deterioration mechanisms on corrosion of reinforcement bar can be understood. After determining the influence of several complex interacting parameters on corrosion of reinforcement bar, more reliable and valid model can be established in order to assess the corrosion of reinforcement bar. This assessment helps to define proactive maintenance plan that enable reduction of lifecycle costs. The discovered knowledge could also assist researchers and designers working in concrete durability to produce concrete that resist well these deterioration mechanisms.

# Chapter 6

# Conclusions

In this dissertation a data-driven framework for evaluating corrosion causing and accelerating factors in concrete structures is presented. The framework was realized through the developed data-driven carbonation depth, chloride profile and hygrothermal performance prediction models. By utilizing these models i) better prediction accuracy for all the models has been achieved; ii) previously overlooked influential carbonation and chloride predicting parameters have been discovered; iii) it has been discovered that the influence of some of the chloride profile predicting parameters vary significantly based on exposure time, and iv) efficient surface-protection materials that protect concrete façade elements from corrosion as well as chemical and frost attacks have been determined.

  The thesis demonstrated that estimation of corrosion onset utilizing conventionally applied corrosion assessment methods in the form of analytical equations is uncertain since they are formulated by considering several simplifications and assumptions. The developed data-driven models mitigate uncertainties by mapping inputs (multiple complex corrosion controlling parameters) to the output that closely approximate the desired output values. The mapping was performed through learning algorithms that are capable of handling highly nonlinear interacting parameters with rational computational time. This allows evaluating all the influential parameters as a group rather than individually, which in turn ensures the reliability of the prediction since imperative dependencies are not overlooked.

  The performance comparison of the developed carbonation depth prediction model with the conventional one confirmed the prediction superiority of the developed data-driven carbonation model. The error difference between the two was significant. The conventional one has about two-fold more MSE than the developed data-driven carbonation model. The performance analysis of the developed data-driven hygrothermal model revealed its prediction capability with low error. This data-driven approach is a better alternative method since understanding the interaction of different surface-protection systems with the substrate concrete is a highly complex process. The integrated exploratory data analysis technique with the hygrothermal model had eased the challenge of selecting appropriate surface-protection materials that protect effectively from corrosion as well as chemical and frost attacks.

The developed carbonation depth and chloride profile prediction models discovered the best predictors of carbonation depth and chloride ingression by measuring the significance of all the considered input parameters. Among the input parameters, the amount of plasticizers and the air content are the most predominant carbonation depth predictors. These two parameters are overlooked in various conventional models despite few studies that demonstrate the effect of plasticizers in improving the carbonation resistance of concrete. In case of chloride penetration, the aggregate size distribution, the amount and type of plasticizers, and the supplementary cementitious materials are among the discovered influential predictors. These parameters are missed in the conventional chloride concentration prediction models. This finding corroborates that several advanced laboratory tests (except those describe the air-void characteristics of concrete) executed at early age are insignificant in predicting chloride concentration in concrete. This proofs that best predictors of chloride permeability of concrete determined from short-term experiment are not always powerful predictors in long term. By employing the determined influential parameters, the MSE of the chloride profile prediction was decreased by up to 19% compared to the results that employ only conventionally agreed variables. This improvement in MSE confirmed that the developed data-driven models have the capability of determining optimal subset of influential variables that best predict the chloride profile. In addition, the parameters importance measure revealed that the effects of supplementary cementitious materials are more pronounced at a later age, whereas chloride migration coefficient influences at earlier age. It also demonstrated that several of the early-age fresh and hardened concrete property tests are insignificant in describing the chloride ingression into concrete. This proofs that evaluation of long-term chloride transport into concrete using short-term tests is unrealistic.

Understanding the influential predictors help companies to produce optimized concrete mix that is able to resist carbonation and chloride penetration and thus enable reduction of lifecycle costs. The same models can be applied to evaluate the performance of the newly designed concrete against carbonation and chloride penetration if they are exposed to similar conditions that are used in training the models. In addition, discovering the optimal influential parameters that best predict the carbonation depth and chloride profile assists researchers to develop efficient and parsimonious corrosion assessment models as well as focus their research on concrete materials that can resist well these deteriorations. Furthermore, the results of this dissertation serve as a show case for researchers to conduct similar scientific researches on other concrete deterioration mechanisms by developing data-driven models.

The developed models play substantial roles in making proactive decisions and/or preparing short- and long-term plans to manage RC structures efficiently. All the developed corrosion assessment models are robust and reproducible with no or little effort by anyone who already has data or can acquire monitoring systems to gather data. The prediction results of the models adaptively adjust depending on the input data (the composition of the case structure and exposure conditions). With the advancement of sensing technologies, the next generation infrastructure will have sensing systems integrated into them, making data driven based condition assessments the primary choice. Therefore, the presented data-driven framework will be one of the key components for smart aging management of RC structures.